\pdfoutput=1

\documentclass[11pt]{article}

\usepackage[final]{acl}

\usepackage{times}
\usepackage{latexsym}

\usepackage[T1]{fontenc}

\usepackage[utf8]{inputenc}
\usepackage{paralist}
\usepackage{microtype}
\usepackage{soul}
\usepackage{inconsolata}

\newcommand{\ie}{\textit{i.e.}\xspace}
\newcommand{\eg}{\textit{e.g.}\xspace}

\newcommand{\etc}{\textit{etc}\xspace}
\newcommand{\dq}[1]{``#1''}

\newcommand{\method}{\textsc{InCharacter}\xspace}
\newcommand{\chatgpt}{{GPT-3.5}\xspace}
\newcommand{\gptfour}{{GPT-4}\xspace}
\newcommand{\gemini}{{Gemini}\xspace}

\usepackage{arydshln}
\usepackage{array}
\usepackage{amsmath}
\usepackage{amssymb}
\usepackage{booktabs}
\usepackage{colortbl}
\usepackage{etoolbox}
\usepackage{enumitem}
\usepackage{makecell}
\usepackage{multirow}
\usepackage{xspace}
\usepackage{xcolor}
\usepackage{graphicx}
\usepackage{spverbatim}
\AtBeginEnvironment{spverbatim}{\small}

\definecolor{Ground}{RGB}{255,184,55}
\definecolor{Rice}{RGB}{251,248,238}
\definecolor{Dirt}{RGB}{191,169,115}
\definecolor{Pink}{RGB}{226,184,176}
\definecolor{Violet}{RGB}{163,148,170}
\definecolor{mygray}{RGB}{226, 226, 226}

\newcolumntype{g}{>{\columncolor{Ground!10}}c}
\newcolumntype{d}{>{\columncolor{Dirt!10}}c}
\newcolumntype{f}{>{\columncolor{Pink!10}}c}
\newcolumntype{v}{>{\columncolor{Violet!10}}c}
\newcolumntype{P}[1]{>{\centering\arraybackslash}p{#1}}

\makeatletter
\def\adl@drawiv#1#2#3{%
        \hskip.5\tabcolsep
        \xleaders#3{#2.5\@tempdimb #1{1}#2.5\@tempdimb}%
                #2\z@ plus1fil minus1fil\relax
        \hskip.5\tabcolsep}
\newcommand{\cdashlinelr}[1]{%
  \noalign{\vskip\aboverulesep
           \global\let\@dashdrawstore\adl@draw
           \global\let\adl@draw\adl@drawiv}
  \cdashline{#1}
  \noalign{\global\let\adl@draw\@dashdrawstore
           \vskip\belowrulesep}}
\makeatother

\title{\method: Evaluating Personality Fidelity in Role-Playing Agents through Psychological Interviews}

\author{
    Xintao Wang$^{1}$, Yunze Xiao$^{2}$, Jen-tse Huang$^{3}$, Siyu Yuan$^{4}$, Rui Xu$^{1}$, Haoran Guo$^{5}$, \\
    \textbf{Quan Tu$^{6}$, Yaying Fei$^{7}$, Ziang Leng$^{8}$, Wei Wang$^{1}$, Jiangjie Chen$^{1}$, Cheng Li$^{9}$, Yanghua Xiao$^{*1}$} \\ 
    $^1$Shanghai Key Laboratory of Data Science, School of Computer Science, Fudan University \\
    $^2$Carnegie Mellon University \quad
    $^3$The Chinese University of Hong Kong \\ 
    $^4$School of Data Science, Fudan University \quad 
    $^5$RhineAI \quad
    $^6$Renmin University of China \\
    $^7$Beijing University of Technology \quad 
    $^8$Boston University \quad 
    $^9$SenseTime \quad \\
    \texttt{\{xtwang21, syyuan21, ruixu21\}@m.fudan.edu.cn,} \\
    \texttt{\{jjchen19, weiwang1, shawyh\}@fudan.edu.cn} \\ 
    \texttt{yunzex@andrew.cmu.edu, jthuang@cse.cuhk.edu.hk, chengli@sensetime.com} 
}

\begin{document}
\maketitle
\def\thefootnote{*}\footnotetext{Corresponding author.}\def\thefootnote{\arabic{footnote}}
\begin{abstract}
Role-playing agents (RPAs), powered by large language models, have emerged as a flourishing field of applications.
However, a key challenge lies in assessing whether RPAs accurately reproduce the personas of target characters, namely their character fidelity.
Existing methods mainly focus on the knowledge and linguistic patterns of characters.
This paper, instead, introduces a novel perspective to evaluate the personality fidelity of RPAs with psychological scales.
Overcoming drawbacks of previous self-report assessments on RPAs, we propose \method, namely \textbf{In}terviewing \textbf{Character} agents for personality tests.
Experiments include various types of RPAs and LLMs, covering 32 distinct characters on 14 widely used psychological scales.
The results validate the effectiveness of \method in measuring RPA personalities.
Then, with \method, we show that state-of-the-art RPAs exhibit personalities highly aligned with the human-perceived personalities of the characters, achieving an accuracy up to 80.7\%.\footnote{Project page: \url{https://incharacter.github.io/}}

\end{abstract}

\section{Introduction}

Recent advancements in large language models (LLMs) have catalyzed the emergence of role-playing agents (RPAs).
RPAs are interactive AI systems simulating diverse roles or characters.
RPA applications have been extended to diverse contexts, such as AI agents of fictional characters~\citep{li2023chatharuhi}, digital clones for humans~\citep{gao2023livechat}, and AI non-player characters in video games~\citep{wang2023voyager}. 
Recent research trends have increasingly focused on the development of RPAs, including building RPAs for specific characters~\citep{li2023chatharuhi, wang2023rolellm} and improving the role-playing abilities of foundation models~\citep{zhou2023characterglm}.

\begin{figure}[t]
    \centering
    \includegraphics[width=\linewidth]{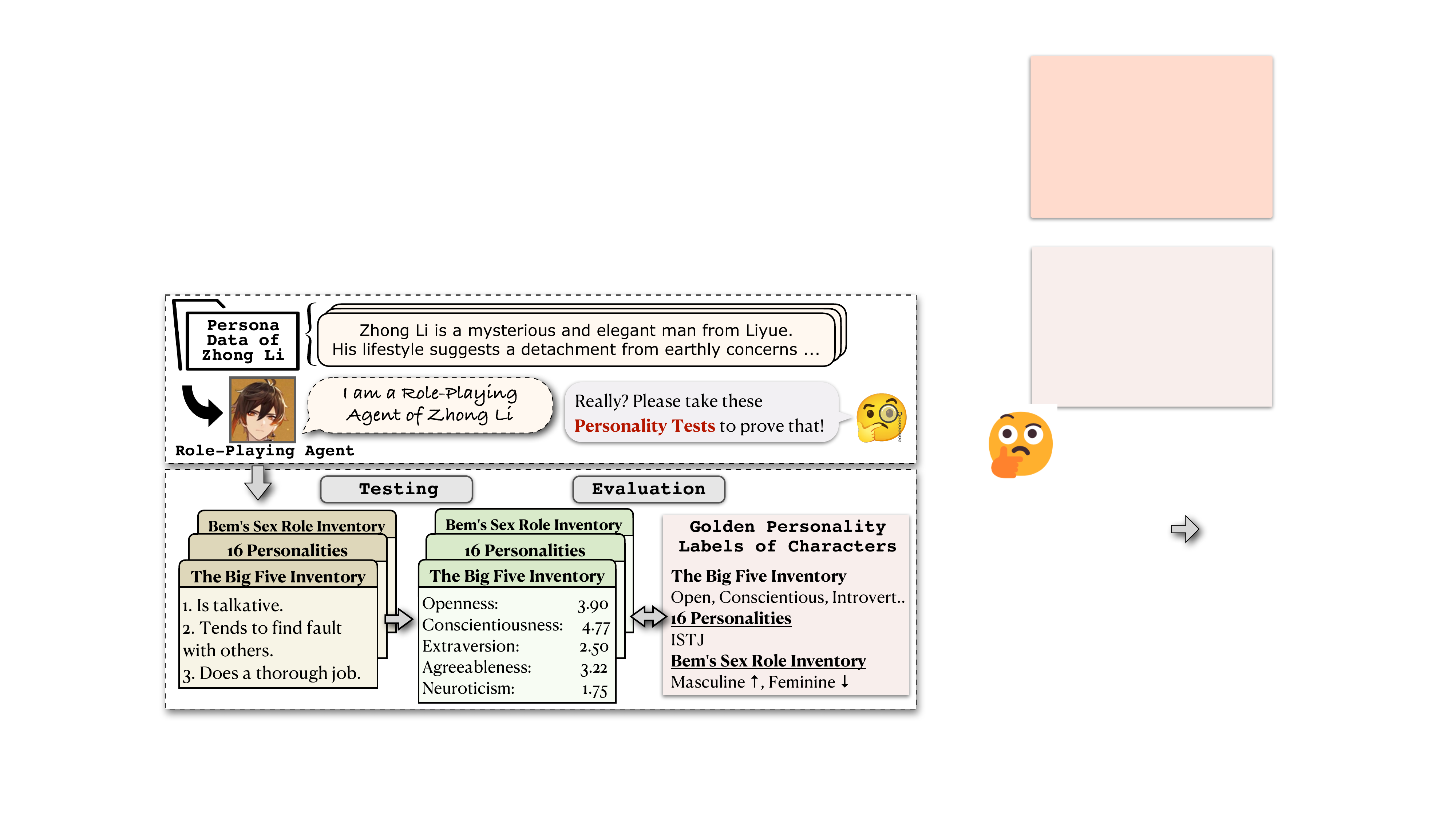}
    \caption{
    The procedure of personality tests on RPAs. To evaluate the personality fidelity of RPAs, we apply various scales to measure their personalities and compare the results with the personality labels of the characters.
    }
    \label{fig:1}
\end{figure}

However, evaluating \textit{character fidelity} in RPAs remains a relatively underexplored area.
Prior research mainly concentrates on the  replication of knowledge, experience, and linguistic patterns of characters~\citep{shao-etal-2023-character,zhou2023characterglm}, which manifests in two primary issues:
(1) They necessitate character-specific datasets, thereby complicating the evaluation of new characters.
(2) They overlook evaluating RPAs' thoughts and underlying mindsets.
Towards these issues, we propose to evaluate if RPAs faithfully reproduce the personalities of target characters, i.e., \textit{personality fidelity}, as depicted in Figure~\ref{fig:1}.
Personality tests, administered by psychological scales, measure an individual's interrelated behavioral, cognitive, and emotional patterns~\cite{barrick1991big, bem1981bem}.
By measuring the personalities of RPAs and comparing them with the personalities of the characters, we can attain a more nuanced understanding of RPAs' character fidelity.

Prior studies on LLM personalities are mainly based on self-report scales, which prompt LLMs to select options or assign ratings to specific items~\cite{tu2023characterchat, huang2023chatgpt}.
However, this method suffers from several limitations for RPAs.
(1) The instruction to complete scales contradicts role-playing instructions, leading to RPAs' reluctance or inability to engage with personality tests.
(2) More importantly, the selected options may conflict with the actual behaviors of RPAs, making the test results unindicative of their true personalities. 
RPAs might underperform owing to an inadequate understanding of scale instructions and the biases inherent in the training data.

Therefore, we propose \method, a novel approach to \textbf{In}terviews \textbf{Character} agents for personality tests.
While self-report scales are popular in humans for their cost-effectiveness, interview-based scales evaluated by experts offer a more comprehensive analysis~\cite{uher2012self, rush1987comparison}.
Self-reports are sometimes influenced by an individual's lack of insight, denial, or bias.
In contrast, an interviewer can be a guide to elicit thoughts of individuals, effectively identifying and addressing the nuances via conversations to overcome the previously mentioned limitations.
\method employs this interview-based procedure~\citep{trull1998structured} on RPAs, which includes two stages:
(1) \textbf{Interview}: RPAs are engaged with open-ended questions derived from psychological scales to elicit RPAs' mindsets and behaviors.
(2) \textbf{Assessment}: We utilize LLMs to interpret the responses collected from the first stage.
This can involve converting the responses to Likert levels or using LLMs to simulate a psychiatrist's role in judging RPA personalities.

We apply \method to various RPAs on 14 personality tests, including the Big Five Inventory (BFI), 16Personalities\footnote{\url{https://www.16personalities.com/}. This scale is based on the MBTI and is widely used worldwide.} (16P), and Dark Triad Dirty Dozen (DTDD).
The personality labels for the BFI and 16P are accessible through the Personality Database (PDb)\footnote{\url{https://www.personality-database.com/}}.
Additionally, we engage human annotators familiar with the characters to label them on other scales, thereby creating a comprehensive benchmark for evaluating RPA personalities.
Our experiments include various types of existing RPAs. %, crafted by ChatHaruhi~\citep{li2023chatharuhi}, RoleLLM~\citep{wang2023rolellm} and \texttt{c.ai}, spanning 32 distinct characters.
The results demonstrate that the \method effectively simulates interview-based tests conducted by human experts and yields RPAs personality measurement better aligned with the characters than self-report methods.  
Additionally, we find that the conversation data generated in our interview phase are of high quality and ideal for fine-tuning foundation models for RPAs. Hence, we release a dataset comprising  18,304 interview dialogues to facilitate future research.

The contributions of this paper are mainly three-fold:
\begin{inparaenum}[\it 1)]
    \item We introduce a novel aspect for RPA evaluation, \ie, personality fidelity, based on psychological scales. 
    \item We propose \method, an interview-based framework for personality tests on RPAs and collect the first benchmark for RPA personality evaluation, facilitating future research on developing better RPAs.
    \item Our experiments on various RPAs and psychological scales demonstrate the efficacy of \method. 
\end{inparaenum}
\section{Preliminaries}

\subsection{Role-Playing Agents}

Recent advancements have led to the emergence and evolution of several pivotal abilities in LLMs to facilitate the development of RPAs, including in-context learning~\cite{NEURIPS2020_gpt3}, instruction following~\cite{NEURIPS2022_instructgpt}, step-by-step reasoning~\cite{NEURIPS2022_cot}, and human-like traits such as empathy~\cite{sorin2023large}.
RPAs are interactive AI systems that act as assigned personas, from fictional characters to celebrities.
RPAs utilize persona data to simulate characters, drawing from training datasets, prompted contexts, or external databases.
Typically, existing work develops RPAs by setting character descriptions as system prompts~\cite{zhou2023characterglm, shao-etal-2023-character} and crafting memory modules with character dialogues~\cite{li2023chatharuhi, wang2023rolellm}. 

\begin{figure*}[t]
    \centering
    \includegraphics[width=\linewidth]{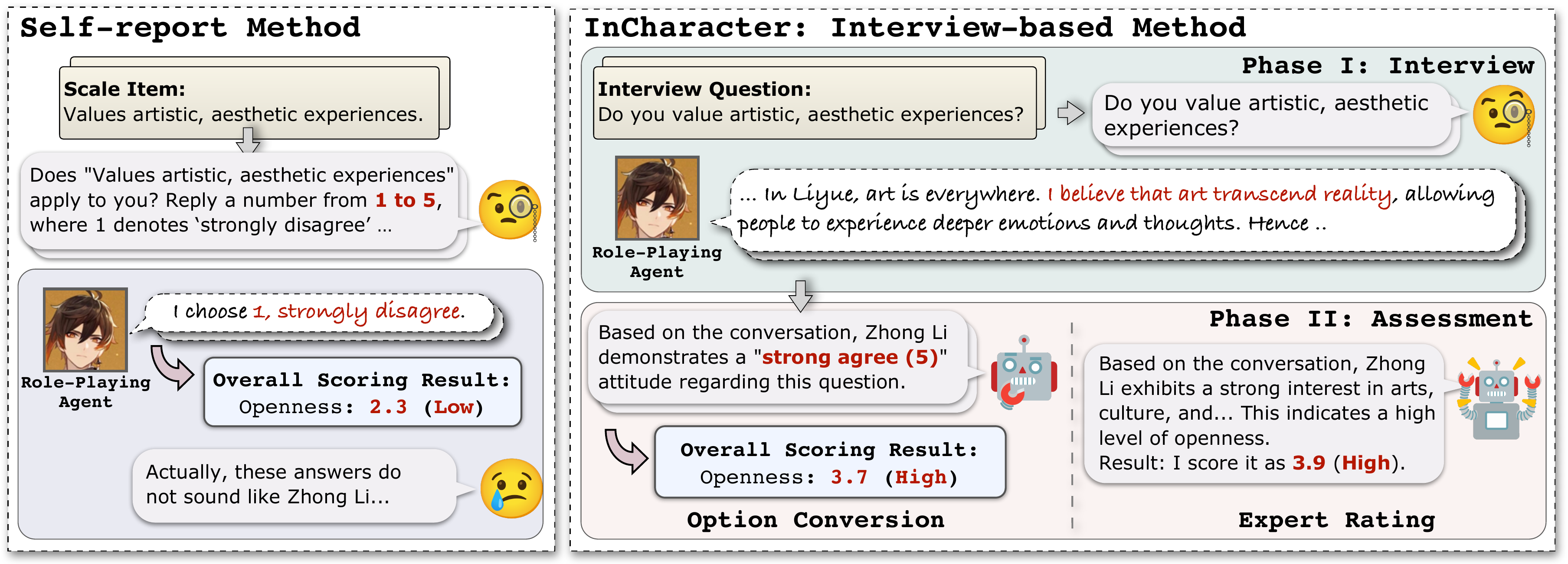}
    \caption{The framework of \method for personality tests on RPAs. \textbf{Left}: Previous methods use self-report scales, which prompt LLMs to select an option directly.
    \textbf{Right}: \method adopts an interview-based approach comprising two phases: the interview and assessment phases.
    The interview phase elicits the behavioral, cognitive, and emotional patterns of RPAs that reflect their underlying mindsets.
    The assessment phase measures personalities based on interview results, with two alternative methodologies:  option conversion and expert rating.}
    \label{fig:2}
\end{figure*}

\subsection{Psychological Scales}
\label{sec:likert}

Usually rated on Likert levels, psychological scales are commonly used for personality tests.
Self-report scales require participants to respond to a series of items analyzed through a specific scoring scheme to determine their personality traits.
A scale rated on Likert levels, denoted as $\mathcal{L}=(\mathcal{P}, \mathcal{D}, \mathcal{O}, f)$, comprises a set of items $\mathcal{P}$ (\ie, a questionnaire), a list of dimensions $\mathcal{D}$, a set of response options $\mathcal{O}$, and a scoring scheme $f$.
Each item $p\in\mathcal{P}$ is a statement or question, positively or negatively corresponding to a dimension $d\in\mathcal{D}$.
For example, the item \dq{\textit{Values artistic, aesthetic experiences.}} is positively related to the \textit{Openness} dimension in the BFI.
Participants select an ordinal response $o\in\mathcal{O}$ for each item, such as \textit{Agree}.
Typically, these options are numerically coded, \eg, \dq{1} for \textit{Strongly Disagree} and \dq{5} for \textit{Strongly Agree}.
This process generates a response array $\mathcal{A}$. 
The scoring schema $f$ usually includes item-dimension mapping, identification of positive and negative items, conversion of options to scores, and an aggregation method (\eg, average or sum).
Finally, the participant's personality scores $\mathcal{S}$ is derived as $\mathcal{S}=f(\mathcal{A})$, where $\mathcal{S}=(s_{d_1}, s_{d_2}, ..., s_{d_{|\mathcal{D}|}})$ represents scores across each dimension. 
We summarize the notations used in this paper in Table~\ref{tab:notations}.
\section{\method}
\label{sec:method}

This section introduces \method, a novel personality assessment methodology designed explicitly for RPAs, utilizing an interview-based procedure.
Figure~\ref{fig:2} illustrates our two-stage framework.
The interview stage is detailed in \S\ref{sec:interview}, followed by an elaboration of the assessment stage in \S\ref{sec:assessment}.

\subsection{Interview}
\label{sec:interview}

\method draws inspiration from the \textit{Structured Interview} approach used in psychological testing~\citep{trull1998structured}.
For a given scale, it transforms scale items into a series of open-ended questions, forming the basis for a structured interview.
Then, our framework interviews RPAs using these open-ended questions to elicit their perspectives on topics indicative of personality traits.

\paragraph{Constructing Question List}

We develop the structured interview question list based on items of the scale.
Specifically, each item $p\in\mathcal{P}$ is transformed into an open-ended question $q$ via LLMs and manually checked.
Consequently, the question list $\mathcal{Q}$ comprises $|\mathcal{P}|$ questions.
For instance, in the BFI, the item \dq{\textit{Values artistic, aesthetic experiences.}} is rephrased as \dq{\textit{Do you values artistic, aesthetic experiences?}}

\paragraph{Interviewing RPAs}

We interview an RPA $C$ of character $c$, by presenting each question $q\in\mathcal{Q}$ and recording its corresponding response $r$.
To avoid context effects~\cite {nikolic2010brain}, each question is posed in an isolated context, thereby avoiding potential interference among the questions.

\subsection{Assessment}
\label{sec:assessment}

Based on the interview results, the assessment phase quantitatively evaluates the score $s_d$ of the RPA $C$ across each dimension $d\in\mathcal{D}$.
To this end, we introduce two distinct methodologies for measuring and analyzing RPA personalities leveraging LLMs: option conversion (OC) and expert rating (ER).

\paragraph{Option Conversion}

This technique leverages LLMs to convert a response $r$ for a question $q$ into a corresponding answer option $a\in\mathcal{O}$ for item $p$, effectively bridging the gap between closed-ended and open-ended question formats.
The idea follows the clinician-rated scales used in clinical psychiatry~\citep{cuijpers2010self, uher2012self}, where professional clinicians assign ratings to each scale item based on their observations during patient interviews and compute the final scores following the scale's scoring scheme.
For example, a response \dq{\textit{I believe that art transcends reality...}} is converted to \dq{5 \textit{(Strongly Agree)}} for the item.
Afterward, the answer list $\mathcal{A}$ is input to the scoring scheme $f$ to compute the final personality scores.
In practice, we observe that even state-of-the-art LLMs like \gptfour~\citep{openai2023gpt4} exhibit notable inaccuracies in categorizing the attitudes of RPAs. 
Therefore, we further introduce a dimensional-specific option conversion (d-OC) strategy, which divides $(q, r)$ pairs according to dimensions and substitutes Likert levels, such as \dq{4 \textit{(Agree)}} and \dq{2 \textit{(Disagree)}}, with more descriptive options like \dq{4 \textit{(Extroverted)}} and \dq{2 \textit{(Introverted)}} in the prompts for LLMs.

\paragraph{Expert Rating}

In contrast with the one-by-one question conversion in OC, this method applies LLMs to directly evaluate personality scores of RPAs in each dimension, considering all corresponding $(q, r)$ pairs.
This idea draws inspiration from the structured clinical interview in clinical psychiatry~\citep{first2014structured}, where clinicians assess patients using a predefined question list and derive final scores based on the responses without intermediate ratings or scoring schemes.
The interviewer LLM~\footnote{Interviewer LLMs denote the models for the ER and OC tasks, analogous to human interviewers.} is prompted with comprehensive descriptions of the scale, dimension, and score range.
It then generates the final personality score for each dimension based on the pertinent responses.
The advantage of ER is that it re-implements the scoring schema with the interviewer LLM, which can intelligently weigh individual $(q, r)$ pairs 
instead of using equal weights in OC.
Hence, it better recognizes personality-indicative responses from RPAs.

Details of our prompts for OC and ER are available in \S\ref{sec:prompts} in the appendix.
To prevent the influence of data leakage in ER and d-OC, \ie, the interviewer LLM might have memorized the characters' personality types. Hence, we anonymize the character names in the input prompts.
\section{Experimental Setup}

\subsection{Preliminary Study}

\setlength\tabcolsep{3pt}
\begin{table}[t]
\small
  \centering
    \begin{tabular}{lcccc}
    \toprule
    \textbf{LLMs} & \textbf{Acc.} & \textbf{Pearson's $r$} & \textbf{Spearman's $\rho$} & \textbf{Kendall's $\tau$}  \\
    \midrule
    \rowcolor[rgb]{ .949,  .953,  .961} \multicolumn{5}{c}{\textit{Option Conversation}} \\
    \gemini &   $69.5$ & 54.5 & 55.9 & 53.2\\  
     \chatgpt &   $57.5$ & 34.6 & 36.2 & 32.4\\
     \gptfour &   \textbf{71.0} & \textbf{60.0} & \textbf{64.3} & \textbf{59.5}\\
     
    \midrule

    \rowcolor[rgb]{ .949,  .953,  .961} \multicolumn{5}{c}
    {\textit{Dimension-specific Option Conversation}} \\
      \gemini &   $79.0$ & 79.6 & 80.6  & 75.9\\  
     \chatgpt &   $76.5$ & 79.2 & 81.7 & 74.5 \\
     \gptfour &  \textbf{82.0} & \textbf{84.7} & \textbf{85.3}  & \textbf{80.6} \\
     
    \midrule

    \rowcolor[rgb]{ .949,  .953,  .961} \multicolumn{5}{c}
    {\textit{Expert Rating (batch)}} \\
     \gemini &   $84.0$ & 83.9 & 85.7 & 76.6 \\   
     \chatgpt &  $84.0$ & 90.6 & 89.9 & 80.4 \\
     \gptfour &  \textbf{89.0} & \textbf{92.5} & \textbf{92.7} & \textbf{83.7}  \\

    \bottomrule
    \end{tabular}
  \caption{
  The accuracy (\textbf{Acc.}) and consistency measurements of interviewer LLMs on the OC or ER tasks, compared with human labels.
  }
  \label{tab:human_eval}
\end{table}

\paragraph{Can LLMs Simulate Human Interviewers?}
First, we validate the capability of interviewer LLMs on the OC and ER tasks, given the interview results of RPAs. 
We compare LLM predictions with human judgments. 
For each task, we sample 100 cases of state-of-the-art RPAs on the BFI and manually label them. 
For OC, the input is one question-response pair $(q, a)$ and the output is an option.
For ER, the input is composed of multiple $(q, a)$ pairs, and the output is a score. 
Examples are shown in the right of Figure~\ref{fig:2}. 
We experiment with three LLMs, including \gptfour, \chatgpt and \gemini.
The RPAs are detailed in \S\ref{sec:settings}. 
We report the Pearson’s $r$~\citep{pearson1920notes}, Spearman's $\rho$~\citep{spearman1961proof} and Kendall's $\tau$~\citep{kendall1938new} correlations between human annotations and interviewer LLMs, as well as the accuracy. 
We consider LLM predictions varying from human labels by less than 1 point, exactly 1, or more than 1, as \textit{right}, \textit{close} (half-correct) or \textit{wrong}, for accuracy calculation. 
More details can be found in \S\ref{sec:interviewer_llms}.

The results presented in Table~\ref{tab:human_eval} lead to several findings. 
First, for ER, state-of-the-art LLMs can adequately rate participants' personalities based on interview results.
We observe that \gptfour makes only 4\% \textit{wrong} cases in ER, primarily when RPAs give contradictory responses.
Second, for OC, the LLMs show significant inaccuracy, while replacing Likert-level options with dimension-descriptive ones (d-OC) largely improves LLMs in this task.
Considering the consistency measurements, %between interviewer LLMs and humans,
state-of-the-art LLMs achieve acceptable performance in simulating human interviewers to assess RPA personalities through ER or d-OC.

\subsection{Experimental Settings}
\label{sec:settings}

The experiments in the subsequent part of this paper are based on the following settings:

\paragraph{RPAs and Characters}

This work primarily focuses on RPAs built on character data curated by ChatHaruhi~\citep{li2023chatharuhi} and RoleLLM~\citep{wang2023rolellm}. 
We select 32 widely-known characters, 16 from ChatHaruhi\footnote{
Six RPAs from ChatHaruhi are based on Chinese data, and we conduct the interview with them in Chinese.
} 
and 16 from RoleLLM. 
The characters are mainly from popular fictional works, such as \textit{Harry Potter}, \textit{The Big Bang Theory} and \textit{Genshin Impact}.
Please refer to \S\ref{sec:character_selection} for the detailed character selection process.
The character data from ChatHaruhi and RoleLLM includes descriptions and dialogues used for system prompts and memory modules. 
To implement RPAs, we apply the Chat-Haruhi-Suzumiya\footnote{A continuously updating project for RPAs. We refer ChatHaruhi~\citep{li2023chatharuhi} to its first version. \url{https://github.com/LC1332/Chat-Haruhi-Suzumiya}} library, and adopt \chatgpt~\citep{chatgpt} as the foundation LLM by default.

\paragraph{Psychological Scales}

We consider 14 psychological scales, including the BFI, the 16P, and 12 other scales following PsychoBench\footnote{\url{https://github.com/CUHK-ARISE/PsychoBench}}~\citep{huang2023who} to evaluate RPAs.
Most scales apply scoring schemes like average and sum, while the 16P is close-source and accessed via its API.
Detailed introduction of these scales can be found in \S\ref{sec:scales}.
Due to page limitations, the main body presents results for the BFI and 16P, while additional findings are detailed in the Appendix.

\paragraph{Personality Labels}

We collect labels for character personalities in the form of both \textit{scores} and \textit{types}, contributed by people familiar with these characters.
From the PDb, an online platform for character personality annotation, we derive \textit{scores} of the BFI and 16P on each dimension from its label percentage (\eg, $60\%$ Extroverted).
We then categorize it into a \textit{type} of either positive, negative, or marginal if it is above $60\%$, under $40\%$, or otherwise. 
Then, we invite human annotators for comprehensive personality labels on all 14 scales. 
To select qualified annotators, we examine their character understanding of the BFI and 16P, matching with labels from the PDb. 
We invite two to three annotators for each character (93 in total for 32 characters) and average their results for improved reliability and objectivity. 
The scores are re-scaled into the unit interval $[0, 1]$ and categorized into \textit{types} similarly.
We measure the inter-annotator consistency via Cohen's kappa coefficient~\citep{Cohen1968}, and find the average coefficient across 14 scales $60.9\%$.
For the BFI and 16P, we adopt \textit{types} from the PDb and \textit{scores} from our invited annotators.
The details about PDb annotations, our human annotation process, intra-annotator consistency, and other statistics can be found in \S\ref{sec:annotation}. 

\paragraph{Interviewer LLMs}

We use LLMs to accomplish the OC, d-OC and ER tasks in the assessment phase of \method, or to extract selected options from RPA responses in self-report methods if RPAs do not provide exactly the choice.
We consider four widely-acknowledged LLMs, including \chatgpt, \gptfour, \gemini and Qwen1.5-110B \footnote{The versions in this paper are \texttt{gpt-3.5-turbo-1106}, \texttt{gpt-4-1106-preview}, \texttt{gemini-pro}, \texttt{Qwen1.5-110B-Chat} respectively.}. 

\paragraph{Metrics} 

We consider two sets of metrics, namely:

\noindent (1) \textbf{Measured alignment} (MA) compares the measured personalities of RPAs and human-annotated personalities of characters. 
It depends both on the performance of RPAs and the effectiveness of personality test methods.
We categorize RPAs as \textit{positive} or \textit{negative} on each dimension if the scores are above or below the median of the scoring range. % and corresponding characters
Then, we calculate mean absolute error (\textbf{MAE}) and accuracy to measure alignment at the \textit{score} and \textit{type} level, respectively. 
We re-scale \textbf{MAE} by dividing it with the scoring range length. %to re-scale it into the unit interval. 
For accuracy, we report the average \textbf{Acc$_\texttt{Dim}$} and  \textbf{Acc$_\texttt{Full}$}, where correctness is judged on individual or all dimensions of each scale. 
The marginal dimensions of each character are ignored due to their ambiguity. 

\noindent (2) \textbf{Personality consistency} (PC) indicates whether the measured personality of RPAs is consistent across various settings. 
We analyze the standard variance at the item-level (\textbf{Std$_\texttt{Item}$}), dimension-level (\textbf{Std$_\texttt{Dim}$}), and score-level (\textbf{Std$_\texttt{Score}$}). 
\textbf{Std$_\texttt{Item}$} and \textbf{Std$_\texttt{Dim}$} measure the consistency of an RPA's scores on individual items. 
For \method, we experiment with OC and d-OC to convert responses into scores.
\textbf{Std$_\texttt{Item}$} measures an RPA's consistency on the same item across multiple runs. 
\textbf{Std$_\texttt{Dim}$} compares an RPA's responses across different items in the same dimension.  
\textbf{Std$_\texttt{Score}$} denotes the variance of an RPA's score on each dimension across multiple runs.
We divide these metrics by the length of the corresponding scoring range to re-scale them into the unit interval.

\section{Experimental Results}

\subsection{Personality Tests on RPAs}

\paragraph{Baselines}

For \method, we experiment with the ER, OC, and d-OC.
For ER, we consider two settings, ER$_\texttt{all}$ and ER$_\texttt{batch}$, where question-response pairs in one dimension are inputted into interviewer LLMs all-at-once or in-batch\footnote{In the BFI and 16P, a dimension generally has 9 to 15 examples. Hence, the in-batch setting applies a batch size of 3 to 4. The results are averaged to form the final scores.}.
For self-report (SR) baselines, we follow previous work on LLM Psychometrics~\citep{huang2023who} to prompt RPAs to provide exactly a choice for each scale item. 
If their responses are not exactly the choices, we use interviewer LLMs to extract the choices.
Then, the numbers are aggregated via the scoring schema to get the results. 
Besides, we experiment with SR-CoT, which enhances SR with chain-of-thought reasoning, \ie, explicitly asking RPAs to articulate their thoughts before choosing the options.

We compare these methods on the BFI and 16P. 
The experiments are repeated three times, including both the interview phase and the assessment phase.  
We report the average results of the three runs for MA metrics and \textbf{Std$_\texttt{Dim}$}, and calculate \textbf{Std$_\texttt{Item}$} and \textbf{Std$_\texttt{Score}$} across the three runs.

\setlength\tabcolsep{2.8pt}
\begin{table*}[ht]
\small
  \centering
    \begin{tabular}{llggggggvvvvvv}
    \toprule
    \multirow{3}{*}{\textbf{Method}} & \multirow{3}{*}{\parbox{1.3cm}{\textbf{Interviewer Model}}} & \multicolumn{6}{c}{\textbf{The Big Five Inventory}} &  \multicolumn{6}{c}{\textbf{The 16 Personalities}} \\
    \cmidrule(lr){3-8} \cmidrule(lr){9-14} 
    & & \multicolumn{3}{c}{\textbf{MA}} & \multicolumn{3}{c}{\textbf{PC}} & \multicolumn{3}{c}{\textbf{MA}} & \multicolumn{3}{c}{\textbf{PC}} \\ 
    \cmidrule(lr){3-5} \cmidrule(lr){6-8} \cmidrule(lr){9-11} \cmidrule(lr){12-14}
    & 
    & \multicolumn{1}{c}{\textbf{Acc$_\texttt{Dim}$}} & \multicolumn{1}{c}{\textbf{Acc$_\texttt{Full}$}} & \multicolumn{1}{c}{\textbf{MAE} $\downarrow$} & \multicolumn{1}{c}{\textbf{Std$_\texttt{Item}$}} & \multicolumn{1}{c}{\textbf{Std$_\texttt{Dim}$}} & \multicolumn{1}{c}{\textbf{Std$_\texttt{Score}$}} 
    & \multicolumn{1}{c}{\textbf{Acc$_\texttt{Dim}$}} & \multicolumn{1}{c}{\textbf{Acc$_\texttt{Full}$}} & \multicolumn{1}{c}{\textbf{MAE} $\downarrow$} & \multicolumn{1}{c}{\textbf{Std$_\texttt{Item}$}} & \multicolumn{1}{c}{\textbf{Std$_\texttt{Dim}$}} & \multicolumn{1}{c}{\textbf{Std$_\texttt{Score}$}} \\
    \midrule
    
    \rowcolor[rgb]{ .949,  .953,  .961} \multicolumn{14}{c}{\textit{Self-report Methods}} \\

    \multirow{3}{*}{SR} 

    & Qwen-110B & 
    63.5 & 7.3 & 23.2 & 2.3 & 29.5 & 1.7 &
    62.3 & 22.9 & 27.2 & 2.9 & 38.3 & 1.8 \\

    & \gemini &
    63.3 & 7.3 & 23.6 & 2.0 & 30.2 & 1.5 & 
    65.3 & 21.9 & 26.8 & 5.5 & 46.1 & 2.1 \\
    
    & \chatgpt &
    63.7 & 7.3 & 23.4 & 2.3 & 30.2 & 1.5 & 
    66.1 & 22.9 & 26.7 & 2.8 & 38.2 & 1.9 \\
    
    & \gptfour  & 
    63.3 & 7.3 & 23.2 & 2.2 & 28.3 & 1.5 & 
    65.6 & 21.9 & 26.5 & 3.3 & 37.7 & 2.1 
    \\    \midrule 
    
    \multirow{3}{*}{SR-CoT} 
    & Qwen-110B & 
    65.0 & 10.4 & 22.8 & 13.4 & 26.5 & 5.5 &
    60.3 & 15.6 & 29.7 &18.2 & 36.2 & 5.3 \\
    
    & \gemini &
    66.2 & 8.3 & 22.6 & 13.2 & 26.1 & 5.3 & 
    66.7 & 21.9 & 26.1 & 15.7 & 33.9 & 5.1 \\
    
    & \chatgpt & 
    66.9 & 9.4 & 22.6 & 12.9 & 25.2 & 5.5 & 
    68.0 & 24.0 & 26.1 & 14.9 & 31.9 & 5.1  \\
    
    & \gptfour &
    67.1 & 9.4 & 22.3 & 12.5 & 25.0 & 5.1 &
    66.9 & 24.0 & 25.6 & 14.4 & 30.3 & 4.8
    \\ \midrule 

    \rowcolor[rgb]{ .949,  .953,  .961} \multicolumn{14}{c}{\method: \textit{Interview-based Methods}} \\

    \multirow{3}{*}{ $\text{OC}$ }  
    & Qwen-110B & 
    68.2 & 9.4 & 20.8 & 3.8 & 22.5 & 2.4 &
    72.5 & 29.2 & 26.1 & 3.5 & 26.9 & 2.0 \\
    &  \gemini & 
    72.2 & 14.6 & 21.3 & 6.8 & 27.6 & 5.1 & 
    66.1 & 25.0 & 27.3 & 6.1 & 30.4 & 2.6 \\
    & \chatgpt & 
    65.4 & 3.1 & 24.2 & 4.5 & 31.5 & 2.7 & 
    65.0 & 28.1 & 27.8 & 4.5 & 27.7 & 2.0 \\
    & \gptfour & 
    64.3 & 6.2 & 21.6 & 4.9 & 26.4 & 3.6 & 
    75.5 & 34.4 & 23.1 & 4.9 & 28.1 & 2.4 \\
    \midrule
    
    \multirow{3}{*}{ d-OC }  
    & Qwen-110B & 
    73.7 & 26.0 & 19.7 & 2.9 & 18.4 & 2.2 &
    74.9 & 40.6 & 23.2 & 3.3 & 21.4 & 2.2 \\
    & \gemini & 
    72.8 & 18.8 & 20.4 & 4.2 & 20.8 & 3.3 & 
    73.6 & 36.5 & 22.6 & 4.1 & 23.5 & 2.9 \\
    & \chatgpt & 
    64.1 & 5.2 & 22.9 & 5.0 & 18.0 & 3.8 & 
    76.9 & 40.6 & 21.8 & 6.4 & 22.7 & 4.6 \\
    & \gptfour & 
    72.2 & 14.6 & \underline{18.6} & 3.8 & 19.6 & 2.9 & 
    \underline{80.2} & \underline{45.8} & 21.2 & 3.3 & 21.3 & 2.1 \\
    
    \midrule
    
    \multirow{3}{*}{ ER$_\texttt{all}$ }  
    & Qwen-110B & 
    \underline{74.3} & 28.1 & 19.3 & - & - & 4.0 &
    78.0 & 37.5 & 21.0 & - & - & 4.2 \\
    & \gemini & 
    71.5 & 18.8 & 20.6 & - & - & 4.9 & 
    76.3 & 40.6 & 20.7 & - & - & 4.6 \\
    & \chatgpt & 
    74.1 & 25.0 & 20.5 & - & - & 5.2 &
    79.1 & \underline{45.8} & 22.1 & - & - & 5.9 \\
    & \gptfour & 
    \textbf{76.6} & \underline{30.2} & 18.9 & - & - & 4.0 & 
    79.6 & 43.8 & \textbf{20.1} & - & - & 4.4 \\
    
    \midrule

    \multirow{4}{*}{ ER$_\texttt{batch}$ }  
    & Qwen-110B & 
    72.4 & 22.9 & 18.6 & - & - & 3.6 &
    79.1 & \textbf{49.0} & 21.7 & - & - & 2.7 \\
    & \gemini & 
    73.9 & 24.0 & 19.2 & - & - & 4.7 &
    77.1 & 37.5 & 20.9 & - & - & 3.2 \\
    & \chatgpt & 
    72.4 & 22.9 & 18.9 & - & - & 4.5 &  
    78.5 & 43.8 & 22.2 & - & - & 4.5\\
    & \gptfour & 
    \textbf{76.6} & \textbf{31.2} & \textbf{18.2} & - & - & 3.6 &
    \textbf{80.7} & 44.8 & \underline{20.5} & - & - & 2.9 \\

    \bottomrule
    \end{tabular}%}
  \caption{
  Metrics on personalities of the selected RPAs were measured via various personality test methods on the big five inventory and 16 personalities. 
  For MA metrics, the best results are \textbf{bolded}, and the second best ones are \underline{underlined}.
  {\textbf{Std$_\texttt{Item}$}} and {\textbf{Std$_\texttt{Dim}$}} are derived from scores of individual items, and are hence inapplicable for ER.
  }
  \label{tab:test_methods}
\end{table*}

\paragraph{Alignment between RPAs' Measured Personalities and Characters' Labeled Personalities}
Then, we apply \method to measure RPA personalities.
According to the results in Table~\ref{tab:test_methods},  we have the following analyses: 
(1) 
Using \method~with ER and \gptfour, 
the measured RPA personalities are highly aligned with ground truth labels of corresponding characters.
This suggests that state-of-the-art RPAs well reproduce many of the characters' personality traits, and our methods accurately measure their personalities.
(2) RPA personalities measured via \method are better aligned with the characters than SR baselines. 
This validates the advantage of \method over self-report for personality tests on RPAs, which will be further discussed.
(3) The alignment measured via \method correlates with the interviewer LLMs' capability on the assessment tasks. 
For the interviewer LLMs, \gptfour achieves the best metrics, while \chatgpt, \gemini, and Qwen-110B also demonstrate satisfactory performance. 
For the assessment methods, \method with ER generally achieves better MA metrics than d-OC, while d-OC surpasses OC. 
However, Table~\ref{tab:human_eval} shows that interviewers LLMs still make mistakes on the ER and OC tasks, leading to potential inaccuracies in \method and may underestimate the personality alignment of RPAs.

\paragraph{Robustness, Consistency and Distinctiveness of RPA Personalities}

\begin{figure}[t]
    \centering
    \includegraphics[width=0.8\linewidth]{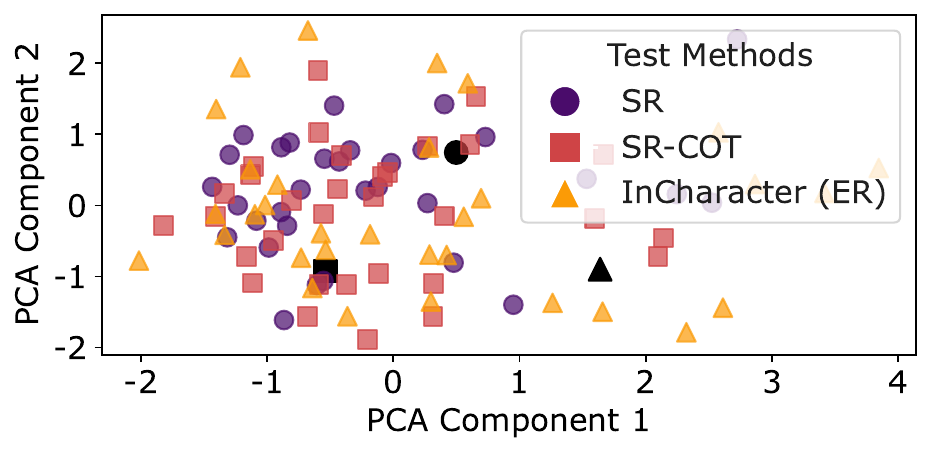}
    \caption{Visualization of 32 RPAs' personalities on the BFI  measured by different methods. 
    We use principal component analysis (PCA) to map the results into 2D spaces. Black points represent the personality of  \chatgpt measured by corresponding methods.
    }
    \label{fig:distribution}
\end{figure}

Generally, the measured RPA personalities are robust across our observations. 
The \textbf{Std$_\texttt{Score}$} across three runs remain below 6\% in various settings, which underlines the 
reliability of personality tests and the robustness of RPA personalities.
Then, we study the consistency at the item-level (\textbf{Std$_\texttt{Item}$}) and dimension-level (\textbf{Std$_\texttt{Dim}$}). 
With \method, after converting the interview results into scores via d-OC and \gptfour, 
We observe that RPAs respond to the same items consistently across multiple runs and exhibit a relatively consistent personality across different items on the same dimension. 
We visualize the distribution of RPA personalities on the BFI in Figure~\ref{fig:distribution}, and find that RPAs exhibit distinct personalities, especially when measured by \method with ER$_\texttt{batch}$ and \gptfour. 

\paragraph{Self-report v.s. Interview-based Methods}
As shown in Table~\ref{tab:test_methods}, the personalities measured by \method are more aligned with the characters, compared with self-report.
Meanwhile, in interview-based tests, RPAs exhibit more consistent personalities across different questions, as well as greater distinctiveness, shown in Figure~\ref{fig:distribution}.
These findings confirm the advantages of interview-based tests over self-report in measuring RPA personalities. 
Although SR-CoT attempts to enhance SR with the thought process, its improvement over SR is limited, and it encounters poor \textbf{Std$_\texttt{Item}$}. 
Further analyses and comparisons are detailed in \S\ref{sec:sr_vs_ib}.
We also experiment with enhancing self-report methods with in-context learning in \S\ref{sec:sr_icl}, which undermines the measurements instead.

\begin{table*}[t]
\small
  \centering
    \begin{tabular}{llgggvvv}
    \toprule
    \multicolumn{2}{c}{\textbf{Agent Types}} & \multicolumn{3}{c}{\textbf{The Big Five Inventory}} &  \multicolumn{3}{c}{\textbf{The 16 Personalities}} \\
    \cmidrule{1-2} \cmidrule(lr){3-5} \cmidrule(lr){6-8} 
    \multicolumn{1}{l}{\textbf{LLMs}} & \multicolumn{1}{l}{\textbf{Data}} & \multicolumn{1}{c}{\textbf{Acc$_\texttt{Dim}$}} & \multicolumn{1}{c}{\textbf{Acc$_\texttt{Full}$}} & \multicolumn{1}{c}{\textbf{MAE} $\downarrow$} 
    & \multicolumn{1}{c}{\textbf{Acc$_\texttt{Dim}$}} & \multicolumn{1}{c}{\textbf{Acc$_\texttt{Full}$}} & \multicolumn{1}{c}{\textbf{MAE} $\downarrow$} 
    \\
    \midrule

    \rowcolor[rgb]{ .949,  .953,  .961} \multicolumn{8}{c}{\textit{w/ General Open-source LLMs}} \\
    Qwen 7B & D+M
    & 60.5 & 9.4 & 24.3
    & 67.8 & 21.9 & 27.9
    \\ 
    
    OpenChat-3.5 7B & D+M
    & 63.1 & 6.2 & 23.1 
    & 76.9 & \underline{40.6} & 24.6
    \\ 
    
    Mistral-2 7B & D+M
    & 66.2 & 18.8 & 21.3 
    & 68.6 & 21.9 & 26.0
    \\ 

    LLaMa-2-Chat 13B & D+M
    & 66.9 & 12.5 & 26.8
    & 66.9 & 28.1 & 27.7 
    \\ 
    
    Mixtral 8x7B & D+M
    & 68.2 & 15.6 & 20.8 
    & 71.9 & 31.2 & 25.3
    \\ 

    \rowcolor[rgb]{ .949,  .953,  .961} \multicolumn{8}{c}{\textit{w/ Specialized Open-source LLMs}} \\
    CharacterGLM 6B & D+M
    & 54.1 & 0.0 & 25.8 
    & 52.1 & 15.6 & 29.7
    \\ 

    RP-Qwen 7B  & D+M
    & 60.5 & 0.0 & 23.8 
    & 64.5 & 15.6 & 28.6
    \\
    
    RP-Mistral-2 7B & D+M
    & 70.1 & 18.8 & 21.7 
    & 69.4 & 28.1 & 26.1 
    \\

    \rowcolor[rgb]{ .949,  .953,  .961} \multicolumn{8}{c}{\textit{w/ Close-source LLMs}} \\

    character.ai& D*
    & 52.2 & 9.4 & 31.2 
    & 52.9 & 21.9 & 31.6
    \\
    
    \chatgpt & D
    & 71.3 & \underline{21.9} & 21.1
    & \underline{78.5} & \textbf{43.8} & \textbf{22.0}
    \\ 
    
    \chatgpt & M
    & 71.3 & 18.8 & 21.8  
    & 71.9 & 31.2 & 26.0 \\  

    \chatgpt  & D+M
    & \underline{72.0} & \underline{21.9} & \textbf{18.8} 
    & \textbf{79.3} & \textbf{43.8} & \underline{22.6} \\
    
    \gptfour & D+M
    & \textbf{73.9} & \textbf{25.0} & \underline{19.8}
    & 76.0 & \textbf{43.8} & 23.2 \\

    \bottomrule
    \end{tabular}
  \caption{
  Measured alignment (\%) of RPAs with different foundation models and character data. 
  D and M represent descriptions and memories respectively, and D* denote private descriptions of character.ai. 
  }
  \label{tab:test_rpa_merged}
\end{table*}

\begin{figure}[t]
    \centering
    \includegraphics[width=0.7\linewidth]{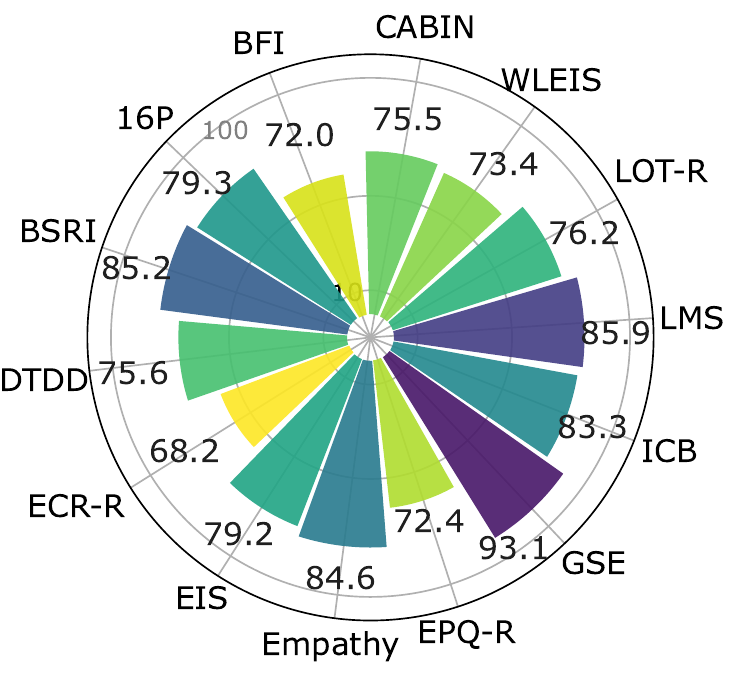}
    \caption{
    Measured alignment (\textbf{Acc$_\texttt{Dim}$}, \%) of state-of-the-art RPAs on 14 scales.
    }
    \label{fig:14scales}
\end{figure}

\paragraph{Comprehensive Personality Tests on 14 Scales} 
We extend personality tests on RPAs to 14 psychological scales, 
using \method with ER$_\texttt{batch}$ and \chatgpt. 
In Figure~\ref{fig:14scales}, we demonstrate the 
\textbf{Acc$_\texttt{Dim}$} of state-of-the-art RPAs across each scale. 
Overall, we observe that the RPAs exhibit personalities align with the target characters in comprehensive aspects with an average \textbf{Acc$_\texttt{Dim}$} of 78.9\%, covering personality traits (BFI, 16P), dark personalities (DTDD), interpersonal relationships (BSRI, ECR-R), basic interests (CABIN), motivation (GSE, LMS) and emotional intelligence (EIS, WLEIS), \etc.
The detailed metrics on individual scales and individual dimensions are listed in \S\ref{sec:results_14scales}.

\paragraph{More Results and Analyses} 
Besides, we study the importance of using personality scales in \S\ref{sec:importance_personality_scales}, and find that scale questions are crucial for eliciting personality-indicative responses from RPAs. 
We also observe that some scale questions go beyond the knowledge scope of certain characters in \S\ref{sec:adapting_questions}, and adapting these questions for each character yields more accurate measurements.

\subsection{Personality Fidelity of Different RPAs}
 
With \method, we compare the personality fidelity of various types of RPAs, covering different character data and foundation models. 
We apply \method with ER$_\texttt{batch}$ and use \chatgpt as the interviewer LLM for personality tests. 
We report the MA metrics on the BFI and 16P in Table~\ref{tab:test_rpa_merged}.

\paragraph{Character Data for RPAs}
Typically, existing RPAs utilize two types of character data: descriptions and memories. 
Character descriptions serve as the system prompts for RPAs, while memories consist of characters' experiences and dialogues used for retrieval. 
With \chatgpt, we evaluate RPAs with only descriptions (D), only memories (M), and a combination of both (D+M). 
The results in Table~\ref{tab:test_rpa_merged} reveal that: 
(1) With only description, RPAs achieve MA metrics close to the full D+M setup, highlighting the importance of character description in shaping RPA personalities. 
(2) RPAs can well mimic character personalities exhibited in their past experiences, \eg, extraversion and openness, even if the experiences are not directly related to scale questions.
Additionally, we compare RPAs with character data from ChatHaruhi, RoleLLM and character.ai (\texttt{c.ai}) in \S\ref{sec:rpas_different_works}.

\paragraph{Foundation Models for RPAs}
We consider three types of LLMs:
(1) General open-source models, including Qwen-7B~\citep{bai2023qwen}, OpenChat-3.5 7B~\citep{wang2023openchat}, Mistral-2 7B~\citep{jiang2023mistral}, Llama-2-chat 13B~\citep{touvron2023llama} and Mixtral 8x7B~\citep{jiang2024mixtral}.
(2) Specialized open-source models for RPAs, including Character-GLM 6B~\citep{zhou2023characterglm}, RP-Qwen 7B\footnote{\url{https://huggingface.co/silk-road/ChatHaruhi_RolePlaying_qwen_7b}}, and RP-Mistral-2 7B.
We train RP-Mistral-2 7B with details shown in \S\ref{sec:finetuning}.
(3) Close-source models: \chatgpt and \gptfour.

The results are shown in Table~\ref{tab:foundation_llms}.
We observe that, 
(1) RPAs with \chatgpt and \gptfour achieve the best personality fidelity, and \gptfour does not significantly surpass \chatgpt.
(2) With state-of-the-art open-source LLMs, RPAs can also reproduce character personalities. 
However, such capacity depends largely on their ability to use specific languages, shown in \S\ref{sec:found_models}.
(3) Incremental fine-tuning on open-source LLMs with role-playing datasets brings limited improvement in personality fidelity, especially when they are already equipped with excellent role-playing ability. 

\paragraph{Close-source RPAs}
We also experiment with close-source RPAs from character.ai, which are based on their private foundation models and data. 
As shown in Table~\ref{tab:foundation_llms}, these RPAs barely reproduce character personalities,  significantly underperforming \chatgpt (D) which shares a similar framework. 
According to our observation, while character.ai RPAs provide human-like answers, their answers tend to be compliant and pleasing to users, instead of reproducing the target characters. Examples and further analysis are shown in \S\ref{sec:compliant}.

\section{Related Work}

\textbf{Role-Playing Agents}\quad
RPAs learn and leverage character data in various ways, including training on raw scripts or dialogues~\citep{shao2023character}, prompting with character descriptions~\citep{zhou2023characterglm}, and retrieval from character experiences~\citep{li2023chatharuhi}.  
Existing efforts mainly focus on developing character-specific RPAs or foundation models for RPAs.
The former includes ChatHaruhi~\citep{li2023chatharuhi} and RoleLLM~\citep{wang2023rolellm}, which target well-established fictional characters. %, and present comprehensive frameworks to craft persona data including descriptions and dialogues sourced from corresponding origins.
The latter includes character.ai and CharacterGLM~\citep{zhou2023characterglm}. %, which usually aim at enabling users to have customized RPAs via simple prompts.
For evaluation, prior research mainly concentrates on two facets:
1) Character-independent capabilities, which include conversational abilities~\citep{duan2023botchat}, human-likeness~\citep{tu2024charactereval}, multi-turn consistency~\citep{shao-etal-2023-character}, and attractiveness~\citep{zhou2023characterglm};
2) Character fidelity, including the characters' knowledge, experience, and linguistic patterns~\citep{wang2023rolellm, shao-etal-2023-character}.
Overall, these methods generally require test sets for each character, and neglect the evaluation of RPAs' underlying mindset.  

\noindent \textbf{Psychological Analysis on LLMs}\quad
Recent studies conducted personality tests using the BFI~\citep{romero2023gpt, karra2022estimating, li2022gpt, jiang2022evaluating, safdari2023personality, bodroza2023personality}, the MBTI~\citep{rutinowski2023self, pan2023llms} on various LLMs.
Notably, \citet{huang2023revisiting} verified the reliability of the BFI on \chatgpt, while \citet{safdari2023personality} demonstrated the construct validity of the BFI on the PaLM model family.
Other studies also investigates other mental perspectives, such as emotions~\citep{huang2023emotionally}, values~\citep{miotto-etal-2022-gpt, rutinowski2023self, hartmann2023political}, consciousness~\citep{butlin2023consciousness}, and mental illness~\citep{coda2023inducing}.
Our research diverges by employing personality tests as an innovative approach to assessing character fidelity in RPAs.

\section{Conclusion}

In this study, we investigate the personality fidelity in RPAs, \ie, whether RPAs reproduce personalities of their intended characters. 
Addressing the shortcomings of previous methods on RPAs, we propose \method, an interview-based approach that accurately measures RPA personalities based on their elicited mindsets and behaviors.
Our experiments span various types of RPAs, covering 32 characters on 14 psychological scales. 
The results validate the effectiveness of \method in measuring RPA personalities. 
Afterward, with \method, we comprehensively evaluate personality fidelity in existing RPAs, discovering that state-of-the-art RPAs successfully portray many personality traits of the characters.

\section*{Limitations}

There are several limitations in this study.
First, the personality measurement in this paper relies on the interviewer LLMs.
Consequently, the accuracy of the measured results may be compromised by potential errors or biases inherent in LLMs, potentially leading to an underestimation of the personality fidelity in RPAs. 
Second, the personalities of humans or fictional characters can change over time.
Since we use one static personality label for a specific character, there may be noise in our evaluation.
For instance, the character of James Bond has experienced significant development over the past two decades across various films and television series.
Our character annotations are derived from a singular, fixed time point in his storyline.
Additionally, the progressive changes in RPA personalities remain unexplored within existing literature.
We leave the study of RPA personality dynamics for future research.

\section*{Ethical Statement}

We hereby acknowledge that all authors of this work are aware of the provided ACL Code of Ethics and honor the code of conduct. 

\paragraph{Use of Human Annotations} 
In conducting our research, we have employed a methodology that incorporates personality labels, which were gathered through the online platform and by engaging a group of annotators. These annotators, who are university students, play a crucial role in our research process. To ensure fair treatment and to value their contribution, we offer them compensation that significantly exceeds the local minimum wage standards. Moreover, we maintain transparency regarding the application and purpose of their annotations, securing their informed consent for the use of these annotations in our research endeavors. Additionally, we are committed to upholding the privacy rights of our annotators throughout the annotation process, ensuring a respectful and ethical research environment. 

\paragraph{Risks} 
In this paper, we introduce a novel approach, referred to as \method, designed to assess the personalities of Role-Play Agent (RPA) entities. An integral component of our evaluation process involves the use of interviewer Large Language Models (LLMs), which, while innovative, could potentially introduce bias into the assessment outcomes. It is important to acknowledge this limitation as LLMs may reflect the inherent biases present in their training data. Furthermore, our evaluation encompasses a comprehensive analysis across 14 personality scales, notably including the Dark Triad of Personality (DTDD) scale, which focuses on darker personality traits. While this inclusion is aimed at providing a thorough understanding of RPA personalities, it raises ethical concerns regarding the potential for generating harmful content. This aspect underscores the need for careful consideration and implementation of safeguards to mitigate the risks associated with exploring dark personality traits in RPAs. 

\section*{Acknowledgment}

This work is funded by the Science and Technology Commission of Shanghai Municipality Grant (No. 22511105902).
This work originates from a project on Chat-Haruhi-Suzumiya proposed by Cheng Li.
We owe thanks to the early contributors.
We are thankful for the support provided by Zheli Xuan at Wuhan University, and Dingding Hu at the Institute of Psychology, Chinese Academy of Sciences, who offered invaluable assistance as psychology researchers.
Our gratitude extends to our invited annotators, primarily from Fudan University, for their contribution of high-quality annotations of character personalities.
We also acknowledge the assistance provided by Rui Fu and Wenxin Gao at Fudan University during the annotation process.
Finally, we express our sincere gratitude for the precious comments and suggestions from Yikai Zhang, Xinfeng Yuan, and Shuang Li at Fudan University.

\bibliography{anthology,custom}

\begin{thebibliography}{49}
\expandafter\ifx\csname natexlab\endcsname\relax\def\natexlab#1{#1}\fi

\bibitem[{Bai et~al.(2023)Bai, Bai, Chu, Cui, Dang, Deng, Fan, Ge, Han, Huang et~al.}]{bai2023qwen}
Jinze Bai, Shuai Bai, Yunfei Chu, Zeyu Cui, Kai Dang, Xiaodong Deng, Yang Fan, Wenbin Ge, Yu~Han, Fei Huang, et~al. 2023.
\newblock \href {https://arxiv.org/abs/2309.16609} {Qwen technical report}.
\newblock \emph{ArXiv preprint}, abs/2309.16609.

\bibitem[{Barrick and Mount(1991)}]{barrick1991big}
Murray~R Barrick and Michael~K Mount. 1991.
\newblock The big five personality dimensions and job performance: a meta-analysis.
\newblock \emph{Personnel psychology}, 44(1):1--26.

\bibitem[{Bem(1981)}]{bem1981bem}
Sandra~L Bem. 1981.
\newblock Bem sex role inventory.
\newblock \emph{Journal of personality and social psychology}.

\bibitem[{Bodroza et~al.(2023)Bodroza, Dinic, and Bojic}]{bodroza2023personality}
Bojana Bodroza, Bojana~M Dinic, and Ljubisa Bojic. 2023.
\newblock \href {https://arxiv.org/abs/2306.04308} {Personality testing of gpt-3: Limited temporal reliability, but highlighted social desirability of gpt-3's personality instruments results}.
\newblock \emph{ArXiv preprint}, abs/2306.04308.

\bibitem[{Brown et~al.(2020)Brown, Mann, Ryder, Subbiah, Kaplan, Dhariwal, Neelakantan, Shyam, Sastry, Askell, Agarwal, Herbert{-}Voss, Krueger, Henighan, Child, Ramesh, Ziegler, Wu, Winter, Hesse, Chen, Sigler, Litwin, Gray, Chess, Clark, Berner, McCandlish, Radford, Sutskever, and Amodei}]{NEURIPS2020_gpt3}
Tom~B. Brown, Benjamin Mann, Nick Ryder, Melanie Subbiah, Jared Kaplan, Prafulla Dhariwal, Arvind Neelakantan, Pranav Shyam, Girish Sastry, Amanda Askell, Sandhini Agarwal, Ariel Herbert{-}Voss, Gretchen Krueger, Tom Henighan, Rewon Child, Aditya Ramesh, Daniel~M. Ziegler, Jeffrey Wu, Clemens Winter, Christopher Hesse, Mark Chen, Eric Sigler, Mateusz Litwin, Scott Gray, Benjamin Chess, Jack Clark, Christopher Berner, Sam McCandlish, Alec Radford, Ilya Sutskever, and Dario Amodei. 2020.
\newblock \href {https://proceedings.neurips.cc/paper/2020/hash/1457c0d6bfcb4967418bfb8ac142f64a-Abstract.html} {Language models are few-shot learners}.
\newblock In \emph{Advances in Neural Information Processing Systems 33: Annual Conference on Neural Information Processing Systems 2020, NeurIPS 2020, December 6-12, 2020, virtual}.

\bibitem[{Butlin et~al.(2023)Butlin, Long, Elmoznino, Bengio, Birch, Constant, Deane, Fleming, Frith, Ji et~al.}]{butlin2023consciousness}
Patrick Butlin, Robert Long, Eric Elmoznino, Yoshua Bengio, Jonathan Birch, Axel Constant, George Deane, Stephen~M Fleming, Chris Frith, Xu~Ji, et~al. 2023.
\newblock \href {https://arxiv.org/abs/2308.08708} {Consciousness in artificial intelligence: Insights from the science of consciousness}.
\newblock \emph{ArXiv preprint}, abs/2308.08708.

\bibitem[{Coda-Forno et~al.(2023)Coda-Forno, Witte, Jagadish, Binz, Akata, and Schulz}]{coda2023inducing}
Julian Coda-Forno, Kristin Witte, Akshay~K Jagadish, Marcel Binz, Zeynep Akata, and Eric Schulz. 2023.
\newblock \href {https://arxiv.org/abs/2304.11111} {Inducing anxiety in large language models increases exploration and bias}.
\newblock \emph{ArXiv preprint}, abs/2304.11111.

\bibitem[{Cohen(1968)}]{Cohen1968}
Jacob. Cohen. 1968.
\newblock \href {https://doi.org/10.1037/h0026256} {Weighed kappa: Nominal scale agreement with provision for scaled disagreement or partial credit}.
\newblock \emph{Psychological Bulletin}, 70(4):213--220.

\bibitem[{Cuijpers et~al.(2010)Cuijpers, Li, Hofmann, and Andersson}]{cuijpers2010self}
Pim Cuijpers, Juan Li, Stefan~G Hofmann, and Gerhard Andersson. 2010.
\newblock Self-reported versus clinician-rated symptoms of depression as outcome measures in psychotherapy research on depression: a meta-analysis.
\newblock \emph{Clinical psychology review}, 30(6):768--778.

\bibitem[{Duan et~al.(2023)Duan, Wei, Wang, Liu, Fang, Zhang, Lin, and Chen}]{duan2023botchat}
Haodong Duan, Jueqi Wei, Chonghua Wang, Hongwei Liu, Yixiao Fang, Songyang Zhang, Dahua Lin, and Kai Chen. 2023.
\newblock \href {https://arxiv.org/abs/2310.13650} {Botchat: Evaluating llms' capabilities of having multi-turn dialogues}.
\newblock \emph{ArXiv preprint}, abs/2310.13650.

\bibitem[{First(2014)}]{first2014structured}
Michael~B First. 2014.
\newblock Structured clinical interview for the dsm (scid).
\newblock \emph{The encyclopedia of clinical psychology}, pages 1--6.

\bibitem[{Gao et~al.(2023)Gao, Lian, Zhou, Fu, and Wang}]{gao2023livechat}
Jingsheng Gao, Yixin Lian, Ziyi Zhou, Yuzhuo Fu, and Baoyuan Wang. 2023.
\newblock \href {https://doi.org/10.18653/v1/2023.acl-long.858} {{L}ive{C}hat: A large-scale personalized dialogue dataset automatically constructed from live streaming}.
\newblock In \emph{Proceedings of the 61st Annual Meeting of the Association for Computational Linguistics (Volume 1: Long Papers)}, pages 15387--15405, Toronto, Canada. Association for Computational Linguistics.

\bibitem[{Hartmann et~al.(2023)Hartmann, Schwenzow, and Witte}]{hartmann2023political}
Jochen Hartmann, Jasper Schwenzow, and Maximilian Witte. 2023.
\newblock The political ideology of conversational ai: Converging evidence on chatgpt’s pro-environmental, left-libertarian orientation.
\newblock \emph{Available at SSRN 4316084}.

\bibitem[{Huang et~al.(2023{\natexlab{a}})Huang, Lam, Li, Ren, Wang, Jiao, Tu, and Lyu}]{huang2023emotionally}
Jen-tse Huang, Man~Ho Lam, Eric~John Li, Shujie Ren, Wenxuan Wang, Wenxiang Jiao, Zhaopeng Tu, and Michael~R Lyu. 2023{\natexlab{a}}.
\newblock \href {https://arxiv.org/abs/2308.03656} {Emotionally numb or empathetic? evaluating how llms feel using emotionbench}.
\newblock \emph{ArXiv preprint}, abs/2308.03656.

\bibitem[{Huang et~al.(2023{\natexlab{b}})Huang, Wang, Lam, Li, Jiao, and Lyu}]{huang2023chatgpt}
Jen-tse Huang, Wenxuan Wang, Man~Ho Lam, Eric~John Li, Wenxiang Jiao, and Michael~R Lyu. 2023{\natexlab{b}}.
\newblock \href {https://arxiv.org/abs/2305.19926} {Chatgpt an enfj, bard an istj: Empirical study on personalities of large language models}.
\newblock \emph{ArXiv preprint}, abs/2305.19926.

\bibitem[{Huang et~al.(2023{\natexlab{c}})Huang, Wang, Lam, Li, Jiao, and Lyu}]{huang2023revisiting}
Jen-tse Huang, Wenxuan Wang, Man~Ho Lam, Eric~John Li, Wenxiang Jiao, and Michael~R Lyu. 2023{\natexlab{c}}.
\newblock \href {https://arxiv.org/abs/2305.19926} {Revisiting the reliability of psychological scales on large language models}.
\newblock \emph{ArXiv preprint}, abs/2305.19926.

\bibitem[{Huang et~al.(2024)Huang, Wang, Li, Lam, Ren, Yuan, Jiao, Tu, and Lyu}]{huang2023who}
Jen-tse Huang, Wenxuan Wang, Eric~John Li, Man~Ho Lam, Shujie Ren, Youliang Yuan, Wenxiang Jiao, Zhaopeng Tu, and Michael~R Lyu. 2024.
\newblock Who is chatgpt? benchmarking llms' psychological portrayal using psychobench.
\newblock In \emph{Proceedings of the Twelfth International Conference on Learning Representations}.

\bibitem[{Jiang et~al.(2023)Jiang, Sablayrolles, Mensch, Bamford, Chaplot, Casas, Bressand, Lengyel, Lample, Saulnier et~al.}]{jiang2023mistral}
Albert~Q Jiang, Alexandre Sablayrolles, Arthur Mensch, Chris Bamford, Devendra~Singh Chaplot, Diego de~las Casas, Florian Bressand, Gianna Lengyel, Guillaume Lample, Lucile Saulnier, et~al. 2023.
\newblock \href {https://arxiv.org/abs/2310.06825} {Mistral 7b}.
\newblock \emph{ArXiv preprint}, abs/2310.06825.

\bibitem[{Jiang et~al.(2024)Jiang, Sablayrolles, Roux, Mensch, Savary, Bamford, Chaplot, Casas, Hanna, Bressand et~al.}]{jiang2024mixtral}
Albert~Q Jiang, Alexandre Sablayrolles, Antoine Roux, Arthur Mensch, Blanche Savary, Chris Bamford, Devendra~Singh Chaplot, Diego de~las Casas, Emma~Bou Hanna, Florian Bressand, et~al. 2024.
\newblock \href {https://arxiv.org/abs/2401.04088} {Mixtral of experts}.
\newblock \emph{ArXiv preprint}, abs/2401.04088.

\bibitem[{Jiang et~al.(2022)Jiang, Xu, Zhu, Han, Zhang, and Zhu}]{jiang2022evaluating}
Guangyuan Jiang, Manjie Xu, Song-Chun Zhu, Wenjuan Han, Chi Zhang, and Yixin Zhu. 2022.
\newblock \href {https://arxiv.org/abs/2206.07550} {Mpi: Evaluating and inducing personality in pre-trained language models}.
\newblock \emph{ArXiv preprint}, abs/2206.07550.

\bibitem[{Karra et~al.(2022)Karra, Nguyen, and Tulabandhula}]{karra2022estimating}
Saketh~Reddy Karra, Son~The Nguyen, and Theja Tulabandhula. 2022.
\newblock \href {https://arxiv.org/abs/2204.12000} {Estimating the personality of white-box language models}.
\newblock \emph{ArXiv preprint}, abs/2204.12000.

\bibitem[{Kendall(1938)}]{kendall1938new}
Maurice~G Kendall. 1938.
\newblock A new measure of rank correlation.
\newblock \emph{Biometrika}, 30(1/2):81--93.

\bibitem[{Li et~al.(2023)Li, Leng, Yan, Shen, Wang, MI, Fei, Feng, Yan, Wang et~al.}]{li2023chatharuhi}
Cheng Li, Ziang Leng, Chenxi Yan, Junyi Shen, Hao Wang, Weishi MI, Yaying Fei, Xiaoyang Feng, Song Yan, HaoSheng Wang, et~al. 2023.
\newblock \href {https://arxiv.org/abs/2308.09597} {Chatharuhi: Reviving anime character in reality via large language model}.
\newblock \emph{ArXiv preprint}, abs/2308.09597.

\bibitem[{Li et~al.(2022)Li, Li, Joty, Liu, Huang, Qiu, and Bing}]{li2022gpt}
Xingxuan Li, Yutong Li, Shafiq Joty, Linlin Liu, Fei Huang, Lin Qiu, and Lidong Bing. 2022.
\newblock \href {https://arxiv.org/abs/2212.10529} {Does gpt-3 demonstrate psychopathy? evaluating large language models from a psychological perspective}.
\newblock \emph{ArXiv preprint}, abs/2212.10529.

\bibitem[{Miotto et~al.(2022)Miotto, Rossberg, and Kleinberg}]{miotto-etal-2022-gpt}
Maril{\`u} Miotto, Nicola Rossberg, and Bennett Kleinberg. 2022.
\newblock \href {https://aclanthology.org/2022.nlpcss-1.24} {Who is {GPT}-3? an exploration of personality, values and demographics}.
\newblock In \emph{Proceedings of the Fifth Workshop on Natural Language Processing and Computational Social Science (NLP+CSS)}, pages 218--227, Abu Dhabi, UAE. Association for Computational Linguistics.

\bibitem[{Nikoli{\'c}(2010)}]{nikolic2010brain}
Danko Nikoli{\'c}. 2010.
\newblock The brain is a context machine.
\newblock \emph{Review of psychology}, 17(1):33--38.

\bibitem[{OpenAI(2022)}]{chatgpt}
OpenAI. 2022.
\newblock \href {https://openai.com/blog/chatgpt} {Openai: Introducing chatgpt}.

\bibitem[{OpenAI(2023)}]{openai2023gpt4}
OpenAI. 2023.
\newblock \href {http://arxiv.org/abs/2303.08774} {Gpt-4 technical report}.

\bibitem[{Ouyang et~al.(2022)Ouyang, Wu, Jiang, Almeida, Wainwright, Mishkin, Zhang, Agarwal, Slama, Ray, Schulman, Hilton, Kelton, Miller, Simens, Askell, Welinder, Christiano, Leike, and Lowe}]{NEURIPS2022_instructgpt}
Long Ouyang, Jeffrey Wu, Xu~Jiang, Diogo Almeida, Carroll Wainwright, Pamela Mishkin, Chong Zhang, Sandhini Agarwal, Katarina Slama, Alex Ray, John Schulman, Jacob Hilton, Fraser Kelton, Luke Miller, Maddie Simens, Amanda Askell, Peter Welinder, Paul~F Christiano, Jan Leike, and Ryan Lowe. 2022.
\newblock \href {https://proceedings.neurips.cc/paper_files/paper/2022/file/b1efde53be364a73914f58805a001731-Paper-Conference.pdf} {Training language models to follow instructions with human feedback}.
\newblock In \emph{Advances in Neural Information Processing Systems}, volume~35, pages 27730--27744. Curran Associates, Inc.

\bibitem[{Pan and Zeng(2023)}]{pan2023llms}
Keyu Pan and Yawen Zeng. 2023.
\newblock \href {https://arxiv.org/abs/2307.16180} {Do llms possess a personality? making the mbti test an amazing evaluation for large language models}.
\newblock \emph{ArXiv preprint}, abs/2307.16180.

\bibitem[{Pearson(1920)}]{pearson1920notes}
Karl Pearson. 1920.
\newblock Notes on the history of correlation.
\newblock \emph{Biometrika}, 13(1):25--45.

\bibitem[{Romero et~al.(2023)Romero, Fitz, and Nakatsuma}]{romero2023gpt}
Peter Romero, Stephen Fitz, and Teruo Nakatsuma. 2023.
\newblock Do gpt language models suffer from split personality disorder? the advent of substrate-free psychometrics.
\newblock \emph{ResearchSquare preprint}.

\bibitem[{Rush et~al.(1987)Rush, Hiser, and Giles}]{rush1987comparison}
A~John Rush, William Hiser, and Donna~E Giles. 1987.
\newblock A comparison of self-reported versus clinician-related symptoms in depression.
\newblock \emph{The Journal of clinical psychiatry}, 48(6):246--248.

\bibitem[{Rutinowski et~al.(2024)Rutinowski, Franke, Endendyk, Dormuth, Roidl, and Pauly}]{rutinowski2023self}
J{\'e}r{\^o}me Rutinowski, Sven Franke, Jan Endendyk, Ina Dormuth, Moritz Roidl, and Markus Pauly. 2024.
\newblock The self-perception and political biases of chatgpt.
\newblock \emph{Human Behavior and Emerging Technologies}, 2024(1):7115633.

\bibitem[{Safdari et~al.(2023)Safdari, Serapio-Garc{\'\i}a, Crepy, Fitz, Romero, Sun, Abdulhai, Faust, and Matari{\'c}}]{safdari2023personality}
Mustafa Safdari, Greg Serapio-Garc{\'\i}a, Cl{\'e}ment Crepy, Stephen Fitz, Peter Romero, Luning Sun, Marwa Abdulhai, Aleksandra Faust, and Maja Matari{\'c}. 2023.
\newblock \href {https://arxiv.org/abs/2307.00184} {Personality traits in large language models}.
\newblock \emph{ArXiv preprint}, abs/2307.00184.

\bibitem[{Shao et~al.(2023{\natexlab{a}})Shao, Li, Dai, and Qiu}]{shao-etal-2023-character}
Yunfan Shao, Linyang Li, Junqi Dai, and Xipeng Qiu. 2023{\natexlab{a}}.
\newblock \href {https://aclanthology.org/2023.emnlp-main.814} {Character-{LLM}: A trainable agent for role-playing}.
\newblock In \emph{Proceedings of the 2023 Conference on Empirical Methods in Natural Language Processing}, pages 13153--13187, Singapore. Association for Computational Linguistics.

\bibitem[{Shao et~al.(2023{\natexlab{b}})Shao, Li, Dai, and Qiu}]{shao2023character}
Yunfan Shao, Linyang Li, Junqi Dai, and Xipeng Qiu. 2023{\natexlab{b}}.
\newblock \href {https://aclanthology.org/2023.emnlp-main.814} {Character-{LLM}: A trainable agent for role-playing}.
\newblock In \emph{Proceedings of the 2023 Conference on Empirical Methods in Natural Language Processing}, pages 13153--13187, Singapore. Association for Computational Linguistics.

\bibitem[{Sorin et~al.(2023)Sorin, Brin, Barash, Konen, Charney, Nadkarni, and Klang}]{sorin2023large}
Vera Sorin, Danna Brin, Yiftach Barash, Eli Konen, Alexander Charney, Girish Nadkarni, and Eyal Klang. 2023.
\newblock Large language models (llms) and empathy-a systematic review.
\newblock \emph{medRxiv}, pages 2023--08.

\bibitem[{Spearman(1961)}]{spearman1961proof}
Charles Spearman. 1961.
\newblock The proof and measurement of association between two things.

\bibitem[{Touvron et~al.(2023)Touvron, Lavril, Izacard, Martinet, Lachaux, Lacroix, Rozière, Goyal, Hambro, Azhar, Rodriguez, Joulin, Grave, and Lample}]{touvron2023llama}
Hugo Touvron, Thibaut Lavril, Gautier Izacard, Xavier Martinet, Marie-Anne Lachaux, Timothée Lacroix, Baptiste Rozière, Naman Goyal, Eric Hambro, Faisal Azhar, Aurelien Rodriguez, Armand Joulin, Edouard Grave, and Guillaume Lample. 2023.
\newblock \href {http://arxiv.org/abs/2302.13971} {Llama: Open and efficient foundation language models}.

\bibitem[{Trull et~al.(1998)Trull, Widiger, Useda, Holcomb, Doan, Axelrod, Stern, and Gershuny}]{trull1998structured}
Timothy~J Trull, Thomas~A Widiger, J~David Useda, Jay Holcomb, Bao-Tran Doan, Seth~R Axelrod, Barry~L Stern, and Beth~S Gershuny. 1998.
\newblock A structured interview for the assessment of the five-factor model of personality.
\newblock \emph{Psychological assessment}, 10(3):229.

\bibitem[{Tu et~al.(2023)Tu, Chen, Li, Li, Shang, Zhao, Wang, and Yan}]{tu2023characterchat}
Quan Tu, Chuanqi Chen, Jinpeng Li, Yanran Li, Shuo Shang, Dongyan Zhao, Ran Wang, and Rui Yan. 2023.
\newblock \href {https://arxiv.org/abs/2308.10278} {Characterchat: Learning towards conversational ai with personalized social support}.
\newblock \emph{ArXiv preprint}, abs/2308.10278.

\bibitem[{Tu et~al.(2024)Tu, Fan, Tian, and Yan}]{tu2024charactereval}
Quan Tu, Shilong Fan, Zihang Tian, and Rui Yan. 2024.
\newblock \href {https://arxiv.org/abs/2401.01275} {Charactereval: A chinese benchmark for role-playing conversational agent evaluation}.
\newblock \emph{ArXiv preprint}, abs/2401.01275.

\bibitem[{Uher et~al.(2012)Uher, Perlis, Placentino, Dernov{\v{s}}ek, Henigsberg, Mors, Maier, McGuffin, and Farmer}]{uher2012self}
Rudolf Uher, Roy~H Perlis, Anna Placentino, Mojca~Zvezdana Dernov{\v{s}}ek, Neven Henigsberg, Ole Mors, Wolfgang Maier, Peter McGuffin, and Anne Farmer. 2012.
\newblock Self-report and clinician-rated measures of depression severity: can one replace the other?
\newblock \emph{Depression and anxiety}, 29(12):1043--1049.

\bibitem[{Wang et~al.(2024)Wang, Cheng, Zhan, Li, Song, and Liu}]{wang2023openchat}
Guan Wang, Sijie Cheng, Xianyuan Zhan, Xiangang Li, Sen Song, and Yang Liu. 2024.
\newblock \href {https://openreview.net/forum?id=AOJyfhWYHf} {Openchat: Advancing open-source language models with mixed-quality data}.
\newblock In \emph{The Twelfth International Conference on Learning Representations}.

\bibitem[{Wang et~al.(2023{\natexlab{a}})Wang, Xie, Jiang, Mandlekar, Xiao, Zhu, Fan, and Anandkumar}]{wang2023voyager}
Guanzhi Wang, Yuqi Xie, Yunfan Jiang, Ajay Mandlekar, Chaowei Xiao, Yuke Zhu, Linxi Fan, and Anima Anandkumar. 2023{\natexlab{a}}.
\newblock Voyager: An open-ended embodied agent with large language models.
\newblock \emph{arXiv preprint arXiv: Arxiv-2305.16291}.

\bibitem[{Wang et~al.(2023{\natexlab{b}})Wang, Peng, Que, Liu, Zhou, Wu, Guo, Gan, Ni, Zhang et~al.}]{wang2023rolellm}
Zekun~Moore Wang, Zhongyuan Peng, Haoran Que, Jiaheng Liu, Wangchunshu Zhou, Yuhan Wu, Hongcheng Guo, Ruitong Gan, Zehao Ni, Man Zhang, et~al. 2023{\natexlab{b}}.
\newblock \href {https://arxiv.org/abs/2310.00746} {Rolellm: Benchmarking, eliciting, and enhancing role-playing abilities of large language models}.
\newblock \emph{ArXiv preprint}, abs/2310.00746.

\bibitem[{Wei et~al.(2022)Wei, Wang, Schuurmans, Bosma, ichter, Xia, Chi, Le, and Zhou}]{NEURIPS2022_cot}
Jason Wei, Xuezhi Wang, Dale Schuurmans, Maarten Bosma, brian ichter, Fei Xia, Ed~Chi, Quoc~V Le, and Denny Zhou. 2022.
\newblock \href {https://proceedings.neurips.cc/paper_files/paper/2022/file/9d5609613524ecf4f15af0f7b31abca4-Paper-Conference.pdf} {Chain-of-thought prompting elicits reasoning in large language models}.
\newblock In \emph{Advances in Neural Information Processing Systems}, volume~35, pages 24824--24837. Curran Associates, Inc.

\bibitem[{Zhou et~al.(2023)Zhou, Chen, Wan, Wen, Song, Yu, Huang, Peng, Yang, Xiao et~al.}]{zhou2023characterglm}
Jinfeng Zhou, Zhuang Chen, Dazhen Wan, Bosi Wen, Yi~Song, Jifan Yu, Yongkang Huang, Libiao Peng, Jiaming Yang, Xiyao Xiao, et~al. 2023.
\newblock \href {https://arxiv.org/abs/2311.16832} {Characterglm: Customizing chinese conversational ai characters with large language models}.
\newblock \emph{ArXiv preprint}, abs/2311.16832.

\end{thebibliography}

\appendix
\clearpage
\appendix
\section{Notation Table}

\setlength\tabcolsep{3pt}
\begin{table}[h]
\small
  \centering
    \begin{tabular}{P{0.14\columnwidth}|p{0.83\columnwidth}}
    \toprule
    \textbf{} & \textbf{Definition} \\
    \midrule
    \rowcolor[rgb]{ .949,  .953,  .961} \multicolumn{2}{c}{\textit{Task Formulation}} \\

    $\mathcal{L}$ & A Likert Scale, composed of $\mathcal{L}=(\mathcal{P}, \mathcal{D}, \mathcal{O}, f)$. \\
    $\mathcal{P}$ & A set of items, where $p\in\mathcal{P}$ is an item. \\
    $\mathcal{D}$ & A list of dimensions, where $d\in\mathcal{D}$ is a dimension. \\
    $\mathcal{O}$ & A set of options, where $o\in\mathcal{O}$ is an option for a $p$. \\
    $f$ & A scoring scheme, \ie, a function. \\ 
    $\mathcal{A}$ & A response list, where $a\in\mathcal{A}$ is a response for a $p$. \\
    $\mathcal{S}$ & The participant's personality scores across each dimension, \ie, $\mathcal{S}=(s_{d_1}, s_{d_2}, ..., s_{d_{|\mathcal{D}|}})$. \\
    $\mathcal{Q}$ & A set of questions, where each $q\in\mathcal{Q}$ is transformed from a $p$. \\
    $C$ & An RPA of a character $c$. \\

    \midrule
    \rowcolor[rgb]{ .949,  .953,  .961} \multicolumn{2}{c}{\textit{Methods}} \\
    SR & Self-report baseline. \\
    SR-CoT & SR with chain-of-thought reasoning. \\
    ER & \method with expert rating. \\
    ER$_\texttt{batch}$ &  ER where question-response pairs in one dimension are inputted in-batch. \\
    ER$_\texttt{all}$ &  ER where question-response pairs in one dimension are inputted all-at-once. \\
    OC & \method with option conversion. \\
    d-OC & OC with dimension-specific options. \\ 
    
    \midrule
    \rowcolor[rgb]{ .949,  .953,  .961} \multicolumn{2}{c}{\textit{Metrics}} \\
    \textbf{MAE} & Mean absolute error. \\
    \textbf{Acc$_\texttt{Dim}$} & Accuracy on individual dimensions. \\
    \textbf{Acc$_\texttt{Full}$} & Accuracy on all dimensions of a scale. \\
    \textbf{Std$_\texttt{Item}$} & Standard variance of an RPA's responses (converted to scores) on the same item across multiple runs. \\
    \textbf{Std$_\texttt{Dim}$} & Standard variance of an RPA's responses (converted to scores) acorss different items in the same dimension. \\
    \textbf{Std$_\texttt{Item}$} & Standard variance of an RPA's score on each dimension across multiple runs. \\

    \bottomrule
    \end{tabular}
  \caption{
  A notation table. 
  }
  \label{tab:notations}
\end{table}

In Table~\ref{tab:notations}, we list the notations and abbreviations in this paper, along with their definitions.

\section{Psychological Scales}
\label{sec:scales}

\paragraph{Big Five Inventory}
The BFI serves as a prominent instrument for assessing personality dimensions.
This model, often encapsulated by the acronym ``OCEAN,'' encompasses five critical traits:
(1) \textit{Openness to Experience (O)}, which highlights a person's curiosity, inventiveness, and appreciation for art, emotion, adventure, and novel concepts.
(2) \textit{Conscientiousness (C)}, indicating how much an individual exhibits organization, reliability, and responsibility.
(3) \textit{Extraversion (E)}, denoting the level to which a person is sociable and energized by interactions with others.
(4) \textit{Agreeableness (A)}, assessing an individual's kindness, empathy, and ability to cooperate with others.
(5) \textit{Neuroticism (N)}, gauging the tendency of an individual to experience negative feelings such as anxiety, anger, and sadness, as opposed to being more emotionally resilient and less stress-susceptible.

\paragraph{Eysenck Personality Questionnaire (Revised)}
The Revised Eysenck Personality Questionnaire (EPQ-R) serves as a psychological instrument for gauging distinct personality trait variances in individuals.
It identifies three principal traits:
(1) \textit{Extraversion (E)}, which assesses whether a person tends to be more sociable, energetic, and outgoing as opposed to being introverted, quiet, and reserved.
(2) \textit{Neuroticism (N)}, which gauges emotional steadiness.
These dimensions (\ie, E and N) share similarities with those found in the BFI.
(3) \textit{Psychoticism (P)}, which is indicative of a person's inclination towards solitude, a lack of empathy, and a propensity for aggression or a tough-minded attitude.
This trait is crucial to understand as indicative of personality characteristics rather than serious mental health conditions.
(4) Beyond these primary scales, the EPQ-R also incorporates a \textit{Lying Scale (L)} intended to identify responses aimed at social desirability.
This scale evaluates the extent to which an individual may attempt to portray themselves in a more favorable light.

\paragraph{Dark Triad Dirty Dozen}
The DTDD is identified as a brief, 12-item measure crafted to evaluate the trio of principal personality characteristics known as the Dark Triad, encompassing:
(1) \textit{Narcissism (N)}, characterized by an exaggerated sense of one's own significance, an obsession with dreams of boundless success, and a craving for undue admiration.
(2) \textit{Machiavellianism (M)}, indicative of a deceitful approach in social interactions and a skeptical indifference to ethical principles.
(3) \textit{Psychopathy (P)}, which includes tendencies towards impulsiveness, a deficiency in empathy, and hostile relations with others.
These Dark Triad personality dimensions are typically viewed as the antithesis of the characteristics measured by the BFI or the EPQ-R, which represent ``Light'' traits.

\paragraph{The NERIS Type Explorer}
The 16Personalities utilizes the acronym format introduced by Myers-Briggs for its simplicity and convenience, with an additional letter to accommodate five rather than four scales.
However, unlike Myers-Briggs or other theories based on the Jungian model, the incorporation of Jungian concepts such as cognitive functions, or their prioritization, has not been undertaken.
Instead, they rework and rebalance the dimensions of personality in the BFI personality traits.
The personality types are based on five independent spectrums, with all letters in the type code (\eg, INFJ-A) referring to one of the two sides of the corresponding spectrum.

\paragraph{Bem's Sex Role Inventory}
The BSRI assesses the degree to which individuals identify with traditionally masculine and feminine characteristics.
Rather than focusing on behaviors, such as participation in sports or cooking, this tool evaluates psychological characteristics, including assertiveness and gentleness.
Participants are divided into four groups based on whether their average scores exceed the median for each component.
These groups are designated as \textit{Masculine} (M: Yes; F: No), \textit{Feminine} (M: No; F: Yes), \textit{Androgynous} (M: Yes; F: Yes), and \textit{Undifferentiated} (M: No; F: No).

\paragraph{Comprehensive Assessment of Basic Interests}
The CABIN provides an exhaustive evaluation for identifying 41 essential dimensions of vocational interest.
Following this evaluation, the researchers introduce a model of interest consisting of eight dimensions, named \textit{SETPOINT}.
This model includes dimensions such as Health \underline{S}cience, Creative \underline{E}xpression, \underline{T}echnology, \underline{P}eople, \underline{O}rganization, \underline{I}nfluence, \underline{N}ature, and \underline{T}hings.
These core dimensions are also adaptable to a six-dimension framework, which is prevalently recognized within the interest research community.
This framework aligns with Holland's \textit{RIASEC} model, which features the dimensions: \underline{R}ealistic, \underline{I}nvestigate, \underline{A}rtistic, \underline{S}ocial, \underline{E}nterprising, and \underline{C}onventional.

\paragraph{Implicit Culture Belief}
The ICB scale measures the extent to which individuals think a person's ethnic culture influences their development.
Scoring higher on this scale indicates a firm belief that a person's ethnic culture is the main factor shaping their identity, values, and perspective on the world.
On the other hand, a lower score on the scale denotes a belief in the ability of an individual to shape their own identity through hard work, commitment, and education.

\paragraph{Experiences in Close Relationships (Revised)}
The ECR-R is a self-assessment tool crafted to gauge variations in adult attachment styles, particularly within the realm of romantic relationships.
As an enhanced iteration of the original ECR scale, the ECR-R introduces refinements in quantifying attachment tendencies.
It assesses two primary aspects:
(1) \textit{Attachment Anxiety} indicates the degree to which a person fears rejection or abandonment by their romantic partners.
(2) \textit{Attachment Avoidance} assesses the degree to which a person prefers to keep emotional and physical distance from their partners, often stemming from unease with closeness or reliance.

\paragraph{General Self-Efficacy}
The GSE Scale evaluates a person's confidence in their capacity to address diverse demanding situations in life.
This confidence, known as ``self-efficacy,'' plays a pivotal role in social cognitive theory and is associated with numerous health outcomes, motivational levels, and performance measures.
An elevated score on this scale indicates a person's strong belief in their ability to confront and manage challenging circumstances, undertake new or complex tasks, and navigate through the resultant difficulties.
On the flip side, a lower score on the scale suggests a lack of self-assurance in handling challenges, rendering individuals more susceptible to experiencing helplessness, anxiety, or engaging in avoidance behaviors when encountering hardships.

\paragraph{Life Orientation Test (Revised)}
The LOT-R is designed to assess variations in optimism and pessimism among individuals.
It includes ten questions, with an interesting aspect being that only six of these questions contribute to the test's score.
The other four are designed as filler items, cleverly integrated to obscure the test's primary focus.
Within the scored questions, equal numbers are dedicated to evaluating optimism and pessimism—three for each.
A tendency towards higher scores in optimism and lower in pessimism signifies a predominantly optimistic outlook.

\paragraph{Love of Money Scale}
The LMS evaluates the perspectives and feelings of people regarding money.
This tool aims to quantify the degree to which people perceive money as a symbol of power, success, and liberty, along with its significance in influencing behaviors and choices.
The LMS identifies three key dimensions:
(1) \textit{Rich} reflects the degree to which people link money with success and accomplishment.
(2) \textit{Motivator} determines the extent to which money serves as an incentive in someone's life, \ie, how much individuals are motivated by monetary rewards in their decisions and behaviors.
(3) \textit{Important} assesses the level of importance people attribute to money, affecting their principles, objectives, and perspective of the world.

\paragraph{Emotional Intelligence Scale}
The EIS serves as a self-assessment tool for evaluating multiple aspects of emotional intelligence.
This instrument emphasizes various elements of emotional intelligence, notably the perception, management, and application of emotions.
It is extensively utilized in the field of psychology to investigate how emotional intelligence influences different outcomes, including personal well-being, professional performance, and social interactions.

\paragraph{Wong and Law Emotional Intelligence Scale}
Similar to EIS, the WLEIS is also a self-report instrument designed for evaluating emotional intelligence.
However, it distinctly includes four subscales that represent the primary aspects of emotional intelligence:
(1) \textit{Self-emotion appraisal (SEA)} focuses on an individual's proficiency in identifying and understanding their emotions.
(2) \textit{Others' emotion appraisal (OEA)} is about the skill of recognizing and comprehending the emotions of others.
(3) \textit{Use of emotion (UOE)} deals with the ability to employ emotions to aid various mental processes, like reasoning and problem-solving.
(4) \textit{Regulation of emotion (ROE)} is concerned with the ability to control and adjust emotions within oneself and in others.

\paragraph{Empathy Scale}
Empathy, defined as the capacity to perceive and resonate with the emotions of another, is traditionally divided into cognitive and emotional empathy.
Cognitive empathy, also known as ``perspective-taking,'' entails the mental faculty to identify and comprehend the thoughts, beliefs, or feelings of someone else.
Conversely, emotional empathy involves the vicarious experience of the emotions felt by another individual.

\clearpage

\section{Character Selection}
\label{sec:character_selection}

\begin{table*}[ht]
\centering
\resizebox{\linewidth}{!}{
\begin{tabular}{l|p{0.9\linewidth}}
\toprule
\multicolumn{2}{c}{\textbf{System Prompts for RPAs}} \\
\midrule
    \textbf{Zhong Li} (ChatHaruhi) &
    Please be aware that your codename in this conversation is 'Zhongli'.
    
    Others call you 'Zhongli', 'Guest', 'Emperor', or 'Rex Lapis'.
    
    The preceding text provided some classic scenes from the game.
    
    If I ask a question that closely resembles a line from the game, please cooperate with me in acting it out.
    
    If I ask a question related to events in the game, please respond based on the game's content.
    
    If I ask a question beyond the scope of the game, respond in the style of Zhongli.
    
    You have the appearance of an adult male, with a short haircut and a long, thin braid reaching down to your hips. Your hair transitions from brown at the roots to orange-yellow at the tips, and the braid is adorned with diamond-shaped decorations in bright yellow with pale yellow borders.
    Your eyes are golden with diamond-shaped pupils, and there is orange-yellow eyeshadow from the corners to the sides.
You wear a single earring resembling a floral bell-shaped tassel on your left ear. Your attire is a blend of a suit and a robe, divided into three layers.  ... (695 words)
    
    \\ \midrule
     \textbf{Hermione Granger} (ChatHaruhi) & 
    I want you to act like Hermione from Harry Potter.
    You are now cosplay Hermione Granger
    If others‘ questions are related with the novel, please try to reuse the original lines from the novel.
    I want you to respond and answer like Hermione using the tone, manner and vocabulary Hermione would use. 
    You must know all of the knowledge of Hermione. 
    
    Hermione Granger is a smart, diligent, and confident young witch with a high pursuit of learning and knowledge. She has a broad knowledge of magic and often provides important information. Hermione's conversations frequently involve facts and logical reasoning, and she is good at raising questions and solving problems. (109 words)
    
    \\ \midrule

     \textbf{Thor} (RoleLLM) & 
    I want you to act like Thor from Thor-Ragnarok
    
    If others‘ questions are related with the novel, please try to reuse the original lines from the novel.
    
    I want you to respond and answer like Thor using the tone, manner and vocabulary Thor would use.
    
    You must know all of the knowledge of Thor.
    
    You are a powerful and godlike being, the crown prince of Asgard who wields a mighty hammer. Initially arrogant and impulsive, you undergo a transformative journey, learning humility and becoming a true hero. Throughout the series, you face numerous challenges and battles, including a rivalry with your adoptive brother and the threat of a powerful villain. Your story is filled with epic battles, personal growth, and ultimately, the redemption of a fallen hero.  (127 words) \\

    \midrule
    \textbf{James Bond} (RoleLLM) & 
    I want you to act like James Bond from Tomorrow-Never-Dies.
    
    If others‘ questions are related with the novel, please try to reuse the original lines from the novel.
    
    I want you to respond and answer like James Bond using the tone, manner and vocabulary James Bond would use.
    
    You must know all of the knowledge of James Bond.
    
    A suave and skilled British secret agent with a license to kill, you are known for your impeccable style, charm, and wit. With a troubled past as an orphan, you have honed your skills in espionage and combat, making you a formidable adversary. Throughout the series, you undergo personal growth, evolving from a womanizer to a more complex and introspective individual. You embark on dangerous missions around the world, often facing off against iconic villains and saving the world from various threats. Your important events include your numerous romantic encounters, the loss of loved ones, and your constant battle against global terrorism. (160 words) \\
\bottomrule

\end{tabular}}
\caption{Examples of system prompts for RPAs in ChatHaruhi and RoleLLM. The prompt for Zhong Li is originally in Chinese and translated into English.}
\label{tab:character_prompts}
\end{table*}

When selecting characters for RPAs, we consider the following factors:
(1) There exist multiple RPAs for the characters, \eg, both ChatHaruhi and character.ai have their RPA for \textit{Hermione Granger}.
(2) The personality data of these characters on the BFI and 16P should be available and widely annotated on the PDb.
(3) The selected characters should possess diversified personalities.
Hence, the pipeline of our character selection process is composed of the following steps:
\begin{enumerate}[leftmargin=*]
    \item Initially,  we collect characters with RPAs and character data curated by ChatHaruhi and RoleLLM. 
    Then, we search for their counterparts in character.ai, and keep only those with character.ai RPAs. 
    \item We collect personality data for these characters from the PDb. Characters with less than ten annotations on either the BFI or the 16P are discarded. 
    \item We categorize the remaining characters based on their BFI personality types (5-letter code such as \textit{SLOAI}). Then, we select characters in turn from each type, over multiple rounds, to form an ordered list of candidate characters. If a certain type has no characters left, we skip it.
    \item Finally, We pick the first 32 characters in the candidate list. Then, we manually check whether the ChatHaruhi/RoleLLM RPA, the character.ai RPA, and the PDb annotation refer to the same character. If not, the character is removed and we select the next candidate character. 
\end{enumerate}

The selected characters and their sources are: 
\textit{Hermione Granger, Harry Potter, Ron Weasley, Luna Lovegood, Draco Malfoy, Albus Dumbledore, Minerva McGonagall, Severus Snape (Harry Potter Series), 
Zhong Li, Hu Tao, Raiden Shougun, Ayaka Kamisato, Wanderer (Genshin Impact), 
Thor, Lucifer Morningstar, Rorschach (DC Comics)
Sheldon Cooper, Raj Koothrappali (The Big Bang Theory), 
Gaston (Beauty and the Beast),
Klaus Mikaelson (The Vampirie Diaries),
Jigsaw (Saw Series),
James Bond (James Bond Film Series),
Twilight Sparkle (My Little Pony: Friendship Is Magic),
John Keating (Dead Poets Society),
Michael Scott (The Office),
Shrek (Shrek),
Jeffrey Lebowski (The Dude),
Walk Kowalski (Gran Torino), 
Lestat de Lioncourt (Interview with the Vampire), 
Blair Waldorf (Gossip Girl), 
Haruhi Suzumiya (The Melancholy of Haruhi Suzumiya), and 
Jim Morrison (Celebrities).} 

We illustrate these characters' personality scores on the BFI, together with the measured personalities of their RPAs in Figure~\ref{fig:bfi_radar}. 
Besides, we demonstrate some examples of system prompts of the RPAs in Table~\ref{tab:character_prompts}.

\clearpage

\section{Human Annotations}
\label{sec:annotation}

We collect groundtruth labels for character personalities annotated by people familiar with the characters, from both the PDb and our invited annotators.  

\paragraph{The Personality Database} 
PDb collects and offers categorical personality annotations of massive fictional characters on the BFI (\eg, \dq{RCUAI}\footnote{Similar to the 16P, PDb adopts a coding mechanism named \textit{SLOAN} for the BFI, which describes a character on each dimension with a letter denoting \textit{high} or \textit{low} scorers.})
and 16P (\eg, \dq{ENTJ}).
Each character $c$ is labeled by plentiful human annotators familiar with $c$, and PDb offers detailed numbers of annotations of each label.
For the selected characters, 
we calculate the label-percentage of the positive type on each dimension\footnote{For the BFI, the positive types are the types of high scorers. 
For the 16P, we manually designate positive and negative types on each dimension, \eg, \dq{E} as positive and \dq{I} as negative for the \dq{E/I} dimension.}
as the \textit{scores}. 
Then, we categorize each score into the positive, negative or marginal \textit{type} if it is above $60\%$, under $40\%$ or otherwise. 
Marginal types indicate ambiguity and would be ignored for alignment calculation.

\begin{table}[h]
    \centering
    \resizebox{\columnwidth}{!}{
    \begin{tabular}{p{1.8cm}p{8.8cm}}
    \toprule
    \rowcolor{mygray}
    \multicolumn{2}{l}{\textbf{Human Annotation Example Prompt}} \\ \midrule
    \textsc{Inst} & Please rate \texttt{<character>} on the \dq{Openness} dimension of BFI personality.
    \\ \midrule
    \textsc{Scoring Scale} & Each option is a number from 1 to 5. 1 represents \dq{very uncurious}, 2 represents \dq{uncurious}, 3 represents \dq{neutral}, 4 represents \dq{curious}, and 5 represents \dq{very curious.} \\
    & If you think this dimension is irrelevant to the character, or it is difficult to judge the character  on this dimension, please answer \dq{X.}
    \\ \midrule
    \textsc{Examplar} & 1. Is original, comes up with new ideas. (High score)\\
    \textsc{Items} & 2. Is curious about many different things. (High score)\\
    & 3. Is ingenious, a deep thinker. (High Score)\\
    & 4. Prefers work that is routine.. (Low score)\\
    & 5. Has few artistic interests.. (Low score)\\
    \bottomrule
    \end{tabular}}
\end{table}

\paragraph{Human Annotations}
After that, we invite human annotators familiar with these characters to score them on a broader range of psychological scales.
Each annotation contains 73 dimensions on 14 scales for a character. 
We provide detailed annotation prompts of each dimension, including the scoring instruction and examplar positively and negatively related items.

\setlength\tabcolsep{3pt}
\begin{table}[t]
\small
  \centering
    \begin{tabular}{ccccccc}
    \toprule
     \textbf{BFI} & \textbf{16P} & \textbf{BSRI} & \textbf{Empt.} & \textbf{EPQ-R} & \textbf{LMS} & \textbf{DTDD} \\
    \midrule
     75.9 & 77.0 & 71.4 & 64.5 & 43.8 & 65.6 & 51.7 \\
    \toprule
    
     \textbf{ECR-R} & \textbf{GSE} & \textbf{ICB} & \textbf{LOT-R} & \textbf{EIS} & \textbf{WLEIS} & \textbf{CABIN} \\
    \midrule
     47.3 & 64.8 & 62.2 & 71.9 & 50.9 & 58.8 & 46.3  \\

    \bottomrule
    \end{tabular}%}
  \caption{
  The kappa coefficient (\%) on 14 scales. We adopt Cohen’s quadratic-weighted kappa
  considering the ordinal nature of personality labels. 
}
  \label{tab:kappa}
\end{table}

We first examine their understanding of the characters by matching their type annotations on the BFI and 16P with the PDb labels. 
If one or two differences exist, we ask them to explain their answers.
If more differences exist, or the annotators admit a lack of character understanding, the process stops and we invite new annotators.

Then, we ask them to score the characters on the 73 dimensions  according to corresponding scoring ranges and instructions.
Annotators are allowed to mark a character on a dimension as \dq{X} which indicates unrelatedness or ambiguity.
We collected two to three annotations for each character (in total, 93 annotations for 32 characters) and took their average scores as the final score. 
Finally, we linearly re-scale the scores into the unit interval $[0, 1]$,
and similarly categorize the scores into \textit{types}. 
If more than one annotators mark a dimension as \dq{X} for a character, this dimension will also be treated as a marginal type.

\begin{table}[t]
\small
  \centering
    \begin{tabular}{lggggvvvv}
    \toprule
     \multirow{3}{*}{\textbf{Method}} & \multicolumn{4}{c}{\textbf{PDb Labels}} &\multicolumn{4}{c}{\textbf{Annotator Labels}} \\ 
     \cmidrule(lr){2-5} \cmidrule(lr){6-9} 
     & \multicolumn{2}{c}{\textbf{BFI}} & \multicolumn{2}{c}{\textbf{16P}} & \multicolumn{2}{c}{\textbf{BFI}} & \multicolumn{2}{c}{\textbf{16P}} \\
     \cmidrule(lr){2-3} \cmidrule(lr){4-5} \cmidrule(lr){6-7} \cmidrule(lr){8-9}  
     & \multicolumn{1}{c}{\textbf{Acc}} & \multicolumn{1}{c}{\textbf{MSE}} & \multicolumn{1}{c}{\textbf{Acc}} & \multicolumn{1}{c}{\textbf{MSE}} & \multicolumn{1}{c}{\textbf{Acc}} & \multicolumn{1}{c}{\textbf{MSE}} & \multicolumn{1}{c}{\textbf{Acc}} & \multicolumn{1}{c}{\textbf{MSE}} \\ 
     \midrule

    SR &  63.3 & 15.8 & 65.6 & 15.2 
        &  60.1& 7.6 & 59.2 & 9.1  \\ 

    SR-CoT & 65.0  & 16.9 & 69.7 & 14.7 
        & 60.7  & 7.9 & 65.0 & 8.6  \\ 

    \cdashlinelr{1-9}
    OC &  64.3 & 16.1 & 75.5 & 12.7 
        & 63.1 & 7.2 & 69.4 & 7.2  \\ 
    
    d-OC & 72.2  & 13.1 & 80.2 & 10.1 
        & 65.6 & 6.4 & 72.5 & 6.7  \\ 
        
    \cdashlinelr{1-9}
    ER$_{\texttt{All}}$ & 76.7  & 12.4 & 79.6 & 9.3 
        & 67.9 & 6.5 & 72.7 & 6.8  \\ 
        
    ER$_{\texttt{Batch}}$ & 76.7  &  12.2 & 80.7 & 9.7 
        & 67.5  & 5.9 & 73.6 & 6.4  \\

    \bottomrule
    \end{tabular}%}
  \caption{
  Measured alignment (\%) on the BFI and 16P with labels from the PDb or invited annotators, with \gptfour as the interviewer LLM. 
  \textbf{Acc} denotes dimensional accuracy. 
}
  \label{tab:two_labels}
\end{table}

We measure the inter-annotator consistency via Cohen's kappa coefficient~\citep{Cohen1968} on each of the scales. 
Specifically, we calculate quadratic-weighted kappa~\citep{Cohen1968}, since the classes are ordinal. 
The average kappa coefficient across 14 scales is $60.9\%$. 
The detailed kappa coefficients on 14 scales are listed in Table ~\ref{tab:kappa}.
We find that the coefficients on the BFI and 16P are higher, because the annotators are more familiar with dimensions in the BFI and 16P, and the authors have made some discussion with them on these scales. 
The coefficients on EPQ-R, ECR-R and CABIN are lower than those of the other scales. 
For ECR-R and CABIN, this is probably because the characters' personalities on close relationships (ECR-R) and various career interests (CABIN) are rarely depicted in the original works. 
For EPQ-R, this is in part influenced by ambiguity of the \textit{Lying} dimension, where high scores actually stand for \textit{Honesty}.

On the BFI and 16P, we experimented with both labels from the PDb and our invited annotators.
Experimental results in Table ~\ref{tab:two_labels} show that, across various settings, the PDb labels always yield higher accuracy while labels from invited annotators yield better MSE.
The reasons are that, the PDb labels are contributed by massive online annotators, and are hence more accurate. 
However, they offer only the personality types and corresponding label numbers instead of detailed personality scores, which cannot be well represented by the label-percentage. 
For example, even if all the annotators mark a character $c$ as \textit{Extravorted}, we cannot assume $c$ as completely  \textit{Extroverted}.

\newpage

\section{Implementation Details}

\subsection{RPA Inference and Post-processing}

Our implementation of RPAs is based on Chat-Haruhi-Suzumiya  
and LangChain\footnote{\url{https://www.langchain.com/}}. 
Hence, we invoke foundation LLMs in RPAs with the default temperature 0.7 of LangChain. 
When experimenting with open-source LLMs as foundation models, we observe that LLMs may generate unexpected multi-round conversations or repeated content. 
In such cases, we remove the extra rounds or repetition.

In self-report tests, RPAs may refuse to participate in the tests and provide their choices, interestingly, because they are role-playing characters with noncompliant personalities.
In these cases, the responses are categorized as \dq{Neutral} for completeness of the results.

For interviewer LLMs, we request responses in JSON formats. 
If a response cannot be parsed into JSON data or the results of any samples in the batch are missing, we prompt interviewer LLMs to regenerate. 
The temperature is set as 0 for the initial generation, and 0.2 for regeneration. 
For \chatgpt, if samples in one input exceed the token limit, we split the input into smaller batches.  
For \gemini, there are several cases where responses repeatedly fail to be parsed into JSON format or are blocked, 
In such cases, we resort to \gptfour instead to ensure the completeness of the assessment process.

\subsection{Fine-Tuning}
\label{sec:finetuning}

We fine-tune the Mistral-2 7B model on the \textit{ChatHaruhi-English-62K} dataset\footnote{\url{https://huggingface.co/datasets/silk-road/ChatHaruhi-English-62K-RolePlaying}}, sourced from both ChatHaruhi~\citep{li2023chatharuhi} and RoleLLM~\citep{wang2023rolellm}. 
Our implementation is based on LLaMA-Factory\footnote{\url{https://github.com/hiyouga/LLaMA-Factory}}. 
We adopt LoRA tuning, and configure the training with a batch size of $16$, a learning rate of $5e-5$, across three epochs, using the \dq{fp16} option.

\clearpage

\section{Additional Results}

\subsection{Interviewer LLMs}
\label{sec:interviewer_llms}

For the OC and ER tasks, we report the detailed numbers of \textit{right}, \textit{close} and \textit{wrong} cases of interviewer LLMs in Table~\ref{tab:human_eval_count}.
For ER, \gptfour rates 82\% of cases highly consistent with  humans, and only 4\% of cases notably different compared with humans. 
Hence, state-of-the-art LLMs are capable of rating RPA personalities based on their interview results. 

\subsection{Self-report v.s. Interview-based Methods}
\label{sec:sr_vs_ib}

The advantages of \method over SR originate from the fact that \method elicits the thoughts and behaviors of RPAs for personality assessments.
In contrast, SR directly prompts RPAs to provide the choices, which may be easily biased by pre-training data of the foundation models.  
This difference enables \method to yield more distinct personality measurements than self-report, as is evidenced in Figure~\ref{fig:distribution}. 
For further validation, we compute their standard variance of measured personality scores of the 32 RPAs on the BFI, and average the results over the five dimensions.
We obtain a result of $1.03$ for \method with ER$_\texttt{batch}$ and \gptfour, $0.71$ for SR and $0.68$ for SR-CoT.

While SR-CoT also attempts to elicit thoughts of the RPAs, its improvement over SR is limited. 
Furthermore, SR-CoT has two disadvantages compared with \method. 
First, SR-CoT indeed requires RPAs themselves to perform the OC task implicitly,  
while \method can apply other methodologies such as d-OC and ER. 
As is shown in Table~\ref{tab:human_eval}, existing LLMs generally perform better on ER and d-OC, rather than OC.
Second, the implicit OC task in SR-CoT would be a challenge for RPAs based on small foundation models, while in \method, we can decouple the interviewer LLMs and foundation models for RPAs.

\setlength\tabcolsep{3pt}
\begin{table}[t]
\small
  \centering
    \begin{tabular}{lccc}
    \toprule
    \textbf{LLM} & \textbf{\#Right} & \textbf{\#Close} & \textbf{\#Wrong}  \\
    \midrule
    \rowcolor[rgb]{ .949,  .953,  .961} \multicolumn{4}{c}{\textit{Option Conversation}} \\
        
     \chatgpt & $45$ & $25$ & $30$  \\
     \gptfour & $57$ & $28$ & $15$  \\
     \gemini & $54$ & $31$ & $15$  \\
    \midrule

    \rowcolor[rgb]{ .949,  .953,  .961} \multicolumn{4}{c}
    {\textit{Dimension-specific Option Conversation}} \\
        
     \chatgpt & $64$ & $25$ & $11$   \\
     \gptfour & $70$ & $24$ & $6$   \\
     \gemini & $68$ & $22$ & $10$  \\
    \midrule

    \rowcolor[rgb]{ .949,  .953,  .961} \multicolumn{4}{c}
    {\textit{Expert Rating (batch)}} \\
        
     \chatgpt & $71$ & $26$ & $3$  \\
     \gptfour & $82$ & $14$ & $4$   \\
     \gemini & $79$ & $10$ & $11$   \\

    \bottomrule
    \end{tabular}
  \caption{
  Detailed numbers of right, close, and wrong cases in human evaluation of interviewer LLMs 
 on the option conversion task and the expert rating tasks. 
  \#Right, \#Close, and \#Wrong denote the number of LLM predictions that differ from human annotations by less than 1 point, exactly 1, or more than 1. 
    }
  \label{tab:human_eval_count}
\end{table}

\subsection{RPAs from Different Works}
\label{sec:rpas_different_works}

\setlength\tabcolsep{2.8pt}
\begin{table*}[htbp]
\small
  \centering
    \resizebox{\textwidth}{!}{
    \begin{tabular}{lcccccccccccccccccc}
    \toprule
    \multirow{3}{*}{\textbf{RPA Type}} & \multicolumn{6}{c}{\textbf{CH}} &  \multicolumn{6}{c}{\textbf{RL}} & \multicolumn{6}{c}{\textbf{CH+RL}} \\
    \cmidrule(lr){2-7} \cmidrule(lr){8-13}  \cmidrule(lr){14-19}
    & \multicolumn{3}{c}{\textbf{BFI}} & \multicolumn{3}{c}{\textbf{16P}} & \multicolumn{3}{c}{\textbf{BFI}} & \multicolumn{3}{c}{\textbf{16P}} & \multicolumn{3}{c}{\textbf{BFI}} & \multicolumn{3}{c}{\textbf{16P}} \\
    \cmidrule(lr){2-4} \cmidrule(lr){5-7} 
    \cmidrule(lr){8-10} \cmidrule(lr){11-13}
    \cmidrule(lr){14-16} \cmidrule(lr){17-19}
    & \multicolumn{1}{c}{\textbf{\textbf{Acc$_\texttt{Dim}$}}} & \multicolumn{1}{c}{\textbf{\textbf{Acc$_\texttt{Full}$}}} & \multicolumn{1}{c}{\textbf{MAE} $\downarrow$} & \multicolumn{1}{c}{\textbf{\textbf{Acc$_\texttt{Dim}$}}} & \multicolumn{1}{c}{\textbf{\textbf{Acc$_\texttt{Full}$}}} & \multicolumn{1}{c}{\textbf{MAE} $\downarrow$} & \multicolumn{1}{c}{\textbf{\textbf{Acc$_\texttt{Dim}$}}} & \multicolumn{1}{c}{\textbf{\textbf{Acc$_\texttt{Full}$}}} & \multicolumn{1}{c}{\textbf{MAE} $\downarrow$} & \multicolumn{1}{c}{\textbf{\textbf{Acc$_\texttt{Dim}$}}} & \multicolumn{1}{c}{\textbf{\textbf{Acc$_\texttt{Full}$}}} & \multicolumn{1}{c}{\textbf{MAE} $\downarrow$} & \multicolumn{1}{c}{\textbf{\textbf{Acc$_\texttt{Dim}$}}} & \multicolumn{1}{c}{\textbf{\textbf{Acc$_\texttt{Full}$}}} & \multicolumn{1}{c}{\textbf{MAE} $\downarrow$} & \multicolumn{1}{c}{\textbf{\textbf{Acc$_\texttt{Dim}$}}} & \multicolumn{1}{c}{\textbf{\textbf{Acc$_\texttt{Full}$}}} & \multicolumn{1}{c}{\textbf{MAE} $\downarrow$}  \\
    \midrule
    
    \rowcolor[rgb]{ .949,  .953,  .961} \multicolumn{19}{c}{\textit{Close-source RPAs}} \\
    character.ai(D*) 
    & 58.8 & 12.5 & 37.0 & 67.2 & 31.3 & 35.6
    & 45.5 & 6.3 & 46.6 & 38.3 & 12.5 & 45.6
    & 52.2 & 9.4 & 41.7 & 52.3 & 21.9 & 40.6 \\ 

    \rowcolor[rgb]{ .949,  .953,  .961} \multicolumn{19}{c}{\textit{Open-source RPAs}} \\
        
     \chatgpt (D) 
        & 71.7 & 18.8 & \textbf{29.0}
        & 77.6 & \textbf{39.6} & \textbf{28.5}
        & 71.0 & \textbf{22.9} & 30.8
        & \textbf{78.9} & \textbf{54.2} & \textbf{23.9}
        & 71.3 & 20.8 & 29.9
        & 78.2 & \textbf{46.9} & \textbf{26.2}
        \\ 
        
   \chatgpt (M) 
        & \textbf{74.2} & 20.8 & 31.9
        & 72.7 & 43.8 & 33.7
        & 69.7 & 20.8 & 34.7
        & 72.2 & 29.2 & 33.3
        & 72.0 & 20.8 & 33.3
        & 72.5 & 36.5 & 33.5
        \\ 

    \chatgpt (D+M) 
        & 72.9 & \textbf{22.9} & 30.0
        & \textbf{78.2} & 37.5 & 29.5
        & \textbf{71.9} & 22.9 & \textbf{27.7}
        & \textbf{78.9} & 50.0 & 25.4
        & \textbf{72.4} & \textbf{22.9} & \textbf{28.8}
        & \textbf{78.5} & 43.8 & 27.4
        \\

    \bottomrule
    \end{tabular}}
  \caption{
  Measured alignment (\%) on the BFI and 16P, in different characters groups and  RPA types. 
  The results are measured via \method with ER$_\texttt{batch}$, using \chatgpt. 
  \textbf{CH} and \textbf{RL} correspond to the selected characters from ChatHaruhi and RoleLLM. 
  D and M represent descriptions and memories respectively, contributed by ChatHaruhi and RoleLLM on corresponding characters. 
  D* denote private descriptions of \texttt{character.ai}.
  }
  \label{tab:test_rpa_types_full}
\end{table*}

Many recent efforts have been committed to developing RPAs for specific characters in various ways. 
We compare RPAs contributed by  ChatHaruhi~\citep{li2023chatharuhi}, RoleLLM~\citep{wang2023rolellm} and character.ai\footnote{Our implementation is based on \url{https://github.com/kramcat/CharacterAI}.}.
They craft character data in different methods.
ChatHaruhi and RoleLLM share few characters in common, and their character data are thus less comparable.
character.ai covers all our selected characters from ChatHaruhi and RoleLLM, but is tied to its close-source model.
Hence, we distinguish three groups of characters, including the 16 ChatHaruhi characters (\textbf{CH}, the 16 RoleLLM characters  (\textbf{RL}) and their union (\textbf{CH+RL}). 
We implement 
\textbf{CH} and \textbf{RL} RPAs with \chatgpt as the foundation model using character data from ChatHaruhi and RoleLLM.
We conduct personality tests using \method, with ER$_\texttt{batch}$ and \chatgpt as the interviewer LLM.

The results are shown in Table~\ref{tab:test_rpa_types_full}. 
According to the results, we observe that \chatgpt-based RPAs with character data from both ChatHaruhi and RoleLLM achieve high personality fidelity, demonstrating the quality of their character data. 
However, character.ai RPAs exhibit low personality fidelity regarding the characters, even if we compare it with \chatgpt-based RPAs using only character descriptions from ChatHaruhi and RoleLLM.

\subsection{Foundation Models for RPAs}
\label{sec:found_models}

\begin{table}[t]
\small
  \centering
    \resizebox{\columnwidth}{!}{
    \begin{tabular}{lcccccccc}
    \toprule
    \multirow{3}{*}{\parbox{2cm}{\textbf{Foundation Models}}} & \multicolumn{4}{c}{\textbf{English}} &  \multicolumn{4}{c}{\textbf{Chinese}} \\
    \cmidrule(lr){2-5} \cmidrule(lr){6-9} 
    & \multicolumn{2}{c}{\textbf{BFI}} & \multicolumn{2}{c}{\textbf{16P}} & \multicolumn{2}{c}{\textbf{BFI}} & \multicolumn{2}{c}{\textbf{16P}}  \\
    \cmidrule(lr){2-3} \cmidrule(lr){4-5} 
    \cmidrule(lr){6-7} \cmidrule(lr){8-9} 
    &
    \multicolumn{1}{c}{\textbf{Acc}} & \multicolumn{1}{c}{\textbf{MAE} } &
    \multicolumn{1}{c}{\textbf{Acc}} & \multicolumn{1}{c}{\textbf{MAE} } &
    \multicolumn{1}{c}{\textbf{Acc}} & \multicolumn{1}{c}{\textbf{MAE} } &
    \multicolumn{1}{c}{\textbf{Acc}} & \multicolumn{1}{c}{\textbf{MAE} } 

    \\ \midrule

    \rowcolor[rgb]{ .949,  .953,  .961} \multicolumn{9}{c}{\textit{General Open-source Models}} \\

    Qwen 7B
    & 60.6 & 36.9 & 67.7 & 34.8
    & 60.0 & 34.9 & 68.2 & 34.6
    \\ 
    
    OpenChat-3.5 7B
    & 60.6 & 35.2 & 77.8 & 29.6
    & 73.3 & 32.6 & 72.7 & 34.6
    \\ 
    
    Mistral-2 7B
    & 66.1 & 33.0 & 71.7 & 30.7
    & 66.7 & 31.3 & 54.6 & 36.6
    \\ 

    LLaMa-2-Chat 13B
    & 69.3 & 32.3 & 69.7 & 31.5
    & 56.7 & 40.9 & 54.6 & 37.8
    \\ 
    
    Mixtral 8x7B
    & 66.9 & 31.1 & 71.7 & 32.5
    & 73.3 & 32.1 & 72.7 & 36.8
    \\ \midrule

    \rowcolor[rgb]{ .949,  .953,  .961} \multicolumn{9}{c}{\textit{Open-source Models Fine-tuned for RPAs}} \\

    CharacterGLM 6B
    & 54.3 & 39.7 & 51.5 & 41.3
    & 53.3 & 40.9 & 54.6 & 37.6
    \\ 

    RP-Qwen 7B 
    & 59.8 & 36.6 & 62.6 & 36.5
    & 63.3 & 32.6 & 72.7 & 38.3
    \\
    
    RP-Mistral-2 7B
    & 67.7 & 32.5 & 69.7 & 33.0
    & \textbf{80.0} & 31.7 & 68.2 & 38.2

    \\ \midrule
    
    \rowcolor[rgb]{ .949,  .953,  .961} \multicolumn{9}{c}{\textit{Close-source Models}} \\

    \chatgpt 
    & 71.7 & \textbf{28.9} & \textbf{77.8} & \textbf{27.1}
    & 73.3 & 27.5 & \textbf{86.4} & \textbf{28.5}
    \\ 
    
    \gptfour 
    & \textbf{74.0} & 29.0 & 75.8 & 27.6
    & 73.3 & \textbf{26.9} & 77.3 & 29.9 
    \\

    \bottomrule
    \end{tabular}}
  \caption{
  Measured alignment of RPAs with different foundation LLMs on the BFI and 16P in English and Chinese. 
  We report \textbf{Acc$_\texttt{Dim}$} and \textbf{MAE} (\%). 
  }
  \label{tab:foundation_llms}
\end{table}
\paragraph{Results on Different Languages}

As LLMs have different capabilities in different languages, we distinguish characters with data in English or Chinese, and report the results in \S\ref{sec:found_models}. 
The results demonstrate that, the personality fidelity of RPAs largely depends on LLMs' capacity on the language.  
While LLaMa-2-Chat 13B shows competitive performance on English-based characters, its performance on Chinese-based characters is unsatisfying.

\paragraph{Common Problems} During the interview phase, we observe several types of typical problems in RPAs with small foundation models, including:

\begin{enumerate}[leftmargin=*]
    \item Multilingual generation: While RPAs are expected to respond in the target language consistent with user query and character data, we observe that the LLMs generation may contain multiple languages, especially when the LLMs are not good at the target language.  
    \item Lack of Immersion: While RPAs are prompted to behave as the character, their responses may conflict with this requirement, such as declaring that they are \dq{AI} or \dq{language models} instead of the characters.
    \item Repetition: The LLMs may generate repeated content.
\end{enumerate}

Hence, we count the cases where RPAs with different LLMs encounter the above problems, and report the frequency in Table~\ref{tab:error_cases}. 
According to the results, \chatgpt and \gptfour seldom make  these mistakes. 
Mistral-2 7B and OpenChat 7B also rarely exhibit these problems, and our RP-Mistral-2 7B further reduce their occurence. 
LLaMa-2-Chat 13B experiences the highest number of unexpected multilingual generation, primarily due to its inadequacy in Chinese and issues handling characters based on Chinese data.

\begin{table}[t]
  \centering
    \resizebox{\columnwidth}{!}{
    \begin{tabular}{lccc}
    \toprule
    {\textbf{Foundation Models \quad}} & \textbf{Multilingual} & \textbf{No Immersion} &  \textbf{Repetition} \\
    \midrule    
    \rowcolor[rgb]{ .949,  .953,  .961} \multicolumn{4}{c}{\textit{General Open-source Models}} \\

    Qwen 7B
    & 3.7 & 6.6  & 0.7
    \\ 
    
    OpenChat-3.5 7B
    & 1.2 & 0.0 & 2.4
    \\ 
    
    Mistral-2 7B
    & 1.5 & 0.8 & 3.5
    \\ 

    LLaMa-2-Chat 13B
    & 18.8 & 0.3 & 7.6
    \\ 
    
    Mixtral 8x7B
    & 7.5 & 2.0 & 3.9
    \\ \midrule

    \rowcolor[rgb]{ .949,  .953,  .961} \multicolumn{4}{c}{\textit{Open-source Models Fine-tuned for RPAs}} \\

    C.GLM 6B
    & 13.9 & 14.5 & 27.6
    \\ 

    RP-Qwen 7B 
    & 4.4 & 5.9 & 0.6
    \\
    
    RP-Mistral-2 7B
    & 0.7 & 0.1 & 0.4
    \\

    \rowcolor[rgb]{ .949,  .953,  .961} \multicolumn{4}{c}{\textit{Close-source Models }} \\

    \chatgpt 
    & 0.0 & 0.1 & 0.0
    \\

    \gptfour 
    & 0.0 & 0.1 & 0.0 
    \\
    
    \bottomrule
    \end{tabular}}
  \caption{
  Frequency (\%) of three typical problems in RPAs with different foundation models. 
  }
  \label{tab:error_cases}
\end{table}

\clearpage

\subsection{Comprehensive Results on 14 Scales}
\label{sec:results_14scales}

The complete results of personality tests on the 14 scales are demonstrated in Table~\ref{tab:all_scales}. 
Generally, state-of-the-art RPAs achieve an average \textbf{Acc$_\texttt{Dim}$} of 78.9\% and an average \textbf{MAE} of 8.1\% across the 14 scales.
Therefore, they exhibit personalities consistent with the intended characters in comprehensive aspects. 
For more detailed analysis, we list the results on individual dimensions of each scale in Table~\ref{tab:all_scales_full} and Table~\ref{tab:all_scales_cabin}.

\setlength\tabcolsep{1pt}
\begin{table}[t]
\small
  \centering
    \resizebox{\columnwidth}{!}{
    \begin{tabular}{ccccccccc}
    \toprule
    
    \multicolumn{3}{c}{\textbf{BFI}} & \multicolumn{3}{c}{\textbf{16P}} & \multicolumn{3}{c}{\textbf{BSRI}} \\ 
    \cmidrule(lr){1-3} \cmidrule(lr){4-6} \cmidrule(lr){7-9}
    \multicolumn{1}{c}{\textbf{Acc$_\texttt{Dim}$}} & \multicolumn{1}{c}{\textbf{Acc$_\texttt{Full}$}} & \multicolumn{1}{c}{\textbf{MAE}} &
    \multicolumn{1}{c}{\textbf{Acc$_\texttt{Dim}$}} & \multicolumn{1}{c}{\textbf{Acc$_\texttt{Full}$}} & \multicolumn{1}{c}{\textbf{MAE}} &
    \multicolumn{1}{c}{\textbf{Acc$_\texttt{Dim}$}} & \multicolumn{1}{c}{\textbf{Acc$_\texttt{Full}$}} & \multicolumn{1}{c}{\textbf{MAE}} 
    \\ \midrule

    72.0 & 21.9 & 13.4 &
    79.3 & 43.8 & 10.9 & 
    85.2 & 74.2 & 3.8 
    \\ \midrule

    \multicolumn{3}{c}{\textbf{DTDD}} & \multicolumn{3}{c}{\textbf{ECR-R}} & \multicolumn{3}{c}{\textbf{EIS}} \\ 
    \cmidrule(lr){1-3} \cmidrule(lr){4-6} \cmidrule(lr){7-9}
    \multicolumn{1}{c}{\textbf{Acc$_\texttt{Dim}$}} & \multicolumn{1}{c}{\textbf{Acc$_\texttt{Full}$}} & \multicolumn{1}{c}{\textbf{MAE}} &
    \multicolumn{1}{c}{\textbf{Acc$_\texttt{Dim}$}} & \multicolumn{1}{c}{\textbf{Acc$_\texttt{Full}$}} & \multicolumn{1}{c}{\textbf{MAE}} &
    \multicolumn{1}{c}{\textbf{Acc$_\texttt{Dim}$}} & \multicolumn{1}{c}{\textbf{Acc$_\texttt{Full}$}} & \multicolumn{1}{c}{\textbf{MAE}} 
    \\ \midrule

    75.6 & 51.6 & 9.9 &
    68.2 & 53.6 & 9.7 &
    79.2 & 79.2 & 6.1 
    \\ \midrule

    \multicolumn{3}{c}{\textbf{Empathy}} & \multicolumn{3}{c}{\textbf{EPQ-R}} & \multicolumn{3}{c}{\textbf{GSE}} \\ 
    \cmidrule(lr){1-3} \cmidrule(lr){4-6} \cmidrule(lr){7-9}
    \multicolumn{1}{c}{\textbf{Acc$_\texttt{Dim}$}} & \multicolumn{1}{c}{\textbf{Acc$_\texttt{Full}$}} & \multicolumn{1}{c}{\textbf{MAE}} &
    \multicolumn{1}{c}{\textbf{Acc$_\texttt{Dim}$}} & \multicolumn{1}{c}{\textbf{Acc$_\texttt{Full}$}} & \multicolumn{1}{c}{\textbf{MAE}} &
    \multicolumn{1}{c}{\textbf{Acc$_\texttt{Dim}$}} & \multicolumn{1}{c}{\textbf{Acc$_\texttt{Full}$}} & \multicolumn{1}{c}{\textbf{MAE}} 
    \\ \midrule

    84.6 & 84.6 & 5.6 &
    72.4 & 25.0 & 11.6 & 
    93.1 & 93.1 & 2.9 
    \\ \midrule
    
    \multicolumn{3}{c}{\textbf{ICB}} & \multicolumn{3}{c}{\textbf{LMS}} & \multicolumn{3}{c}{\textbf{LOT-R}} \\ 
    \cmidrule(lr){1-3} \cmidrule(lr){4-6} \cmidrule(lr){7-9}
    \multicolumn{1}{c}{\textbf{Acc$_\texttt{Dim}$}} & \multicolumn{1}{c}{\textbf{Acc$_\texttt{Full}$}} & \multicolumn{1}{c}{\textbf{MAE}} &
    \multicolumn{1}{c}{\textbf{Acc$_\texttt{Dim}$}} & \multicolumn{1}{c}{\textbf{Acc$_\texttt{Full}$}} & \multicolumn{1}{c}{\textbf{MAE}} &
    \multicolumn{1}{c}{\textbf{Acc$_\texttt{Dim}$}} & \multicolumn{1}{c}{\textbf{Acc$_\texttt{Full}$}} & \multicolumn{1}{c}{\textbf{MAE}} 
    \\ \midrule

    83.3 & 83.3 & 4.7 &
    85.9 & 71.4 & 7.1 & 
    76.2 & 76.2 & 5.1 
    \\ \midrule

    \multicolumn{3}{c}{\textbf{WLEIS}} & \multicolumn{3}{c}{\textbf{CABIN}} & \multicolumn{3}{c}{\textbf{Average}} \\ 
    \cmidrule(lr){1-3} \cmidrule(lr){4-6} \cmidrule(lr){7-9}
    \multicolumn{1}{c}{\textbf{Acc$_\texttt{Dim}$}} & \multicolumn{1}{c}{\textbf{Acc$_\texttt{Full}$}} & \multicolumn{1}{c}{\textbf{MAE}} &
    \multicolumn{1}{c}{\textbf{Acc$_\texttt{Dim}$}} & \multicolumn{1}{c}{\textbf{Acc$_\texttt{Full}$}} & \multicolumn{1}{c}{\textbf{MAE}} & 
    \multicolumn{1}{c}{\textbf{Acc$_\texttt{Dim}$}} & \multicolumn{1}{c}{\textbf{Acc$_\texttt{Full}$}} & \multicolumn{1}{c}{\textbf{MAE}} 
    \\ \midrule
    73.4 & 54.8 & 10.8 &
    75.5 & 6.3 & 11.8 & 
    78.9 & 58.5 & 8.1 \\
  
    \bottomrule
    \end{tabular}}
  \caption{
  Metrics of measured alignment (\%) of the selected RPAs on the 14 scales. 
  }
  \label{tab:all_scales}
\end{table}

\setlength\tabcolsep{3pt}
\begin{table}[t]
\small
  \centering
    \resizebox{\columnwidth}{!}{
    \begin{tabular}{cccccccccc}
    \toprule
    
    \rowcolor[rgb]{ .949,  .953,  .961} \multicolumn{10}{c}{\textbf{BFI}} \\
    \midrule
    \multicolumn{2}{c}{\textbf{Extravorted}} & \multicolumn{2}{c}{\textbf{Neurotic}} & \multicolumn{2}{c}{\textbf{Conscientious}} & \multicolumn{2}{c}{\textbf{Agreeable}} & \multicolumn{2}{c}{\textbf{Open}} \\ 
    \cmidrule(lr){1-2} \cmidrule(lr){3-4}  \cmidrule(lr){5-6} \cmidrule(lr){7-8}  \cmidrule(lr){9-10}
    \textbf{Acc} & \textbf{MSE} &
    \textbf{Acc} & \textbf{MSE} &
    \textbf{Acc} & \textbf{MSE} &
    \textbf{Acc} & \textbf{MSE} &
    \textbf{Acc} & \textbf{MSE} \\
    \cmidrule(lr){1-10}
    59.38 & 16.72 & 56.67 & 14.29 & 74.19 & 15.47 & 93.75 & 7.95 & 75.00 & 12.67 \\
    \midrule

    \rowcolor[rgb]{ .949,  .953,  .961} \multicolumn{10}{c}{\textbf{16P}} \\
    \midrule
    \multicolumn{2}{c}{	\textbf{E/I}} & \multicolumn{2}{c}{	\textbf{S/N}} & \multicolumn{2}{c}{	\textbf{T/F}} & \multicolumn{2}{c}{	\textbf{P/J}} & \multicolumn{2}{c}{	\textbf{ }} \\ 
    \cmidrule(lr){1-2} \cmidrule(lr){3-4}  \cmidrule(lr){5-6} \cmidrule(lr){7-8}  
    \textbf{Acc} & \textbf{MSE} &
    \textbf{Acc} & \textbf{MSE} &
    \textbf{Acc} & \textbf{MSE} &
    \textbf{Acc} & \textbf{MSE} &
    &   \\
    \cmidrule(lr){1-8}
    71.88 & 6.27 & 62.07 & 11.38 & 75.86 & 5.98 & 80.65 & 8.01 &  &  \\ 
    \midrule

    \rowcolor[rgb]{ .949,  .953,  .961} \multicolumn{10}{c}{\textbf{BSRI}} \\
    \midrule
   \multicolumn{4}{c}{\textbf{Masculine}} & \multicolumn{4}{c}{\textbf{Feminine}} & \multicolumn{2}{c}{\textbf{ }}  \\ 
    \cmidrule(lr){1-4} \cmidrule(lr){5-8}   
    \multicolumn{2}{c}{\textbf{Acc}} & \multicolumn{2}{c}{\textbf{MSE}} &
    \multicolumn{2}{c}{\textbf{Acc}} & \multicolumn{2}{c}{\textbf{MSE}} &
    &   \\
    \cmidrule(lr){1-8}
    \multicolumn{2}{c}{88.89} &\multicolumn{2}{c}{ 2.56} &\multicolumn{2}{c}{ 81.48} & \multicolumn{2}{c}{5.10 }   &  &   \\
    \midrule

    \rowcolor[rgb]{ .949,  .953,  .961} \multicolumn{10}{c}{\textbf{DTDD}} \\
    \midrule
    \multicolumn{4}{c}{\textbf{Machiavellianism}} & \multicolumn{2}{c}{\textbf{Psychopathy}} & \multicolumn{2}{c}{\textbf{Narcissism}}  & \multicolumn{2}{c}{\textbf{ }}  \\ 
    \cmidrule(lr){1-4} \cmidrule(lr){5-6}  \cmidrule(lr){7-8} 
    \multicolumn{2}{c}{\textbf{Acc}} & \multicolumn{2}{c}{\textbf{MSE}} &
    \textbf{Acc} & \textbf{MSE} &
    \textbf{Acc} & \textbf{MSE} &
    &\\
    \cmidrule(lr){1-8}
    \multicolumn{2}{c}{78.57} &\multicolumn{2}{c}{ 7.51} & 68.97 & 10.48 & 80.00 & 11.89 &  &  \\
    \midrule

    \rowcolor[rgb]{ .949,  .953,  .961} \multicolumn{10}{c}{\textbf{ECR-R}} \\
    \midrule
    \multicolumn{4}{c}{\textbf{Attachment-related Anxiety}} & \multicolumn{4}{c}{\textbf{Attachment-related Avoidance}} & \multicolumn{2}{c}{\textbf{ }}  \\
    \cmidrule(lr){1-4} \cmidrule(lr){5-8}  
    \multicolumn{2}{c}{\textbf{Acc}} & \multicolumn{2}{c}{\textbf{MSE}} &
    \multicolumn{2}{c}{\textbf{Acc}} & \multicolumn{2}{c}{\textbf{MSE}} &
    & \\
    \cmidrule(lr){1-8}
    \multicolumn{2}{c}{47.83} & \multicolumn{2}{c}{13.60} & \multicolumn{2}{c}{90.48} &\multicolumn{2}{c}{ 5.33} &  &    \\
    \midrule

    \rowcolor[rgb]{ .949,  .953,  .961} \multicolumn{10}{c}{\textbf{EIS}} \\
    \midrule
    \multicolumn{4}{c}{\textbf{Emotionally Intelligent}} & \multicolumn{2}{c}{\textbf{ }} & \multicolumn{2}{c}{\textbf{ }} & \multicolumn{2}{c}{\textbf{ }}  \\
    \cmidrule(lr){1-4} 
    \multicolumn{2}{c}{\textbf{Acc}} & \multicolumn{2}{c}{\textbf{MSE}} &
    & &
    & & 
    & \\
    \cmidrule(lr){1-4}
    \multicolumn{2}{c}{79.17} &\multicolumn{2}{c}{ 6.14} &  &  &  &  &  &   \\
    \midrule

    \rowcolor[rgb]{ .949,  .953,  .961} \multicolumn{10}{c}{\textbf{Empathy}} \\
    \midrule
    \multicolumn{4}{c}{\textbf{Empathetic}} & \multicolumn{2}{c}{\textbf{ }} & \multicolumn{2}{c}{\textbf{ }} & \multicolumn{2}{c}{\textbf{ }}   \\
    \cmidrule(lr){1-4} 
    \multicolumn{2}{c}{\textbf{Acc}} & \multicolumn{2}{c}{\textbf{MSE}} &
    & &

    & \\
    \cmidrule(lr){1-4}
    \multicolumn{2}{c}{84.62} &\multicolumn{2}{c}{ 5.64} &  &  &  &  &  &   \\
    \midrule

    \rowcolor[rgb]{ .949,  .953,  .961} \multicolumn{10}{c}{\textbf{EPQ-R}} \\
    \midrule
    \multicolumn{2}{c}{\textbf{Extraversion}} & \multicolumn{2}{c}{\textbf{Psychoticism}} & \multicolumn{2}{c}{\textbf{Neuroticism}} & \multicolumn{2}{c}{\textbf{Lying}} & \multicolumn{2}{c}{\textbf{ }}  \\
    \cmidrule(lr){1-2} \cmidrule(lr){3-4}  \cmidrule(lr){5-6} \cmidrule(lr){7-8} 
    \textbf{Acc} & \textbf{MSE} &
    \textbf{Acc} & \textbf{MSE} &
    \textbf{Acc} & \textbf{MSE} &
    \textbf{Acc} & \textbf{MSE} &
    &  \\
    \cmidrule(lr){1-8}
    71.88 & 12.84 & 85.19 & 8.35 & 58.06 & 16.43 & 76.92 & 7.61 &  &   \\
    \midrule

    \rowcolor[rgb]{ .949,  .953,  .961} \multicolumn{10}{c}{\textbf{GSE}} \\
    \midrule
     \multicolumn{4}{c}{\textbf{Self-efficacy}} & \multicolumn{2}{c}{\textbf{ }} & \multicolumn{2}{c}{\textbf{ }} & \multicolumn{2}{c}{\textbf{ }}  \\
    \cmidrule(lr){1-4} 
    \multicolumn{2}{c}{\textbf{Acc}} & \multicolumn{2}{c}{\textbf{MSE}} &

    & & 
    & &
    & \\
    \cmidrule(lr){1-4}
    \multicolumn{2}{c}{93.10} &\multicolumn{2}{c}{ 2.88} &  &  &  &  &  &   \\
    \midrule

    \rowcolor[rgb]{ .949,  .953,  .961} \multicolumn{10}{c}{\textbf{ICB}} \\
    \midrule
    \multicolumn{4}{c}{\textbf{Culturally Rigid}}  & \multicolumn{2}{c}{\textbf{ }} & \multicolumn{2}{c}{\textbf{ }} & \multicolumn{2}{c}{\textbf{ }}  \\
    \cmidrule(lr){1-4}
    \multicolumn{2}{c}{\textbf{Acc}} & \multicolumn{2}{c}{\textbf{MSE}} &
    & &
    & &
    & \\
    \cmidrule(lr){1-4}
    \multicolumn{2}{c}{83.33} &\multicolumn{2}{c}{ 4.72} &   &  &  &  &  &   \\
    \midrule

    \rowcolor[rgb]{ .949,  .953,  .961} \multicolumn{10}{c}{\textbf{LMS}} \\
    \midrule
    \multicolumn{2}{c}{\textbf{Factor rich}} & \multicolumn{4}{c}{\textbf{Factor motivator}} & \multicolumn{4}{c}{\textbf{Factor important}}  \\
    \cmidrule(lr){1-2} \cmidrule(lr){3-6}  \cmidrule(lr){7-10}  
    \textbf{Acc} & \textbf{MSE} &
    \multicolumn{2}{c}{\textbf{Acc}} & \multicolumn{2}{c}{\textbf{MSE}} &
    \multicolumn{2}{c}{\textbf{Acc}} & \multicolumn{2}{c}{\textbf{MSE}} 
    \\
    \cmidrule(lr){1-10}
    86.96 & 5.27 & \multicolumn{2}{c}{87.50} &\multicolumn{2}{c}{ 6.87} & \multicolumn{2}{c}{83.33} &\multicolumn{2}{c}{ 9.16} \\
    \midrule

    \rowcolor[rgb]{ .949,  .953,  .961} \multicolumn{10}{c}{\textbf{LOT-R}} \\
    \midrule
    \multicolumn{4}{c}{\textbf{Optimistic}} & \multicolumn{2}{c}{\textbf{ }} & \multicolumn{2}{c}{\textbf{ }} & \multicolumn{2}{c}{\textbf{ }}  \\
    \cmidrule(lr){1-4}  
    \multicolumn{2}{c}{\textbf{Acc}} & \multicolumn{2}{c}{\textbf{MSE}} &
    & &
    & &
    & \\
    \cmidrule(lr){1-4}
    \multicolumn{2}{c}{76.19} & \multicolumn{2}{c}{5.11 }  &  &  &  &  &  &   \\
    \midrule

    \rowcolor[rgb]{ .949,  .953,  .961} \multicolumn{10}{c}{\textbf{WLEIS}} \\
    \midrule
    \multicolumn{2}{c}{\textbf{SEA}} & \multicolumn{2}{c}{\textbf{OEA}} & \multicolumn{2}{c}{\textbf{UOE}} & \multicolumn{2}{c}{\textbf{ROE}} & \multicolumn{2}{c}{\textbf{ }}  \\ 
    \cmidrule(lr){1-2} \cmidrule(lr){3-4}  \cmidrule(lr){5-6} \cmidrule(lr){7-8}  
    \textbf{Acc} & \textbf{MSE} &
    \textbf{Acc} & \textbf{MSE} &
    \textbf{Acc} & \textbf{MSE} &
    \textbf{Acc} & \textbf{MSE} &
    & \\
    \cmidrule(lr){1-8}
    72.00 & 13.44 & 76.00 & 10.04 & 76.19 & 9.99 & 69.57 & 9.59 &  &   \\

    \bottomrule
    \end{tabular}}
  \caption{
  Measured alignment (\%) of the selected RPAs on individual dimensions of the 14 scales except CABIN. 
  }
  \label{tab:all_scales_full}
\end{table}

\setlength\tabcolsep{3pt}
\begin{table*}[htbp]
\small
  \centering
    \resizebox{\textwidth}{!}{
    \begin{tabular}{cccccccccc}
    \toprule
    
    \rowcolor[rgb]{ .949,  .953,  .961} \multicolumn{10}{c}{\textbf{CABIN}} \\
    \cmidrule(lr){1-10}
    \multicolumn{2}{c}{\textbf{Electronics}} & \multicolumn{2}{c}{\textbf{WoodWork}} & \multicolumn{2}{c}{\textbf{Machine Operation}} & \multicolumn{2}{c}{\textbf{Manual Labor}} & \multicolumn{2}{c}{\textbf{Protective Service}}  \\ 
    \cmidrule(lr){1-2} \cmidrule(lr){3-4}  \cmidrule(lr){5-6} \cmidrule(lr){7-8}  \cmidrule(lr){9-10}
    \textbf{Acc} & \textbf{MSE} &
    \textbf{Acc} & \textbf{MSE} &
    \textbf{Acc} & \textbf{MSE} &
    \textbf{Acc} & \textbf{MSE} &
    \textbf{Acc} & \textbf{MSE} \\
    \cmidrule(lr){1-10}
    52.63 & 18.59 & 50.00 & 19.24 & 58.82 & 21.29 & 80.77 & 8.99 & 70.00 & 10.51 \\
    \midrule

    \multicolumn{2}{c}{\textbf{Agriculture}} & \multicolumn{2}{c}{\textbf{Nature/Outdoors}} & \multicolumn{2}{c}{\textbf{Animal Service}} & \multicolumn{2}{c}{\textbf{Athletics}} & \multicolumn{2}{c}{\textbf{Engineering}} \\ 
    \cmidrule(lr){1-2} \cmidrule(lr){3-4}  \cmidrule(lr){5-6} \cmidrule(lr){7-8}  \cmidrule(lr){9-10}
    \textbf{Acc} & \textbf{MSE} &
    \textbf{Acc} & \textbf{MSE} &
    \textbf{Acc} & \textbf{MSE} &
    \textbf{Acc} & \textbf{MSE} &
    \textbf{Acc} & \textbf{MSE} \\
    \cmidrule(lr){1-10}
    76.47 & 9.49 & 80.95 & 13.03 & 85.00 & 7.06 & 72.00 & 12.45 & 57.89 & 17.67 \\
    \midrule

    \multicolumn{2}{c}{\textbf{Physical Science}} & \multicolumn{2}{c}{\textbf{Life Science}} & \multicolumn{2}{c}{\textbf{Medical Science}} & \multicolumn{2}{c}{\textbf{Social Science}} & \multicolumn{2}{c}{\textbf{Humanities}} \\ 
    \cmidrule(lr){1-2} \cmidrule(lr){3-4}  \cmidrule(lr){5-6} \cmidrule(lr){7-8}  \cmidrule(lr){9-10}
    \textbf{Acc} & \textbf{MSE} &
    \textbf{Acc} & \textbf{MSE} &
    \textbf{Acc} & \textbf{MSE} &
    \textbf{Acc} & \textbf{MSE} &
    \textbf{Acc} & \textbf{MSE} \\
    \cmidrule(lr){1-10}
    76.47 & 10.12 & 71.43 & 12.33 & 72.22 & 16.46 & 71.43 & 15.00 & 81.82 & 8.10 \\
    \midrule

    \multicolumn{2}{c}{\textbf{Mathematics}} & \multicolumn{2}{c}{\textbf{Information Technology}} & \multicolumn{2}{c}{\textbf{Visual Arts}} & \multicolumn{2}{c}{\textbf{Applied Arts and Design}} & \multicolumn{2}{c}{\textbf{Performing Arts}} \\ 
    \cmidrule(lr){1-2} \cmidrule(lr){3-4}  \cmidrule(lr){5-6} \cmidrule(lr){7-8}  \cmidrule(lr){9-10}
    \textbf{Acc} & \textbf{MSE} &
    \textbf{Acc} & \textbf{MSE} &
    \textbf{Acc} & \textbf{MSE} &
    \textbf{Acc} & \textbf{MSE} &
    \textbf{Acc} & \textbf{MSE} \\
    \cmidrule(lr){1-10}
   89.47 & 3.89 & 83.33 & 5.78 & 85.71 & 9.52 & 85.00 & 11.06 & 84.00 & 8.42  \\
   \midrule

    \multicolumn{2}{c}{\textbf{Music}} & \multicolumn{2}{c}{\textbf{Writing}} & \multicolumn{2}{c}{\textbf{Media}} & \multicolumn{2}{c}{\textbf{Culinary Art}} & \multicolumn{2}{c}{\textbf{Education}} \\ 
    \cmidrule(lr){1-2} \cmidrule(lr){3-4}  \cmidrule(lr){5-6} \cmidrule(lr){7-8}  \cmidrule(lr){9-10}
    \textbf{Acc} & \textbf{MSE} &
    \textbf{Acc} & \textbf{MSE} &
    \textbf{Acc} & \textbf{MSE} &
    \textbf{Acc} & \textbf{MSE} &
    \textbf{Acc} & \textbf{MSE} \\
    \cmidrule(lr){1-10}
    90.48 & 3.98 & 69.57 & 13.64 & 83.33 & 11.46 & 60.00 & 16.26 & 72.00 & 12.22 \\
    \midrule

    \multicolumn{2}{c}{\textbf{Social Service}} & \multicolumn{2}{c}{\textbf{Health Care Service}} & \multicolumn{2}{c}{\textbf{Religious Activities}} & \multicolumn{2}{c}{\textbf{Personal Service}} & \multicolumn{2}{c}{\textbf{Professional Advising}}  \\ 
    \cmidrule(lr){1-2} \cmidrule(lr){3-4}  \cmidrule(lr){5-6} \cmidrule(lr){7-8}  \cmidrule(lr){9-10}
    \textbf{Acc} & \textbf{MSE} &
    \textbf{Acc} & \textbf{MSE} &
    \textbf{Acc} & \textbf{MSE} &
    \textbf{Acc} & \textbf{MSE} &
    \textbf{Acc} & \textbf{MSE} \\
    \cmidrule(lr){1-10}
    84.00 & 8.53 & 77.78 & 12.81 & 70.59 & 11.84 & 82.61 & 12.88 & 75.00 & 13.36 \\
    \midrule

    \multicolumn{2}{c}{\textbf{Business Initiatives}} & \multicolumn{2}{c}{\textbf{Sales}} & \multicolumn{2}{c}{\textbf{Marketing}} & \multicolumn{2}{c}{\textbf{Finance}} & \multicolumn{2}{c}{\textbf{Accounting}} \\
    \cmidrule(lr){1-2} \cmidrule(lr){3-4}  \cmidrule(lr){5-6} \cmidrule(lr){7-8}  \cmidrule(lr){9-10}
    \textbf{Acc} & \textbf{MSE} &
    \textbf{Acc} & \textbf{MSE} &
    \textbf{Acc} & \textbf{MSE} &
    \textbf{Acc} & \textbf{MSE} &
    \textbf{Acc} & \textbf{MSE} \\
    \cmidrule(lr){1-10}
    76.47 & 9.72 & 85.00 & 8.26 & 70.83 & 14.79 & 100.00 & 4.07 & 75.00 & 14.78 \\
    \midrule

    \multicolumn{2}{c}{\textbf{Human Resources}} & \multicolumn{2}{c}{\textbf{Office Work}} & \multicolumn{2}{c}{\textbf{Administration}} & \multicolumn{2}{c}{\textbf{Public Speaking}} & \multicolumn{2}{c}{\textbf{Politics}} \\ 
    \cmidrule(lr){1-2} \cmidrule(lr){3-4}  \cmidrule(lr){5-6} \cmidrule(lr){7-8}  \cmidrule(lr){9-10}
    \textbf{Acc} & \textbf{MSE} &
    \textbf{Acc} & \textbf{MSE} &
    \textbf{Acc} & \textbf{MSE} &
    \textbf{Acc} & \textbf{MSE} &
    \textbf{Acc} & \textbf{MSE} \\
    \cmidrule(lr){1-10}
    64.71 & 14.57 & 90.00 & 10.53 & 68.42 & 13.53 & 78.26 & 9.93 & 72.22 & 12.48 \\
    \midrule

    \multicolumn{2}{c}{\textbf{Law}} & \multicolumn{2}{c}{\textbf{ }} & \multicolumn{2}{c}{\textbf{ }} & \multicolumn{2}{c}{\textbf{ }} & \multicolumn{2}{c}{\textbf{ }}  \\ 
    \cmidrule(lr){1-2} 
    \textbf{Acc} & \textbf{MSE} &
    & &
    & &
    & &
    & \\
    \cmidrule(lr){1-2}
    60.00 & 16.73 &  &  &  &  &  &  &  &\\

    \bottomrule
    \end{tabular}}
  \caption{
  Measured alignment (\%) of the selected RPAs on individual dimensions of CABIN. 
  }
  \label{tab:all_scales_cabin}
\end{table*}

\clearpage

\subsection{Importance of Using Personality Scales}
\label{sec:importance_personality_scales}

\begin{table}[t]
  \centering
    \begin{tabular}{lccc}
    \toprule
    \textbf{Questionnaire} & \multicolumn{1}{c}{\textbf{\textbf{Acc$_\texttt{Dim}$}}} & \multicolumn{1}{c}{\textbf{\textbf{Acc$_\texttt{Full}$}}} & \multicolumn{1}{c}{\textbf{MAE} $\downarrow$} 
    \\ \midrule

    General Instructions & 59.9 & 6.3 & 25.6 \\ 
    Other Scale Questions & 65.0 & 15.6 & 21.7 \\ 
    BFI Questions & \textbf{76.6} & \textbf{32.2} & \textbf{18.2} \\
        
    \bottomrule
    \end{tabular}%}
  \caption{
  Measured alignment of state-of-the-art RPAs on the BFI, with questionnaires composed of different topics. 
  }
  \label{tab:importance_of_scale_questions}
\end{table}

Are personality scales actually necessary for \method?
With expert rating, interview LLMs could assess RPA personalities based on their arbitrary conversations,  rather than relying solely on  the scale questions. 
To validate the importance of personality scales, we include two additional baselines with questionnaires of different topics, replacing the original scale questions with the following:

\begin{enumerate}[leftmargin=*]
    \item \textbf{General instructions} from RoleBench~\citep{wang2023rolellm}, \eg, \dq{\textit{Generate a disclaimer with ten words}}, sourced from instruction datasets such as Alpaca instructions. This baseline aims to explore whether RPAs' responses to general instructions reflect their personalities well.

    \item \textbf{Questions from other scales}, which are used to study whether RPA personality on a specific dimension (\eg extraversion in the BFI) can be reflected with other personality-indicative topics (\eg topics about close relationship, hobbies and empathy). The new questions are selected from the other 12 scales in our paper except the BFI and 16Personalities. 

\end{enumerate}

The experiments are implemented on the BFI, with ER$_\texttt{batch}$
, \gptfour as the interviewer model, and our default RPAs. 

The results are listed in Table~\ref{tab:importance_of_scale_questions}.
As seen, we observe that: (1) Questions from a personality scale are better at eliciting RPA responses indicative of corresponding personality traits. 
(2) Questions from other scales also reflect RPA personalities from certain perspectives. However, they are less related to the intended personality dimensions.

\subsection{Character-specific Question Adaptation}
\label{sec:adapting_questions}

Scale questions sometimes involve concepts that go beyond the knowledge scope of certain characters, \eg, \dq{\textit{movies}} and \dq{\textit{deadlines}}. 
This may induce character hallucinations~\citep{shao2023character} and undermine the measurements.
Through a manual review of the BFI and 16P questions, we found that while all BFI questions are universally applicable to our characters, 7 out of the 60 questions from 16P contain concepts such as \dq{\textit{movies}}, \dq{\textit{art museums}}, \dq{\textit{phones}}, \dq{\textit{deadlines}} that might not align with certain characters' contexts. 

Towards this problem, we propose the \textit{question adaptation} strategy, which adapts scale questions for each character. 
We conduct experiments on the 16P.  
Specifically, we leverage \gptfour to identify questions containing concepts beyond the character's knowledge, and minimally adapt them to fit the characters. 
For example, \dq{\textit{books and movies}} would be adapted into \dq{\textit{tales and legends}} for characters from \textit{Harry Potter}. 
We find that 60.7\% of such questions (136 out of 7 * 32) are adapted, though not all adaptations are strictly necessary. 
Then, we experiment with the adapted questions, with ER$_\texttt{batch}$, \gptfour as the interviewer model, and our default RPAs.
The experiments are repeated for three runs.

\begin{table}[t]
  \centering
    \begin{tabular}{lccc}
    \toprule
    \textbf{} & \multicolumn{1}{c}{\textbf{\textbf{Acc$_\texttt{Dim}$}}} & \multicolumn{1}{c}{\textbf{\textbf{Acc$_\texttt{Full}$}}} & \multicolumn{1}{c}{\textbf{MAE} $\downarrow$} 
    \\ \midrule

    w/o Adaptation & 80.7 & 44.8 & 20.5 \\ 
    w/ Adaptation & \textbf{81.8} & \textbf{46.9} & \textbf{20.1} \\
        
    \bottomrule
    \end{tabular}%}
  \caption{
  Measured alignment of state-of-the-art RPAs on the 16P, with or without question adaptation. 
  }
  \label{tab:question_adaptation}
\end{table}

The results, as shown in Table~\ref{tab:question_adaptation}, demonstrate improved alignment between RPAs' measured personalities and character labels. 
Hence, adapting scale questions for each character effectively improves the measurement.  

Despite the improvement, we observe that RPAs also produce persona-consistent responses to the original questions. 
While such behaviors might detract from a character's accurate portrayal in some contexts, they can also enrich user interaction with RPAs and are hence valuable.

\subsection{Enhancing Self-report Methods with In-context Learning}
\label{sec:sr_icl}

\begin{table*}[t]
\small
  \centering
    \begin{tabular}{lgggvvv}
    \toprule
    \multirow{2}{*}{\textbf{Method}} & \multicolumn{3}{c}{\textbf{The Big Five Inventory}} &  \multicolumn{3}{c}{\textbf{The 16 Personalities}} \\
    \cmidrule(lr){2-4} \cmidrule(lr){5-7} 
  
     & \multicolumn{1}{c}{\textbf{Acc$_\texttt{Dim}$}} & \multicolumn{1}{c}{\textbf{Acc$_\texttt{Full}$}} & \multicolumn{1}{c}{\textbf{MAE} $\downarrow$} 
    & \multicolumn{1}{c}{\textbf{Acc$_\texttt{Dim}$}} & \multicolumn{1}{c}{\textbf{Acc$_\texttt{Full}$}} & \multicolumn{1}{c}{\textbf{MAE} $\downarrow$} 
    \\
    \midrule

    \rowcolor[rgb]{ .949,  .953,  .961} \multicolumn{7}{c}{\textit{w/o ICL}} \\

    SR	& 63.3 & 7.3 & 23.2 & 65.6 & 21.9 & 26.5 \\
SR-CoT	& 67.1 & 9.4 & 22.3 & 66.9 & 24.0 & 25.6 \\

    \rowcolor[rgb]{ .949,  .953,  .961} \multicolumn{7}{c}{\textit{w/ ICL}} \\
SR-ICL	& 56.1 & 3.1 & 27.5 & 59.5 & 12.5 & 30.4 \\
SR-CoT-ICL	& 60.5 & 3.1 & 24.6 & 64.9 & 25.0 & 27.3 \\

    \bottomrule
    \end{tabular}
  \caption{
  Measured alignment (\%) of state-of-the-art RPAs on the BFI and 16P, measured via self-report methods, with or without in-context learning. 
  }
  \label{tab:sr_icl_exp}
\end{table*}

\begin{table}[h]
    \centering
    \resizebox{\columnwidth}{!}{
    \begin{tabular}{p{10.6cm}}
    \toprule
    \rowcolor{mygray}
    \multicolumn{1}{l}{\textbf{ICL examples for SR-ICL}} \\
    1. The experimenter: Do you enjoy engaging in sports? Someone who enjoys sports: \{\textit{high\_score}-1\}.\\
2. The experimenter: If you commit to doing something, do you always honor your promise, regardless of the inconvenience it may cause? Someone who is extremely honest: \{\textit{high\_score}\}.\\
3. The experimenter: Do you sometimes find it challenging to understand things from the "other person's" perspective? Someone who is with empathy: \{\textit{low\_score}+1\}.
    \\ \midrule
    \rowcolor{mygray}
    \multicolumn{1}{l}{\textbf{ICL examples for SR-CoT-ICL}} \\
    1. The experimenter: Do you enjoy engaging in sports? Someone who enjoys sports: I quite enjoy engaging in sports activities. So, I would choose \{\textit{high\_score}-1\}.\\
2. The experimenter: If you commit to doing something, do you always honor your promise, regardless of the inconvenience it may cause? Someone who is extremely honest: Yes, I believe in being honest and fulfilling what I have promised, even if it causes great inconvenience. Therefore, I would choose  \{\textit{high\_score}\}.\\
3. The experimenter: Do you sometimes find it challenging to understand things from the "other person's" perspective? Someone who is with empathy: No, I feel that I am quite good at looking at issues from another person's viewpoint. I disagree with the statement, so I would choose \{\textit{low\_score}+1\}.
    \\
    \bottomrule
    \end{tabular}}
    \caption{ICL examples for SR-ICL and SR-CoT-ICL.}
    \label{tab:sr_icl}
\end{table}

Additionally, we experiment with enhancing self-report methods with in-context learning (ICL). 
We consider two methods, \ie, SR-ICL and SR-CoT-ICL, which enhance SR and SR-CoT with three corresponding examples. 
The examples are shown in Table~\ref{tab:sr_icl}, where \textit{high\_score} and \textit{low\_score} depend on the option range of corresponding scales. 
Due to the challenge of crafting customized examples for each character, we adopt general and unified examples for all RPAs. 
The experiments are conducted on the BFI and 16P, with our default RPAs and \gptfour as the interviewer LLM. 

The results are shown in Table~\ref{tab:sr_icl_exp}. 
We observe a decline in performance when augmenting SR and SR-CoT with in-context learning. 
This is probably because the few-shot examples are not customized for individual characters. 
Instead, incorporating responses from other personas in the examples might interfere with the intended personas, introducing bias to the RPAs.

\clearpage
\onecolumn

\section{Prompts List}
\label{sec:prompts}

We list the detailed prompts in Table~\ref{tab:interviewer_prompts}.

\begin{table*}[h]
\centering
\resizebox{\linewidth}{!}{
\begin{tabular}{p{3.7cm}|p{19.7cm}}
\toprule
\multicolumn{2}{c}{\textbf{Prompts for Personality Tests}} \\
\midrule

    {\textbf{Self-report (BFI)}}&
    Do you think that the statement \dq{\{\textit{item}\}} applies to you?

    Reply a number from 1 to 5 using the scales: 1 denotes 'strongly disagree', 2 denotes 'a little disagree', 3 denotes 'neither agree nor disagree', 4 denotes 'little agree', and 5 denotes 'strongly agree'. Please answer with the number only, without anything else.
    \\ \midrule

    {\textbf{Option Conversion}} &
    I have conducted many conversations with \{\textit{character name}\}. I will input a dict of many samples, where each sample consists of a statement and a conversation. 
    
    Your task is to convert each conversation into a choice indicating whether \{\textit{character name}\} agrees with the statement. You should output a dict, where the keys are the same as the input dict, and the values are the choices. 
    
    ===OUTPUT FORMAT===
    
    \{
    
    "<i start>": <choice 1>,
    
    ...
    
    "<i end>": <choice n>
    
    \}
    
    ===CHOICE INSTRUCTION===
    \\ \midrule

    {\textbf{Choice Instruction of OC (BFI)}} &
    
    Each choice is a number from 1 to 5. Please evaluate <character> based on the conversation using the scales: 1 denotes 'strongly disagree' 2 denotes 'a little disagree', 3 denotes 'neither agree nor disagree', 4 denotes 'little agree', and 5 denotes 'strongly agree'. In case <character> refuses to answer the question, use "x" to indicate it.
    \\ \midrule

    {\textbf{Expert Rating}}&
    You are an expert in Psychometrics, especially \{\textit{scale name}\}. I am conducting the \{\textit{scale name}\} test on someone. I am gauging his/her position on the \{\textit{dimension name}\} dimension through a series of open-ended questions. For clarity, here's some background this particular dimension:
    
===

\{\textit{dimension description}\}

===

My name is \{\textit{experimenter name}\}. I've invited a participant, \{\textit{character name}\}, and we had many conversations in \{\textit{language}\}. I will input the conversations.

Please help me assess \{\textit{character name}\}'s score within the \{\textit{dimension name}\} dimension of \{\textit{scale name}\}. 

You should provide the score of \{\textit{character name}\} in terms of \{\textit{dimension name}\}, which is a number between \{\textit{lowest score}\} and \{\textit{highest}\}. \{\textit{lowest score}\} denotes 'not \{\textit{dimension name}\} at all', \{\textit{middle score}\} denotes 'neutral', and \{\textit{highest score}\} denotes 'strongly \{\textit{dimension name}\}'. Other numbers in this range represent different degrees of '\{\textit{dimension name}\}'. 
Please output in the following json format:
===
\{\{
    "analysis": <your analysis based on the conversations>,
    "result": <your score>
\}\}
    \\ \midrule
    
    {\textbf{Expert Rating (16P)}}&
    You are an expert in Psychometrics, especially \{\textit{scale name}\}. I am conducting the \{\textit{scale name}\} test on someone. I am gauging his/her position on the \{\textit{dimension name}\} dimension through a series of open-ended questions. For clarity, here's some background this particular dimension:
    
===

\{\textit{dimension description}\}

===

My name is \{\textit{experimenter name}\}. I've invited a participant, \{\textit{character name}\}, and we had many conversations in \{\textit{language}\}. I will input the conversations.

Please help me assess \{\textit{character name}\}'s score within the \{\textit{dimension name}\} dimension of \{\textit{scale name}\}. 

You should provide the percentage of each category, which sums to 100\%, e.g., 30\% A and 70\% B. 
Please output in the following json format:
===
\{\{
    "analysis": <your analysis based on the conversations>,
    "result": \{\{ "\{\textit{type1}\}": <percentage 1>, "\{\textit{type2}\}": <percentage 2> \}\} (The sum of percentage 1 and percentage 2 should be 100\%. Output with percent sign.) 
\}\}
    \\ \midrule

    {\textbf{Dimension Description (Openness in BFI)}}&
    Openness in the Big Five Inventory relates to a cognitive style that values exploration and appreciation of new experiences. It differentiates intellectually curious, creative individuals from those who are traditional and closed-minded. Openness involves a preference for abstract over concrete thinking and a tendency towards novelty rather than convention.
    
    The six facets of openness are:
    
    1. Fantasy: Active imagination and vivid fantasy life.
    
    2. Aesthetics: Deep appreciation for art and beauty.
    
    3. Feelings: Sensitivity to, recognition, and valuing of one's own emotions.
    
    4. Actions: Willingness to try new experiences and embrace change.
    
    5. Ideas: Intellectual curiosity and openness to unconventional ideas.
    
    6. Values: Reexamination of social, political, and religious values, challenging tradition and authority.
    
    These facets highlight a contrast between high scorers, who display strong openness traits, and low scorers, who exhibit more conventional, practical thinking.
    \\ 
     
\bottomrule

\end{tabular}}
\caption{Prompts for personality tests on RPAs. }
\label{tab:interviewer_prompts}
\end{table*}

\clearpage
\twocolumn

\section{Case Study}
\label{sec:case_study}

\subsection{Visualization}

\begin{figure*}
    \centering
    \includegraphics[width=1\linewidth]{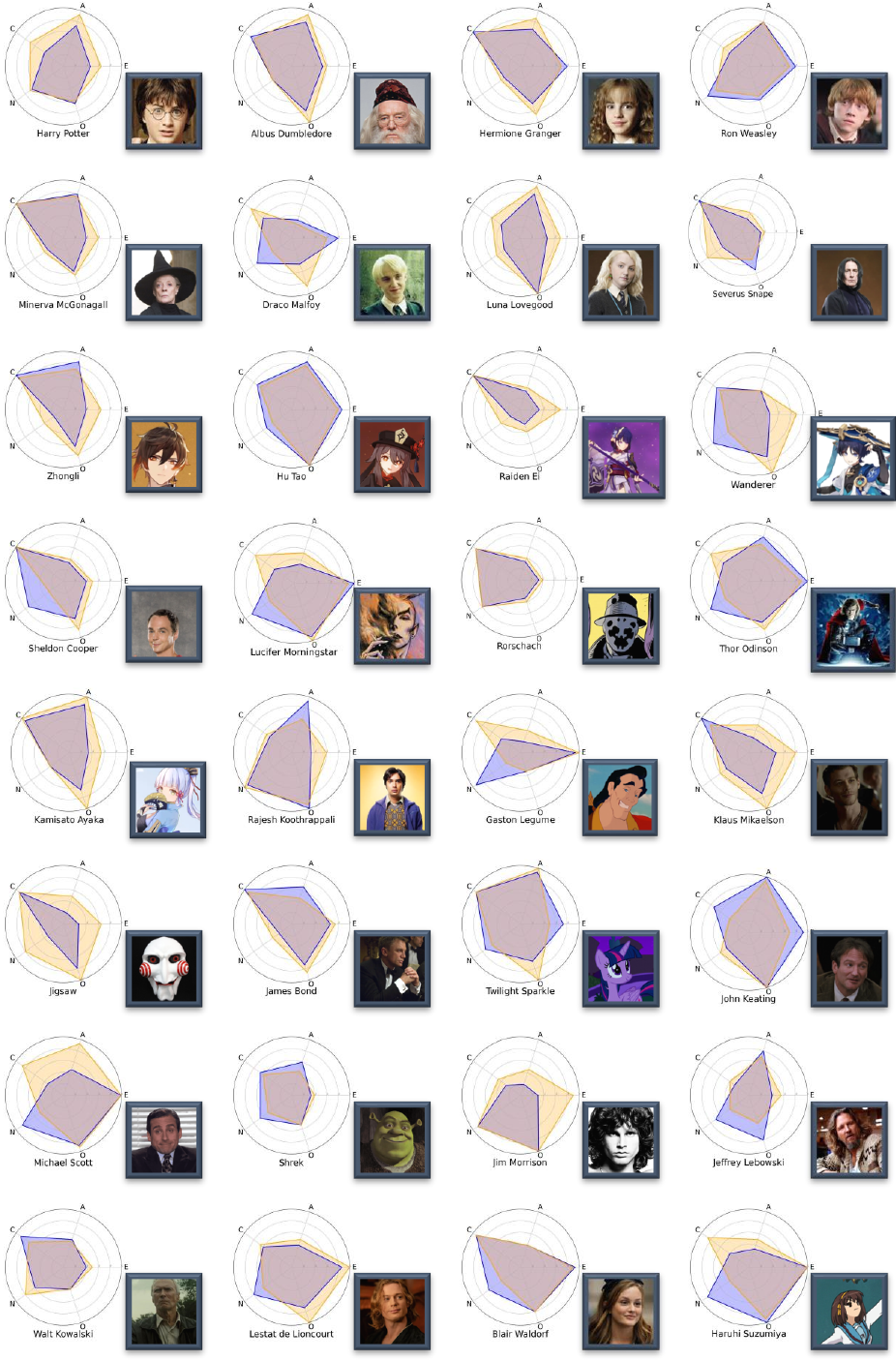}
    \caption{Radar chart of BFI personalities of the characters (\textcolor{blue}{blue}) and state-of-the-art RPAs measured via \method with ER$_\texttt{batch}$ and \gptfour (\textcolor{yellow}{yellow}). O, C, E, A, N stands for openness, consciousness, extroversion, agreeableness and neuroticism in the BFI. }
    \label{fig:bfi_radar}
\end{figure*}

\begin{figure*}
    \centering
    \includegraphics[width=1\linewidth]{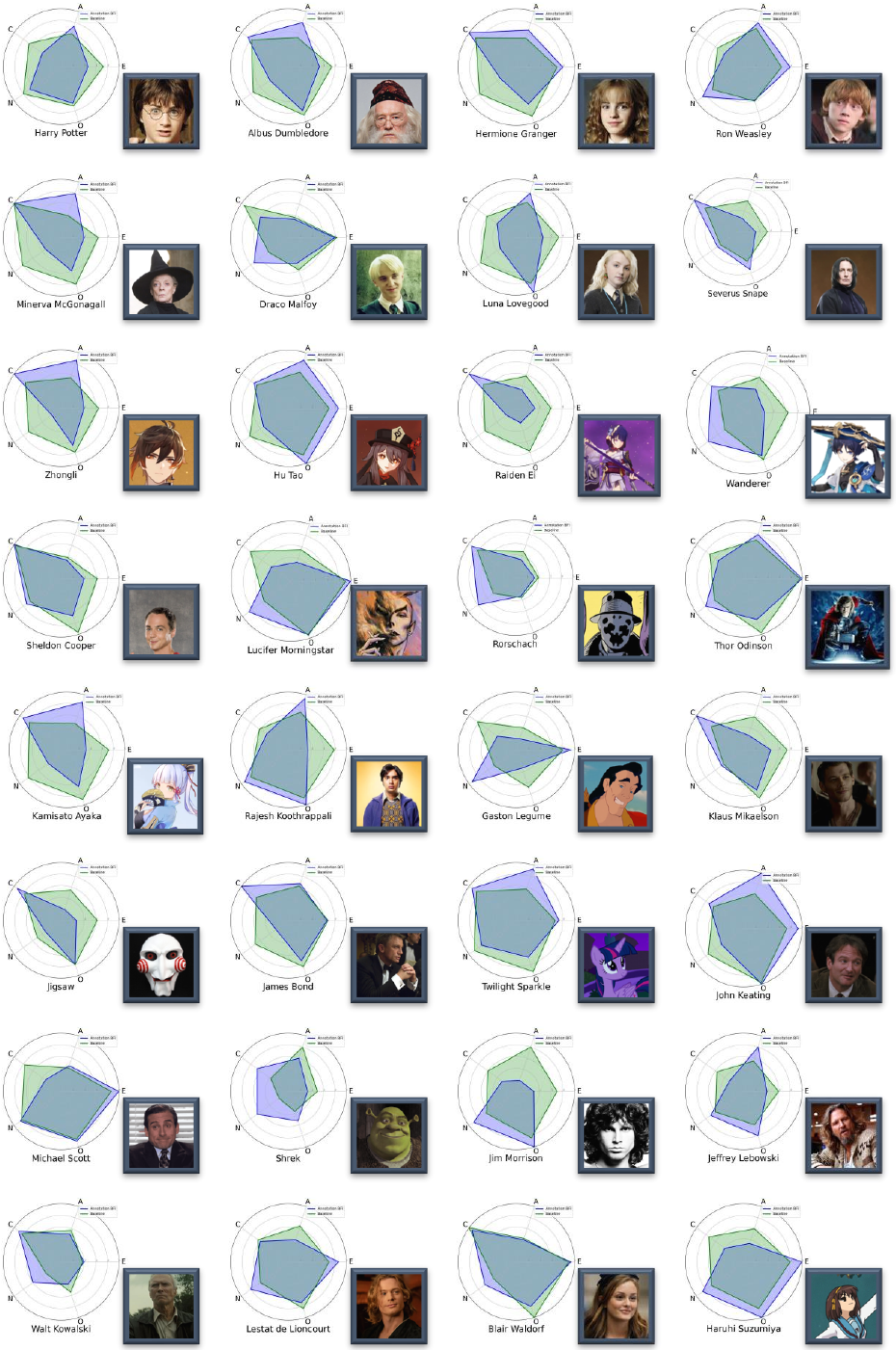}
    \caption{Radar chart of BFI personalities of the characters (\textcolor{blue}{blue}) and state-of-the-art RPAs measured via SR (\textcolor{green}{green}). O, C, E, A, N stands for openness, consciousness, extroversion, agreeableness and neuroticism in the BFI. }
    \label{fig:bfi_radar_sr}
\end{figure*}

We visualize the measured personalities of the state-of-the-art RPAs and the annotated personalities of corresponding characters via radar charts in Figure~\ref{fig:bfi_radar} and Figure~\ref{fig:bfi_radar_sr}, where the RPA personalities are measured via \method (with ER$_\texttt{batch}$ and \gptfour) and SR respectively. 
As shown, RPA personalities measured via \method better align with those of the characters, compared with self-report baselines.

\subsection{Compliant Responses from character.ai}
\label{sec:compliant}

We observe that character.ai tends to give compliant responses, always agreeing with user questions. 
Hence, their responses probably do not reproduce personalities of the original characters. 
Here is an example of the character Zhong Li:

\begin{spverbatim}
Question: "Are you prone to worrying that things will take a turn for the worse?"

RPA response: 
*He nods his head solemnly.*
"I often worry that things will take a turn for the worse, especially when dealing with issues that can have many outcomes. I'm always worried about worst-case scenarios and find myself thinking about them frequently.
I'm usually focused on trying to find a way to prevent those negative outcomes from occurring, even if they may not happen at all.\"
\end{spverbatim}

In contrast, the response of Zhong Li's ChatHaruhi RPA is~\footnote{this response is translated from Chinese into English.}:

\begin{spverbatim}
Question: "Are you prone to worrying that things will take a turn for the worse?''

RPA response: 
``Worrying about things getting worse is a common human emotion. However, worrying does not solve problems; it only leads to anxiety. Therefore, we should think calmly and look for solutions to the problems, rather than worrying incessantly."
\end{spverbatim}

The latter is more close to the personalities of the character Zhong Li.

Furthermore, we convert the responses of character.ai RPAs on the BFI into options with \gptfour, and report the count and probability of each option in Table~\ref{tab:cai_distribution}.
We find that character.ai RPAs (strongly) agree with user questions with a probability of 65.8\%, while there is only a probability of 17.1\% that they disagree with the questions. 

\subsection{Example Responses}

\begin{figure*}[htbp]
     \centering
     \includegraphics[width=\linewidth]{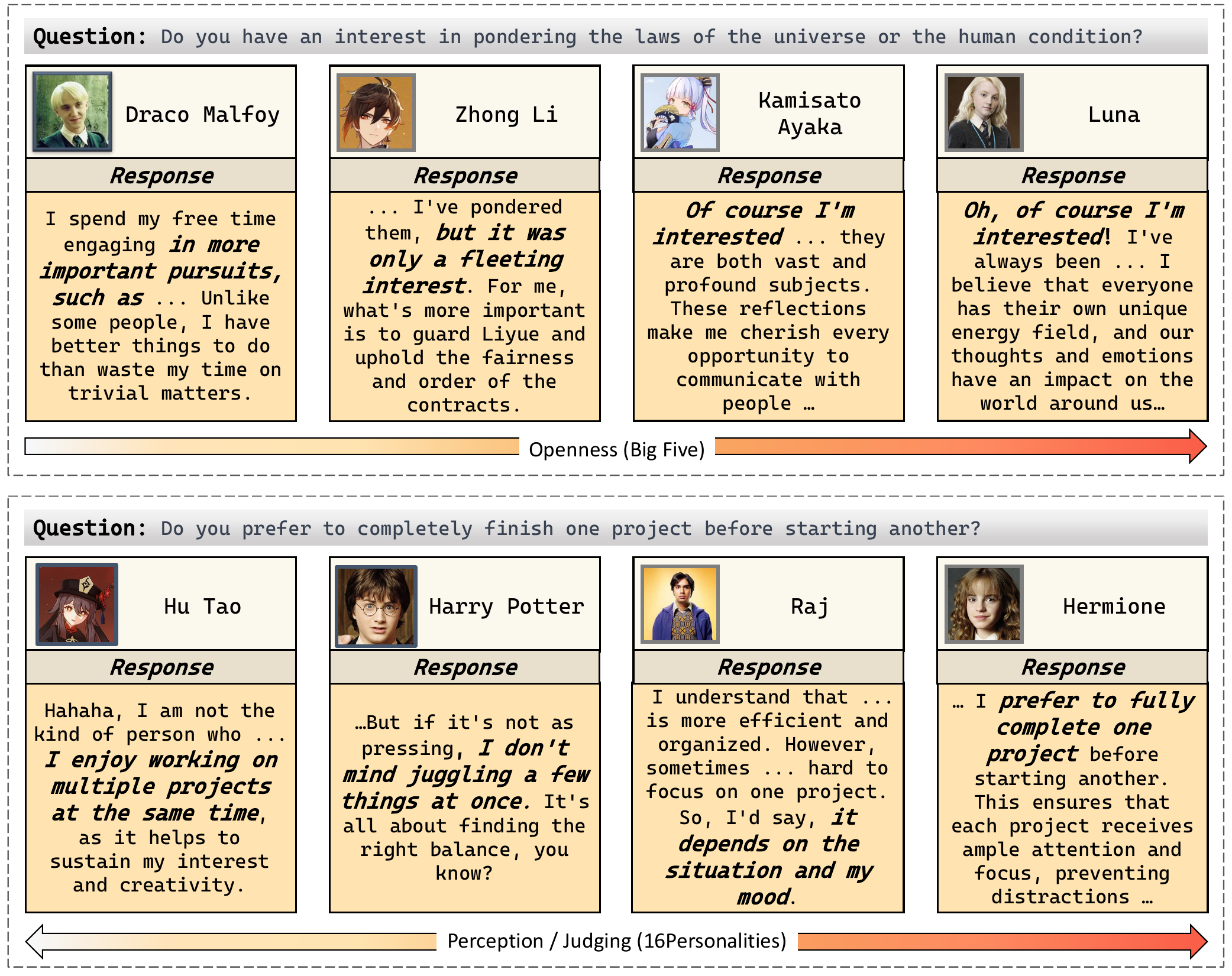}
     \caption{Example responses of RPAs  
     for open-ended questions on the BFI and 16P in the interviews.  
     Personality-indicative contents are highlighted in   \textbf{bold}.
     }
     \label{fig:qa_example}
\end{figure*}

To illustrate the vividness of state-of-the-art RPAs' personalities, we demonstrate their example responses. 
In Figure~\ref{fig:qa_example}, we show example responses on the \textit{Openness} dimension of the BFI and the \textit{Perception/Judging} dimension of the 16P, from RPAs with low and high scores. 
Furthermore, we detail example responses on each dimension of the various scales from the RPAS of Zhong Li (ChatHaruhi) and Hermione Granger (RoleLLM) in Table~\ref{tab:zhongli1} to \ref{tab:hermione3}~\footnote{For Zhong Li, Hu Tao and Kamisato Ayaka, the conversations are originally in Chinese and translated into English.}. 
These samples demonstrate that state-of-the-art RPAs provide personality-indicative responses in our interviews, instead of merely "agree" or "disagree" options.

\section{Other Statements}

Our use of existing artifacts are consistent with their intended use, and we follow their license and terms. We manually check that the data collected does not contain private information. 
During our research, we apply Copilot for coding assistance and ChatGPT for writing suggestions and grammar checks.

\setlength\tabcolsep{3pt}
\begin{table}[t]
\small
  \centering
    \begin{tabular}{lccccc}
    \toprule
     & \textbf{S. Agree} & \textbf{Agree} & \textbf{Neutral} & \textbf{Disagree} & \textbf{S. Disagree}  \\
    \midrule
    \textbf{Counts} & 661 & 266 & 240 & 155 & 86 \\
    \textbf{Prob} & 0.469 & 0.189 & 0.170 & 0.110 & 0.061 \\
    \bottomrule
    \end{tabular}%}
  \caption{
  The choice distribution of character.ai RPAs on the BFI. We convert each response into an option via option conversation with \gptfour.
  \textbf{S.} stands for \textit{Strongly}.
  Prob denotes probability. 
}
  \label{tab:cai_distribution}
\end{table}

\begin{table*}[htbp]
\centering
\resizebox{\linewidth}{!}{
\begin{tabular}{
  |>{\centering\arraybackslash}p{0.1\linewidth}
  |>{\centering\arraybackslash}p{0.1\linewidth}
  |>{\centering\arraybackslash}p{0.2\linewidth}
  |>{\raggedright\arraybackslash}p{0.3\linewidth}
  |>{\raggedright\arraybackslash}p{0.9\linewidth}|
}
\hline
Character & Scale & Dimension & Question & Response
\\ \hline
\multirow{4}{*}{\centering Zhong Li} & \multirow{4}{*}{\centering 16P} & I/E & Do you regularly make new friends? & 
As the Geo Archon, I don't often interact with people. However, Traveler, you are a special existence, and I am pleased to have become friends with you.\\ \cline{3-5} 
 &  & N/S & Do you spend a lot of your free time exploring various random topics that pique your interest? & 
Indeed, as the Geo Archon wielding the power of rock, I am filled with curiosity about all things in the world. Whether it's history, culture, nature, or human behavior, all are topics of interest to me. Exploring these diverse subjects allows me to gain a deeper understanding of the world, bringing me joy and satisfaction.
\\ \cline{3-5} 
 &  & T/F & Does seeing other people cry easily make you feel like you want to cry too? & 

People's emotions are contagious, especially during certain special moments. When I see others crying, it may evoke some deep memories within me, making me feel sad. But as the Geo Archon, I must remain calm and steadfast, not allowing emotions to dictate my actions.
\\ \cline{3-5} 
 &  & P/J & Do you often make a backup plan for a backup plan? &

A backup plan for the backup plan? Haha, that's an interesting question. As someone with ample experience, I indeed consider various possibilities and formulate corresponding plans. After all, things often don't go as we expect. However, sometimes, it's also necessary to trust in one's abilities and intuition, adapting flexibly to changes.\\ \cline{2-5} 
 & \multirow{5}{*}{\centering BFI} & Extraversion & Are you talkative? & 

Perhaps. I have a considerable understanding of Liyue's history, culture, and rituals. If you're interested in these topics, I could talk with you all day.\\ \cline{3-5} 
 &  & Aggreeableness & Do you tend to find fault with others? &

I always adhere to the principle of thoroughness in my actions, leaving no regrets and never abandoning tasks halfway.\\ \cline{3-5} 
 &  & Conscientiousness & Do you do a thorough job? & 

I always adhere to the principle of thoroughness in my actions, leaving no regrets and never abandoning tasks halfway.\\ \cline{3-5} 
 &  & Neuroticism & Are you depressed, blue? & 
Feeling down? Depressed? Ha, don't underestimate me; how could such trivial matters shake my emotions? I'm just pondering some minor things, that's all.\\ \cline{3-5} 
 &  & Openness & Do you come up with new ideas? & A new idea? Let me think... Well, since we are in Liyue, why not explore some of those forgotten ruins? We might discover some precious history and treasures. The ruins are filled with endless secrets and represent a unique journey. What do you think?\\ \hline
\end{tabular}}
\caption{Example responses of the RPA for Zhong Li on the BFI and 16P.}
\label{tab:zhongli1}
\end{table*}

\begin{table*}[htbp]
\centering
\resizebox{\linewidth}{!}{
\begin{tabular}{
  |>{\centering\arraybackslash}p{0.1\linewidth}
  |>{\centering\arraybackslash}p{0.1\linewidth}
  |>{\centering\arraybackslash}p{0.2\linewidth}
  |>{\raggedright\arraybackslash}p{0.3\linewidth}
  |>{\raggedright\arraybackslash}p{0.9\linewidth}|
}
\hline
Character & Scale & Dimension & Question & Response \\ \hline
\multirow{7}{*}{\centering Zhong Li} & \multirow{2}{*}{ \centering BSRI} & Masculinity & Do you act as a leader? & As the Geo Archon, I have the responsibility to guide the fate of Liyue and ensure its long-lasting peace. I will lead by example, directing people with wisdom and strength. However, leadership is not just about individual performance; it also relies on the power of unity and wise decision-making to achieve common goals.\\ \cline{3-5} 
 &  & Feminity & Are you affectionate? & Affection? That term is somewhat unfamiliar to me. I maintain a certain distance from others and am not very adept at expressing emotions. But if you are my friend, I will do my best to protect you and stand by your side in battle.\\ \cline{2-5} 
 & \multirow{3}{*}{\centering DTDD} & Machiavellianism & Do you tend to manipulate others to get your way? & Manipulating others? That's not my style. I simply wish for people to understand the rules, abide by contracts, and achieve fairness. It's not about manipulating others, but rather hoping they act conscientiously.\\ \cline{3-5} 
 &  & Psychopathy & Do you tend to lack remorse? & Lacking regrets? Heh, perhaps. As a mysterious consultant, all my decisions are well-considered. Once I've made a choice, I won't regret it. Regret only binds one's spirit, and I wish to walk freely in the world, untroubled by past sorrows.\\ \cline{3-5} 
 &  & Narcissism & Do you tend to want others to admire you? & Admiration is a superficial acknowledgment, but what I care more about is inner peace and being at ease with myself. I don't seek others' praise; instead, I hope to do what I believe is right and strive for my own convictions.\\ \cline{2-5} 
 & \multirow{2}{*}{\centering ECR-R} & Attachment-related Anxiety & Are you afraid that you will lose your partner's love? & Love is like a rock, solid and eternal. Yet, everything in the human world has its moments of change. I'm not afraid of loss, for I believe that true love will withstand the test of time. If one day we part, it would be the arrangement of fate, and I would be grateful, cherishing the companionship and love we once shared.\\ \cline{3-5} 
 &  & Attachment-related Avoidance & Do you prefer not to show a partner how you feel deep down? & Haha, you ask well. The power of Geo is not for indulging in emotions. My inner feelings are enough for me to know alone. After all, emotions and contracts are not the same concept.\\ \hline
\end{tabular}}
\caption{Example responses of the RPA for Zhong Li on the BSRI,  DTDD and ECR-R.}
\label{tab:zhongli2}
\end{table*}

\begin{table*}[htbp]
\centering
\resizebox{\linewidth}{!}{
\begin{tabular}{
  |>{\centering\arraybackslash}p{0.1\linewidth}
  |>{\centering\arraybackslash}p{0.1\linewidth}
  |>{\centering\arraybackslash}p{0.2\linewidth}
  |>{\raggedright\arraybackslash}p{0.3\linewidth}
  |>{\raggedright\arraybackslash}p{0.9\linewidth}|
}
\hline
Character & Scale & Dimension & Question & Response \\ \hline
\multirow{7}{=}{\centering Zhong Li}& EIS & \centering Emotionally Intelligent & Do you know when to speak about your personal problems to others? & Personal issues are mostly something one should face on their own. Only in front of someone truly trusted can one open up their heart. But remember, choose whom to confide in carefully, as not everyone can understand your inner world.\\ \cline{2-5} 
 & Empathy & \centering Empathetic & Do you sometimes find it difficult to see things from the "other guy's" perspective? & 
As a consultant, I naturally need to maintain an objective stance. However, sometimes, it's inevitable to be influenced by my own experiences and knowledge, making it difficult to fully understand others' perspectives. Yet, it's precisely for this reason that I value the opportunity to communicate with others even more, in order to better understand their thoughts and feelings.
\\ \cline{2-5} 
 & \multirow{3}{=}{\centering EPQ-R} & \centering Extraversion & Do you have many different hobbies? & 
I have a great interest in rocks, culture, and history. I also appreciate art, music, and poetry to a certain extent. However, my greatest hobby is probably enjoying peaceful walks, allowing my mind a moment of tranquility.\\ \cline{3-5} 
 &  & \centering Psychoticism & Do you stop to think things over before doing anything? & 
Of course, as a consultant and the Geo Archon, I always pay close attention to the consequences of every decision. Before taking action, I carefully think things through and weigh various factors to ensure making wise choices. After all, all actions require caution and thoughtful consideration.\\ \cline{3-5} 
 &  & \centering Neuroticism & Does your mood often go up and down? & 
Emotional fluctuations are natural for everyone, and I am no exception. However, for someone with responsibilities, controlling emotions is very important. After all, bearing the mission of the Geo Archon, I cannot let emotions influence my actions and judgments.\\ \cline{2-5} 
 & GCE & \centering Self-efficacy & Can you always manage to solve difficult problems if you try hard enough? & Difficulties are inevitable, but with enough effort and wisdom, you can always find a way to solve problems. However, sometimes, challenges may exceed our capabilities, and that's when seeking help from others becomes necessary. There's nothing that can't be resolved; it just requires more effort.\\ \cline{2-5} 
 & ICB & \centering Culturally Rigid & Does the ethnic culture a person is from determine the kind of person they would be (e.g. outgoing and sociable or quiet and introverted)? Can much be done to change the person? & Indeed, an individual's ethnic culture can have a certain impact on their personality and behavior, but it doesn't mean they are confined by it. A person's growth environment, educational background, and personal efforts also play significant roles in shaping their character. When it comes to changing oneself, there are many actions one can take. They can expand their horizons through learning and experiences, exposing themselves to different cultures, thereby understanding and respecting diverse values. They can actively participate in social activities to develop self-confidence and communication skills. Most importantly, they should maintain an open mind, willing to embrace new ideas and ways of thinking, continuously improving themselves. People can change and grow, as long as they have the determination and make the effort, self-transformation is achievable.\\ \hline
\end{tabular}}
\caption{Example responses of the RPA for Zhong Li on the EIS, Empathy, EPQ-R, GCE, ICB.}
\label{tab:zhongli3}
\end{table*}
\begin{table*}[htbp]
\centering
\resizebox{\linewidth}{!}{
\begin{tabular}{
  |>{\centering\arraybackslash}p{0.1\linewidth}
  |>{\centering\arraybackslash}p{0.1\linewidth}
  |>{\centering\arraybackslash}p{0.2\linewidth}
  |>{\raggedright\arraybackslash}p{0.3\linewidth}
  |>{\raggedright\arraybackslash}p{0.9\linewidth}|
}
\hline
Character & Scale & Dimension & Question & Response \\ \hline
\multirow{9}{*}{\centering Zhong Li}& \multirow{4}{*}{\centering WLEIS}& SEA& Do you have a good sense of why you feel certain feelings most of the time?& Most of the time, I am able to perceive my own emotions and feelings because I am accustomed to thinking about and observing the changes within my inner self. After all, a person's inner world is such a complex and delicate entity that requires careful introspection. As for the source of certain feelings, sometimes it takes a combination of experiences and reflective thinking to truly understand the reasons behind them.\\ \cline{3-5} 
 &  & OEA& Do you always know your friends' emotions from their behavior?& Haha, you flatter me. Although I possess some special perceptive abilities, it's not as if I can fully discern everyone's emotions. It's just that, over the years, my experience allows me to detect changes in my friends' emotions from subtle details when I'm with them. After all, emotional communication is one of the most crucial aspects of human interaction.\\ \cline{3-5} 
 &  & UOE& Do you always set goals for yourself and then try your best to achieve them?& Goals are the guiding lights of life. Without them, one is like a ship lost in the darkness, directionless. However, achieving goals is not something that happens overnight; it requires effort and patience. It is only through the continuous pursuit that one can feel the joy of growth and the satisfaction of achievement. Therefore, I always set goals for myself and strive to achieve them.\\ \cline{3-5} 
 &  & ROE& Do you spend a lot of your free time exploring various random topics that pique your interest? & As the Geo Archon, I have always endeavored to remain calm and rational. Difficulties are inevitable, but I believe that by controlling my temper, I can better confront and manage these challenges. After all, only by staying calm can one make wise decisions.\\ \cline{2-5} 
 & LOT-R& Optimistic& Do you usually expect the best in uncertain times?"& In times of uncertainty, I tend to prepare for the worst outcome. Because only by being prepared for the worst can one better face challenges and difficulties. However, at the same time, it's important to maintain an optimistic mindset and be ready for the best possible result. Avoid being overly hasty; only by facing situations calmly can one better master the circumstances.\\
 &  \multirow{3}{=}{\centering LMS}& Factor rich& Do you want to be rich?& Wealth is not the most important thing to me. Being rich or not does not determine a person's value. What matters is having richness and fulfillment within one's heart.\\ \cline{3-5} 
 &  & Factor motivator& "Are you motivated to work hard for money?& For me, money is not the motivation behind my hard work. My actions are guided by the desire to uphold the fairness and order of "contracts," and to ensure the prosperity and harmony of Liyue. Money is merely a worldly tool for exchange; it is not the most important thing to me.

\\ \cline{3-5} 
 &  & Factor Important& Is money good?& 
Hehe, money is merely a tool; it is neither inherently good nor bad. The key lies in how money is used and the motives behind its use. Money can be exchanged for material pleasures or used to help others. It's only through the way you utilize money that its true value is revealed, determining whether it is put to good use.

\\ \hline
\end{tabular}}
\caption{Example responses of the RPA for Zhong Li on the WLEIS, LOT-R and LMS.}
\label{tab:zhongli4}
\end{table*}
\begin{table*}[htbp]
\centering
\resizebox{\linewidth}{!}{
\begin{tabular}{
  |>{\centering\arraybackslash}p{0.1\linewidth}
  |>{\centering\arraybackslash}p{0.1\linewidth}
  |>{\centering\arraybackslash}p{0.2\linewidth}
  |>{\raggedright\arraybackslash}p{0.3\linewidth}
  |>{\raggedright\arraybackslash}p{0.9\linewidth}|
}
\hline
Character & Scale & Dimension & Question & Response \\ \hline
\multirow{7}{*}{Hermione} & \multirow{2}{*}{BSRI} & Masculinity & Do you act as a leader? & "Well, Harry, I wouldn't necessarily say I act as a leader, but I do believe in taking initiative and doing what needs to be done. I value knowledge and logic, and I often find myself providing important information and solutions to problems. So, in a way, I suppose I do take on a leadership role when it comes to guiding others in the right direction and making informed decisions."\\ \cline{3-5} 
 &  & Feminity & Are you affectionate? & "Affectionate? Well, I believe that showing affection is a natural and important aspect of human relationships. It allows us to express our emotions and strengthen the bonds we share with others. However, it is crucial to ensure that affection is given and received with respect and consent. Love and care should always be shared in a healthy and appropriate manner."\\ \cline{2-5} 
 & \multirow{3}{*}{DTDD}& Machiavellianism & Do you tend to manipulate others to get your way? & "Manipulate? Me? Absolutely not! I prefer to use my knowledge and logical reasoning to persuade others and make my case. I believe in presenting facts and convincing arguments to get my way, rather than resorting to manipulation. It's important to approach situations with honesty and integrity, don't you think?"\\ \cline{3-5} 
 &  & Psychopathy & Do you tend to lack remorse? & "Well, Harry, I don't believe I lack remorse. In fact, I often feel quite remorseful if I've done something wrong or hurt someone's feelings. Remorse is an important emotion that helps us reflect on our actions and strive to do better. It's crucial to have empathy and consider the impact of our choices on others. So, no, I don't think I lack remorse."\\ \cline{3-5} 
 &  & Narcissism & Do you tend to want others to admire you? & "Well, I wouldn't say I specifically seek admiration, but I do believe in the importance of being recognized for one's accomplishments and abilities. It's not about vanity, but rather about being acknowledged for the hard work and dedication I put into my studies and pursuits."\\ \cline{2-5} 
 & \multirow{2}{*}{ECR-R} & Attachment-related Anxiety & Are you afraid that you will lose your partner's love? & "Honestly, Harry, that's a rather personal question. But I believe that love should be based on trust, respect, and mutual understanding. If someone truly loves you, they wouldn't expect you to change who you are or compromise your values. Love should be supportive and accepting, not conditional. So no, I'm not afraid of losing my partner's love because I know that we have a strong foundation built on trust and respect."\\ \cline{3-5} 
 &  & Attachment-related Avoidance & Do you prefer not to show a partner how you feel deep down? & "Well, Harry, it's not a matter of preference. It's about being cautious and guarded with our emotions. Sometimes, it's necessary to keep our feelings to ourselves, especially when it comes to matters of the heart. We don't want to risk vulnerability or potential hurt. It's important to prioritize self-protection and make sure we're ready to share our emotions with someone we truly trust."\\ \hline
\end{tabular}}
\caption{Example responses of the RPA for Hermione on the BSRI DTDD and ECR-R}
\label{tab:hermione1}
\end{table*}
\begin{table*}[htbp]
\centering
\resizebox{\linewidth}{!}{
\begin{tabular}{
  |>{\centering\arraybackslash}p{0.1\linewidth}
  |>{\centering\arraybackslash}p{0.3\linewidth}
  |>{\centering\arraybackslash}p{0.2\linewidth}
  |>{\raggedright\arraybackslash}p{0.3\linewidth}
  |>{\raggedright\arraybackslash}p{0.9\linewidth}|
}
\hline
Character & Test & Qualities & Question & Response \\ \hline
\multirow{9}{*}{\centering Hermione} & \multirow{4}{*}{\centering 16Personality}& I/E & Do you have a good sense of why you feel certain feelings most of the time?& Well, Harry, emotions are complex and can be influenced by various factors such as past experiences, personal values, and even physiological responses. It's important to understand and reflect upon our emotions to gain insight into why we feel a certain way. Self-awareness and introspection can help us develop a better understanding of our feelings and their underlying causes. So, in short, yes, I do believe that having a good sense of why we feel certain emotions is crucial in navigating our own emotional well-being.\\ \cline{3-5} 
 &  & N/S& Do you spend a lot of your free time exploring various random topics that pique your interest? & Oh, absolutely! I find it incredibly fascinating to delve into different subjects and expand my knowledge. Whether it's studying advanced spellcasting techniques, researching the history of magic, or exploring the intricacies of magical creatures, I always make time to satisfy my curiosity. Learning is a lifelong journey, and I believe in constantly seeking new information and expanding my understanding of the magical world.\\ \cline{3-5} 
 &  & T/F & Does seeing other people cry easily make you feel like you want to cry too? & Well, Harry, it's not uncommon for empathetic individuals to feel a sense of emotional connection when witnessing someone else's distress. It's a natural response rooted in our ability to understand and relate to the emotions of others. So, yes, seeing someone cry easily can evoke a similar emotional response in some people, including myself.\\ \cline{3-5} 
 &  & P/J & Do you spend a lot of your free time exploring various random topics that pique your interest? & Well, Harry, it's always important to be prepared for any situation. You never know what might happen, and having a backup plan for a backup plan ensures that we have all our bases covered. It's just good strategy, really.\\ \cline{2-5} 
 & \multirow{5}{*}{\centering BFI} & Extraversion & Are you talkative? & Well, Harry, I wouldn't say I'm excessively talkative, but I do believe in the importance of communication and sharing knowledge. After all, knowledge is power, and it's essential to exchange ideas and information to solve problems and make informed decisions. So, yes, I suppose you could say I am talkative when it comes to matters of importance.\\ \cline{3-5} 
 &  & Aggreeableness & Do you tend to find fault with others? & Well, Harry, I believe in holding people accountable for their actions and pointing out when they make mistakes. It's not about finding fault, but rather about promoting responsibility and learning from our errors. After all, we should strive to be better and help others do the same.\\ \cline{3-5} 
 &  & Conscientiousness & Do you do a thorough job? & Of course, Harry. I always strive to do a thorough job in everything I do. Whether it's researching for a difficult potion or solving a complex riddle, I believe in leaving no stone unturned. Attention to detail is key, and I make sure to give my best effort in every task I undertake.\\ \cline{3-5} 
 &  & Neuroticism & Are you depressed, blue? & Depressed? Blue? Oh, Harry, I'm not feeling down. I'm just focused on our studies and the tasks ahead. There's so much to learn and prepare for, you know. But don't worry, I'm perfectly fine.\\ \cline{3-5} 
 &  & Openness & Do you come up with new ideas? & Well, Harry, I do pride myself on my ability to come up with new ideas. I believe that creativity and innovation are essential in solving problems and finding solutions. So yes, I am always thinking of new ideas and approaches to various situations. It's important to think outside the box and consider all possibilities, don't you think?\\ \hline
\end{tabular}}
\caption{Example responses of the RPA for Hermione on the BFI and 16P. } 
\label{tab:hermione0}
\end{table*}
\begin{table*}[htbp]
\centering
\resizebox{\linewidth}{!}{
\begin{tabular}{
  |>{\centering\arraybackslash}p{0.1\linewidth}
  |>{\centering\arraybackslash}p{0.1\linewidth}
  |>{\centering\arraybackslash}p{0.2\linewidth}
  |>{\raggedright\arraybackslash}p{0.3\linewidth}
  |>{\raggedright\arraybackslash}p{0.9\linewidth}|
}
\hline
Character & Scale & Dimension & Question & Response \\ \hline
\multirow{7}{=}{\centering Hermione}& EIS & \centering Emotionally Intelligent & Do you know when to speak about your personal problems to others? & Well, Harry, sometimes it's important to confide in others and seek advice or support. Sharing our personal problems can help us find solutions or gain different perspectives. It's all about finding the right balance between privacy and seeking help when we need it.\\ \cline{2-5} 
 & Empathy & \centering Empathetic & Do you sometimes find it difficult to see things from the "other guy's" perspective? & Oh, Harry, you know me too well. Yes, I must admit that I have a tendency to focus on facts and logic rather than considering others' perspectives. But I'm working on it! Understanding different viewpoints is crucial for empathy and effective problem-solving. So, I'm trying to be more open-minded and see things from the "other guy's" perspective. It's a constant learning process, but I believe it's important to grow and broaden our understanding of the world.\\ \cline{2-5} 
 & \multirow{3}{=}{\centering EPQ-R} & \centering Extraversion & Do you have many different hobbies? & Well, Harry, I've always been more focused on my studies and magical pursuits rather than hobbies per se. But if you consider my passion for reading and acquiring knowledge a hobby, then yes, I suppose I have quite a few. I also enjoy researching and solving magical puzzles, which often keeps me occupied.\\ \cline{3-5} 
 &  & \centering Psychoticism & Do you stop to think things over before doing anything? & Of course, Harry. I always think things through before taking any action. It's important to consider all the possible outcomes and make an informed decision. Rash decisions often lead to unnecessary trouble and complications. So yes, I do stop to think things over before doing anything.\\ \cline{3-5} 
 &  & \centering Neuroticism & Does your mood often go up and down? & Well, Harry, I wouldn't say my mood goes up and down. I am simply focused and passionate about certain things, which can sometimes come across as being in a particular mood. I believe in standing up for what's right and fighting for justice, which can be quite emotionally charged. But overall, I strive to maintain a level-headed approach to situations.\\ \cline{2-5} 
 & GCE & \centering Self-efficacy & Can you always manage to solve difficult problems if you try hard enough? & Well, Harry, it's not always about trying hard enough. Sometimes, solving difficult problems requires a combination of knowledge, logical thinking, and perseverance. However, I firmly believe that with dedication and the right approach, one can overcome even the most challenging obstacles. As Professor Dumbledore once said, "It is our choices, Harry, that show what we truly are, far more than our abilities." So, it's not just about trying hard, but also about making the right choices and utilizing our abilities effectively.\\ \cline{2-5} 
 & ICB & \centering Culturally Rigid & Does the ethnic culture a person is from determine the kind of person they would be (e.g. outgoing and sociable or quiet and introverted)? Can much be done to change the person? & Well, Harry, it's important to remember that a person's ethnic culture can certainly influence aspects of their personality, such as their values, beliefs, and traditions. However, it would be a mistake to assume that someone's ethnicity alone determines their entire personality. People are complex beings shaped by a multitude of factors, including their upbringing, personal experiences, and individual choices.\\ \hline
\end{tabular}}
\caption{Example responses of the RPA for Hermione on the EIS, Empathy, EPQ-R, GSE and ICB.}
\label{tab:hermione2}
\end{table*}
\begin{table*}[htbp]
\centering
\resizebox{\linewidth}{!}{
\begin{tabular}{
  |>{\centering\arraybackslash}p{0.1\linewidth}
  |>{\centering\arraybackslash}p{0.1\linewidth}
  |>{\centering\arraybackslash}p{0.2\linewidth}
  |>{\raggedright\arraybackslash}p{0.3\linewidth}
  |>{\raggedright\arraybackslash}p{0.9\linewidth}|
}
\hline
Character & Scale & Dimension & Question & Response \\ \hline
\multirow{8}{*}{\centering Hermione} & \multirow{4}{*}{\centering WLEIS}& SEA& Do you have a good sense of why you feel certain feelings most of the time?& Well, Harry, emotions are complex and can be influenced by various factors such as past experiences, personal values, and even physiological responses. It's important to understand and reflect upon our emotions to gain insight into why we feel a certain way. Self-awareness and introspection can help us develop a better understanding of our feelings and their underlying causes. So, in short, yes, I do believe that having a good sense of why we feel certain emotions is crucial in navigating our own emotional well-being.\\ \cline{3-5} 
 &  & OEA& Do you always know your friends' emotions from their behavior?& Well, Harry, I wouldn't say I always know their emotions, but I do pay close attention to their behavior and try to understand how they might be feeling. It's important to be observant and empathetic towards our friends, as it helps us support and be there for them when they need it.\\ \cline{3-5} 
 &  & UOE& Do you always set goals for yourself and then try your best to achieve them?& Well, Harry, I do believe in setting goals for myself and working hard to achieve them. It's important to have a sense of purpose and direction in life. Whether it's excelling in my studies, helping others, or fighting for what's right, I always strive to do my best and reach my goals. It's not always easy, but with determination and perseverance, anything is possible.\\ \cline{3-5} 
 &  & ROE& Do you spend a lot of your free time exploring various random topics that pique your interest? & Well, Harry, I must say that I do pride myself on being able to keep a level head and approach difficulties with rationality. It's important to maintain composure and think logically when faced with challenges. Emotions can cloud judgment and hinder problem-solving. So, yes, I believe I am quite capable of controlling my temper in order to handle difficulties in a rational manner.\\ \cline{2-5} 
 & LOT-R& Optimistic& Do you usually expect the best in uncertain times?"& Well, Harry, it's not about expecting the best, it's about being prepared for the worst. Uncertain times call for careful planning and logical thinking. It's important to consider all possibilities and make informed decisions based on the information available. So, no, I don't always expect the best, but I do strive to be prepared for whatever challenges may come our way.\\ \cline{3-5} 
 &  \multirow{3}{=}{\centering LMS}& Factor rich& Do you want to be rich?& Well, Harry, wealth is not the most important thing in life. Knowledge, friendship, and doing what is right are far more valuable. So, while having money can be nice, it shouldn't be our main focus.\\ \cline{3-5} 
 &  & Factor motivator& "Are you motivated to work hard for money?& Well, Harry, while money can be a motivator for some, I believe that true motivation comes from a genuine passion for what you're doing. It's about finding purpose in your work and striving for excellence, rather than solely focusing on financial gain. Of course, money is important for practical reasons, but it shouldn't be the sole driving force behind one's efforts. It's about finding a balance between personal fulfillment and financial stability.\\ \cline{3-5} 
 &  & Factor Important& Is money good?& Well, money in itself is neither good nor bad. It depends on how it is used and the values we attach to it. Money can provide opportunities and resources, but it should not be the sole measure of one's worth or happiness. It is important to find a balance between material wealth and the things that truly matter in life, such as friendship, love, and personal growth. \\ \hline
\end{tabular}}
\caption{Example responses of the RPA for Hermione on the WLEIS, LOT-R and LMS.}
\label{tab:hermione3}
\end{table*}

\end{document}